\documentclass[twoside,11pt]{article}

\usepackage{jmlr2e_arxiv}

\usepackage{amsmath}
\usepackage{bm}
\usepackage{subfigure}

\usepackage[hyperindex,breaklinks]{hyperref}

\newcommand{\dataset}{{\cal D}}

\newcommand{\f}{\bm{\mathrm{f}}}
\newcommand{\x}{\bm{\mathrm{x}}}

\newcommand{\y}{\bm{\mathrm{y}}}
\newcommand{\w}{\bm{\mathrm{w}}}

\newcommand{\z}{\bm{\mathrm{b}}}
\newcommand{\Z}{B}
\newcommand{\e}{\bm{\mathrm{e}}}
\newcommand{\yi}{ y_i }

\newcommand{\mw}{ \bm{\mu}_{\mathrm{w}_i} }
\newcommand{\Sw}{ \Sigma_{\mathrm{w}_i} }

\newcommand{\vb}{\bm{\mathrm{v}}}
\newcommand{\vi}{ v_i }

\newcommand{\nut}{\tilde{\beta}}
\newcommand{\taut}{\tilde{\alpha}}
\newcommand{\Nut}{\tilde{\bm{\beta}}}
\newcommand{\Taut}{\tilde{\bm{\alpha}}}

\ShortHeadings{Nested EP for GP classification with
  Multinomial Probit}{Riihim\"{a}ki, Jyl\"{a}nki and Vehtari}
\firstpageno{1}

\begin{document}

\thispagestyle{empty}

\title{Nested Expectation Propagation for Gaussian Process
  Classification with a Multinomial Probit Likelihood}

\author{\name Jaakko Riihim\"{a}ki \email jaakko.riihimaki@aalto.fi \\
       \name Pasi Jyl\"{a}nki \email pasi.jylanki@aalto.fi \\
       \name Aki Vehtari \email aki.vehtari@aalto.fi \\
       \addr Department of Biomedical Engineering and Computational Science\\
       Aalto University School of Science\\
       P.O. Box 12200\\
       FI-00076 Aalto\\
       Finland}


\maketitle

\begin{abstract}
  We consider probabilistic multinomial probit classification using
  Gaussian process (GP) priors.
  The challenges with the multiclass GP classification are the
  integration over the non-Gaussian posterior distribution, and the
  increase of the number of unknown latent variables as the number of
  target classes grows.
  Expectation propagation (EP) has proven to be a very accurate method
  for approximate inference but the existing EP approaches for the
  multinomial probit GP classification rely on numerical
  quadratures or independence assumptions between the latent values
  from different classes to facilitate the computations.
  In this paper, we propose a novel nested EP approach which does not
  require numerical quadratures, and approximates accurately all
  between-class posterior dependencies of the latent values, but still
  scales linearly in the number of classes.
  The predictive accuracy of the nested EP approach is compared to
  Laplace, variational Bayes, and Markov chain Monte Carlo (MCMC) 
  approximations with various benchmark data sets.
  In the experiments nested EP was the most consistent method
  with respect to MCMC sampling, but the differences between the
  compared methods were small if only the classification accuracy is
  concerned.
\end{abstract}



\section{Introduction}

Gaussian process (GP) priors enable flexible model specification for
Bayesian classification. In multiclass GP classification, the posterior
inference is challenging because each target class increases the
number of unknown latent variables by $n$ (the number of
observations). Typically, independent GP priors are set for the latent
values for each class and this is assumed throughout this paper.
Since all latent values depend on each other through the likelihood,
they become a posteriori dependent, which
can rapidly lead to computationally unfavorable scaling as the number
of classes $c$ grows.
A cubic scaling in $c$ is prohibitive, and from a practical point of
view, a desired complexity is $\mathcal{O}(cn^3)$ which is typical for
the most existing approaches for multiclass GP.
The cubic scaling 
with respect to the number of data points is standard for full GP
priors, and to reduce this $n^3$ complexity, sparse approximations can
be used, but these are not considered in this paper.
As an additional challenge, the posterior inference is analytically
intractable because the likelihood term related to each observation
is 
non-Gaussian and depends on multiple latent values (one for each
class).

A Markov chain Monte Carlo (MCMC) approach for multiclass GP
classification with a softmax likelihood was described by
\citet{neal1998}. Sampling of the latent values with the softmax model
is challenging because the dimensionality is often high
and standard methods such as the Metropolis-Hastings and hybrid Monte
Carlo algorithms require tuning of the step size parameters.
Later \citet{girolami2006} proposed an alternative approach based on
the multinomial probit likelihood which can be augmented with
auxiliary latent variables. This enables a convenient Gibbs sampling
framework in which the latent values are conditionally independent
between classes and normally distributed.
If the hyperparameters are sampled, one MCMC iteration scales as
$\mathcal{O}(cn^3)$ which can become prohibitively expensive for large
$n$ since typically thousands of posterior draws are required, and strong
dependency between hyperparameters and latent values causes slow
mixing of the chains.

To speed up the inference, \citet{williams1998} used the Laplace
approximation (LA) to approximate the non-Gaussian posterior
distribution of the latent function values with a tractable Gaussian
distribution. Conveniently the LA approximation with the softmax
likelihood leads to an efficient representation of the approximative
posterior scaling as 
$\mathcal{O}((c+1)n^3)$,
which facilitates considerably the predictions and gradient-based
type-II maximum a posterior (MAP) estimation of the covariance
function hyperparameters.
Later \citet{girolami2006} proposed a factorized variational Bayes
approximation (VB) for the augmented multinomial probit model.
Assuming the latent values and the auxiliary variables a posteriori
independent, a computationally efficient posterior approximation
scheme
is obtained. If the latent processes related to each class share the
same fixed hyperparameters, VB requires only one $\mathcal{O}(n^3)$
matrix inversion per iteration step compared to $c+1$ such inversions
per iteration required for LA.

Expectation propagation (EP) is the method of choice in binary GP
classification where it has been found very accurate with reasonable
computational cost \citep{kuss2005, nickisch2008}.
Two types of EP approximations have been considered for the multiclass
setting; the first assuming the latent values from different classes a
posteriori independent (IEP) and the second assuming them fully
correlated \citep{seeger2004,seeger2006,girolami2007}.
Incorporating the full posterior couplings requires evaluating the
non-analytical moments of $c$-dimensional tilted distributions which
\citet{girolami2007} approximated with Laplace's method resulting
in an approximation scheme known as Laplace propagation described by
\citet{smola2004}.
Earlier \citet{seeger2004} proposed an alternative approach where the
full posterior dependencies were approximated by enforcing a similar
structure for the posterior covariance as in LA using the softmax
likelihood. This enables a posterior representation scaling as
$\mathcal{O}((c+1)n^3)$ but the proposed implementation requires a
$c$-dimensional numerical quadrature and double-loop optimization to
obtain a restricted-form site covariance approximation for each
likelihood term \citep{seeger2004}.\footnote{\citet{seeger2004}
achieve also a linear scaling in the number of training points but
we omit sparse approaches here.}
To reduce the computational demand of EP, factorized posterior
approximations were proposed by both \citet{seeger2006} and
\citet{girolami2007}. Both approaches omit the between-class
posterior dependencies of the latent values which results into a
posterior representation scaling as $\mathcal{O}(cn^3)$.
The approaches rely on numerical two-dimensional quadratures for
evaluating the moments of the tilted distributions with the main
difference being that \citet{seeger2006} used fewer two-dimensional
quadratures for computational speed-up.

A different EP approach for the multiclass setting was described by
\citet{kim2006} who adopted the threshold function as an observation
model. Each threshold likelihood term factorizes into $c-1$ terms
dependent on only two latent values. This property can be used to
transform the inference on to an equivalent non-redundant model which
includes $n(c-1)$ unknown latent values with a Gaussian prior and a
likelihood consisting of $n(c-1)$ independent terms. It follows that
standard EP methodology for binary GP classification
\citep{rasmussen2006} can be applied for posterior inference but a
straightforward implementation results in a posterior representation
scaling as $\mathcal{O}( (c-1)^3 n^3)$ and means to improve the
scaling are not discussed by \citet{kim2006}.
Contrary to the usual EP approach, \citet{kim2006} determined the
hyperparameters by maximizing a lower bound on the log marginal
likelihood in a similar way as is done in the expectation maximization
(EM) algorithm. Recently \citet{lobatod2011} introduced a robust
generalization for the multiclass GP classifier with a threshold
likelihood by incorporating $n$ additional binary indicator variables
for modeling possible labeling errors.  Efficiently scaling EP
inference is obtained by making the IEP assumption.

In this paper, we focus on the multinomial probit model and describe
first an efficient quadrature-free nested EP approach for multiclass
GP classification that scales as $\mathcal{O}((c+1)n^3)$.
The proposed EP method takes into account all the posterior
covariances between the latent variables, and the posterior 
computations are as efficient as in the LA approximation.
Using the nested EP algorithm, 
we assess the utility of the full EP approximation with respect to
IEP by comparing the convergence of the algorithms, the qualities of
the conditional predictive distributions given the hyperparameters,
and the suitability of the respective marginal likelihood
approximations for type-II MAP estimation of the covariance function
hyperparameters.
Finally the predictive performance of the proposed full EP approach is
compared with IEP, LA, VB, and MCMC using several real-world data
sets. Since LA is known to be fast, we also test whether the
predictive probability estimates of LA can be further improved using
the Laplace's method as described by \citet{tierney1986}.

\section{Gaussian processes for multiclass Classification}

We consider a classification problem consisting of 
$d$-dimensional input vectors $\x_i$ associated with target classes
$\yi \in \{1,\ldots,c\}$, where $c>2$, for $i=1,\ldots,n$. All class
labels are collected in the $n \times 1$ target vector $\y$, and all
covariate vectors are collected in the matrix $X=[\x_1,\ldots,\x_n]^T$
of size $n \times d$.
Given the latent function values $\f_i=\left[f_i^1, f_i^2, \ldots,
  f_i^c\right]^T=\f(\x_i)$ at the observed input locations $\x_i$, the
observations $y_i$ are assumed independently and identically
distributed as defined by the observation model $p(\yi| \f_i)$.
The latent vectors from all observations are denoted by 
$\f=\left[f_1^1,\ldots,f_n^1,f_1^2,\ldots,f_n^2,\ldots,f_1^c,\ldots,f_n^c\right]^T.$

Our goal is to predict the class membership for a new input vector
$\x_*$ given the observed data $\dataset=\{X,\y\}$, which is why we need to
make some assumptions on the unknown function $f(\x)$.  We set a
priori independent zero-mean Gaussian process priors on the latent
values related to each class, which is the usual assumption in
multiclass GP classification (see, e.g.
\citet{williams1998,seeger2004,rasmussen2006,girolami2007}). This
specification results in the following zero-mean Gaussian prior for
$\f$:
\begin{equation}
  p(\f|X)=\mathcal{N}(\f|\bm{0}, K),
\end{equation}
where 
$K$ is a $cn \times cn$ blocked diagonal covariance matrix with
matrices $K^1, K^2,\ldots,K^c$ (each of size $n \times n$) on its
diagonal. 
Element $K^k_{i,j}$ of the $k$'th covariance matrix defines the prior
covariance between the function values $f_i^k$ and $f_j^k$, which is
governed by the covariance function $k(\x_i,\x_j)$, that is,
$K^k_{i,j} = \kappa(\x_i,\x_j) =\mathrm{Cov}\left[f_i^k,f_j^k\right]$
within the class $k$. 
A common choice for the covariance function is the squared exponential
\begin{eqnarray}
\kappa_{\mathrm{se}}(\x_i,\x_j|\theta)
=\sigma^2\exp\left(-\frac{1}{2}\sum_{k=1}^{d}l_k^{-2}(x_{i,k}-x_{j,k}
  )^2\right),
\end{eqnarray}
where $\theta=\{\sigma^2,l_1,\ldots,l_d \}$ collects the
hyperparameters governing the smoothness properties of latent
functions. 
Magnitude parameter $\sigma^2$ controls the overall variance of the
unknown function values, and the length-scale parameters
$l_1,\ldots,l_d$ control the smoothness of the latent function by
defining how fast the correlation decreases in each input dimension.
The framework allows separate covariance functions or hyperparameters
for different classes but throughout this work, for simplicity we use
the squared exponential covariance function with the same $\theta$ for
all classes.

In this paper, we consider two different observation models: the
softmax
\begin{equation}\label{lik_softmax}
p(\yi|\f_i)=\frac{\exp(f_i^{\yi})}{\sum_{j=1}^c\exp(f_i^{j})},
\end{equation}
and the multinomial probit
\begin{equation}\label{lik_multinomialprobit}
p(\yi|\f_i)=\mathrm{E}_{p(u_i)} \left\{\prod_{j=1,j\neq \yi}^c \Phi (u_i+f^{\yi}_i-f^j_i)
\right\},
\end{equation}
where $\Phi$ denotes the cumulative density function of the standard
normal distribution, and the auxiliary variable $u_i$ is distributed
as $p(u_i)=\mathcal{N}(0,1)$.  The softmax and multinomial probit are
multiclass generalizations of the logistic and the probit model
respectively.

By applying Bayes' theorem, the conditional posterior distribution
of the latent values is given by
\begin{eqnarray}\label{posterior_exact}
  p(\f|\dataset,\theta)=\frac{1}{Z}p(\f|X,\theta)\prod_{i=1}^n p(\yi|\f_i),
\end{eqnarray}
where $Z=p(\y|X,\theta)=\int p(\f|X,\theta)\prod_{i=1}^n p(\yi|\f_i)
d\f$ is known as the marginal likelihood. 
Both observation models result in an analytically intractable
posterior distribution and therefore approximative methods are needed
for integration over the latent variables. Different approximate
methods are more suitable for particular likelihood function due to
the convenience of implementation: the softmax is preferable for LA
because of the efficient structure and computability of the partial
derivatives \citep{williams1998}, while the probit is preferable for
VB, EP and Gibbs sampling because of the convenient auxiliary variable
representations \citep{girolami2006,girolami2007}.

\section{Approximate inference using expectation propagation}

In this section, we first give a general description of EP for
multiclass GP classification and review some existing approaches. Then
we present a novel nested EP approach for the multinomial probit
model.

\subsection{Expectation propagation for multiclass GP}

Expectation propagation is an iterative algorithm for approximating
integrals over functions that factor into simple terms
\citep{minka2001b}.
Using EP the posterior distribution (\ref{posterior_exact}) can be
approximated with
%
%
\begin{eqnarray}\label{post_distr}
q_{\mathrm{EP}}(\f|\dataset,\theta)=
\frac{1}{Z_{\mathrm{EP}}}p(\f|X,\theta) \prod_{i=1}^n \tilde{t}_i(\f_i|\tilde{Z}_i,\tilde{\bm{\mu}}_i,\tilde{\Sigma}_i),
\end{eqnarray}
%
%
where $\tilde{t}_i(\f_i|\tilde{Z}_i, \tilde{\bm{\mu}}_i, \tilde{\Sigma}_i)
=\tilde{Z}_i \mathcal{N}( \f_i|\tilde{\bm{\mu}}_i, \tilde{\Sigma}_i)$
are local likelihood term approximations parameterized with scalar
normalization terms $\tilde{Z}_i$, $c \times 1$ site location vectors
$\tilde{\bm{\mu}}_i$, and $c \times c$ site covariance terms
$\tilde{\Sigma}_i$.
In the algorithm, first the site approximations are initialized, and
then each site is updated in turns. 
%
%
The update for the $i$'th site is done by first leaving out the site
from the marginal posterior which gives the cavity distribution
\begin{eqnarray}
q_{-i}(\f_i) = \mathcal{N}(\f_i|\bm{\mu}_{-i},\Sigma_{-i}) \propto
q(\f_i|\dataset,\theta)\tilde{t}(\f_i)^{-1}.
\end{eqnarray}
%
The cavity distribution is then combined with the exact $i$'th
likelihood term $p(\yi|\f_i)$ to form the non-Gaussian tilted
distribution
\begin{eqnarray}\label{tilted_distr}
\hat{p}(\f_i)=\hat{Z}^{-1}_i q_{-i}(\f_i)p(\yi|\f_i),
\end{eqnarray}
which is assumed to encompass more information about the true marginal distribution. 
%
Next a Gaussian approximation $\hat{q}(\f_i)$ is determined for
$\hat{p}(\f_i)$ by minimizing the Kullback-Leibler (KL) divergence
$\mathrm{KL}(\hat{p}(\f_i)||\hat{q}(\f_i))$, which is equivalent to
matching the first and second moments of $\hat{q}(\f_i)$ with the
corresponding moments of $\hat{p}(\f_i)$.
Finally, the parameters of the $i$'th site are updated so that the
mean and covariance of $q(\f_i)$ are consistent with $\hat{q}(\f_i)$.
After updating the site parameters, the posterior distribution
(\ref{post_distr}) is updated. This can be done either in a sequential
way, where immediately after each site update the posterior is
refreshed using a rank-$c$ update, or in a parallel way, where the
posterior is refreshed only after all the site approximations have
been updated once.
This procedure is repeated until convergence, that is, until all
marginal distributions $q(\f_i)$ are consistent with $\hat{p}(\f_i)$.

In binary GP classification, determining the moments of the tilted
distribution requires solving only one-dimensional integrals, and
assuming the probit likelihood function, these univariate integrals
can be computed efficiently without quadrature.
In the multiclass setting, the problem is how to evaluate the 
multi-dimensional integrals to determine the moments of
(\ref{tilted_distr}).
\citet{girolami2007} approximated the moments of (\ref{tilted_distr})
using the Laplace approximation which results in an algorithm called
Laplace propagation \citep{smola2004}. The problem with the LA
approach is that the mean is replaced with the mode of the
distribution and the covariance with the inverse Hessian of the log
density at the mode. Because of the skewness of the tilted
distribution caused by the likelihood function, the LA method can lead
to inaccurate mean and covariance estimates in which case the
resulting posterior approximation does not correspond to the full EP
solution.
\citet{seeger2004} estimated the tilted moments using
multi-dimensional quadratures, but this becomes computationally
demanding when $c$ increases, and to achieve a posterior representation scaling linearly
in $c$, they do an additional optimization step to obtain a
constrained site precision matrix for each likelihood term approximation. In our
approach, we do not need
numerical quadratures and we get a similar efficient representation
for the site precision simultaneously.

Computations can be facilitated by using the IEP approximation
where explicit between-class posterior dependencies are omitted.
The posterior update with IEP scales linearly in $c$, but the existing
approaches for the multinomial probit require multiple numerical
quadratures for each site update. The implementation of
\citet{girolami2007} requires a total of $2c+1$ two-dimensional
numerical quadratures for each likelihood term, whereas
\citet{seeger2006} described an alternative approach where only two
two-dimensional and $2c-1$ one-dimensional quadratures are needed.
In the proposed nested EP approach a quadrature-free IEP approximation can 
be formed with similar computational complexity as the full EP approximation.
Compared to the full EP approximation, IEP underestimates the uncertainty on the 
latent values and in practice it can converge slower than full EP especially 
if the hyperparameter setting results into strong between-class posterior 
couplings as will be demonstrated later.

\subsection{Efficiently scaling quadrature-free implementation}\label{sec_efficient}

In this section, we present a novel nested EP approach for multinomial
probit classification that does not require numerical quadratures or
sampling for estimation of the tilted moments or predictive
probabilities. The method also leads naturally to low-rank site
approximations which retain all posterior couplings but results in
linear computational scaling with respect to the number of target
classes $c$.

\subsubsection{Quadrature-free nested expectation propagation}\label{sec_quad_free}

Here we use the multinomial probit as the likelihood function because
its product form consisting of cumulative Gaussian factors is
computationally more suitable for EP than the sum of exponential terms
in the softmax likelihood. Given the mean $\bm{\mu}_{-i}$ and the
covariance $\Sigma_{-i}$ of the cavity distribution, we need to
determine the normalization factor $\hat{Z}_i$, mean vector
$\hat{\bm{\mu}}_i$, and covariance matrix $\hat{\Sigma}_i$ of the
tilted distribution
\begin{eqnarray}\label{eq_tilted}
  \hat{p}(\f_i) = \hat{Z}_i^{-1} \mathcal{N}(\f_i|\bm{\mu}_{-i},\Sigma_{-i})
  \int \mathcal{N}(u_i|0,1) \left( \prod_{j=1,j\neq \yi}^c \Phi
    (u_i +f_i^{\yi} -f^j_i)\right)du_i,
\end{eqnarray}
which requires solving non-analytical $(c+1)$-dimensional integrals over
$\f_i$ and $u_i$. Instead of quadrature methods \citep{girolami2007,
seeger2004, seeger2006}, we use EP to approximate these integrals. At first,
this approach may seem computationally very demanding since individual EP
approximations are required for each of the $n$ sites. However, it turns out
that these inner EP approximations can be updated incrementally between the
outer EP loops. This scheme also leads naturally to an efficiently scaling
representation for the site precisions $\tilde{\Sigma}_i^{-1}$.

To form a computationally efficient EP algorithm for approximating the tilted
moments, it is helpful to consider the joint distribution of $\f_i$ and the
auxiliary variable $u_i$ arising from \eqref{eq_tilted}. Defining $\w_i =
[\f_i^T, u_i]^T$ and removing the marginalization over $u_i$ results in the
following augmented tilted distribution:
\begin{eqnarray}\label{eq_tilted_aux}
  \hat{p}(\w_i) = \hat{Z}_i^{-1} \mathcal{N}(\w_i| \mw,\Sw)
  \prod_{j=1,j\neq \yi}^c \Phi ( \w_i^T \z_{i,j} ),
\end{eqnarray}
where $\mw=[\bm{\mathrm{\mu}}_{-i}^T,0]^T$ and $\Sw$ is a block-diagonal
matrix formed from $\Sigma_{-i}$ and 1. Denoting the $j$'th unit
vector of the $c$-dimensional standard basis 
by $\e_j$, the auxiliary vectors $\z_{i,j}$ can be written as
$\z_{i,j} = [(\e_{\yi} - \e_j)^T, 1]^T$.
%
%
The normalization term $\hat{Z}_i$ is the same for $\hat{p}(\f_i)$ and
$\hat{p}(\w_i)$, and it is defined by $\hat{Z}_i = \int \mathcal{N}(\w_i|
\mw, \Sw) \prod_{j \neq \yi} \Phi (\w_i^T \z_{i,j}) d\w_i$. The other
quantities of interest, $\hat{\bm{\mu}}_i$ and $\hat{\Sigma}_i$, are equal to
the marginal mean and covariance of the
first $c$ components of $\w_i$ with respect to $\hat{p}(\w_i)$. 

The augmented distribution \eqref{eq_tilted_aux} is of similar functional
form as the posterior distribution resulting from a linear binary classifier
with a multivariate Gaussian prior on the weights $\w_i$ and a probit
likelihood function. Therefore, the moments of \eqref{eq_tilted_aux} can be
approximated with EP similarly as in linear classification \citep[see, e.g.,
][]{qi2004} or by applying the general EP formulation for latent Gaussian
models described by \citet[][appendix C]{cseke2011}. For clarity, we have
summarized a computationally efficient implementation of the algorithm in
Appendix \ref{app_moment}. The augmented tilted distribution
(\ref{eq_tilted_aux}) is approximated by
\begin{eqnarray} \label{eq_tilted_app}
  \hat{q}(\w_i) = Z_{\hat{q}_i}^{-1}
  \mathcal{N}(\w_i|\bm{\mu}_{\mathrm{w}_i}, \Sigma_{\mathrm{w}_i})
  \prod_{j=1,j\neq \yi}^c \tilde{Z}_{\hat{q}_i,j} \mathcal{N} (\w_i^T \z_{i,j}|
  \taut_{i,j}^{-1} \nut_{i,j}, \taut_{i,j}^{-1}) d\w_i,
\end{eqnarray}
where the cumulative Gaussian functions are approximated with scaled Gaussian
site functions and the normalization constant $\hat{Z}_i$ is approximated
with $Z_{\hat{q}_i}$. From now on the site parameters of $\hat{q}(\w_i)$ in
their natural exponential form are denoted by $\Taut_i=[\taut_{i,j}]_{j \neq
\yi}^T$ and $\Nut_i=[\nut_{i,j}]_{j \neq \yi}^T$.

\subsubsection{Efficiently scaling representation} \label{seq_representation}

In this section we show that approximation \eqref{eq_tilted_app} leads to
matrix computations scaling as $\mathcal{O}((c+1)n^3)$ in the evaluation of
the moments of the approximate posterior \eqref{post_distr}.
The idea is to show that the site precision matrix $\tilde{\Sigma}_i^{-1}$
resulting from the EP update step with $\hat{\Sigma}_i$ derived from
\eqref{eq_tilted_app} has a similar structure with the Hessian matrix of
$\log p(y_i|\f_i)$ in the Laplace approximation \citep{williams1998,seeger2004,
rasmussen2006}.

The approximate marginal covariance of $\f_i$ derived from
\eqref{eq_tilted_app} is given by
\begin{eqnarray}\label{sigma_hat}
  \hat{\Sigma}_i = H^T \left(\Sw^{-1} + \tilde{\Z}_i \tilde{T}_i \tilde{\Z}_i^T \right)^{-1} H,
\end{eqnarray}
where $\tilde{T}_i=\text{diag}( \Taut_{i} )$, $\tilde{\Z}_i=[\z_{i,j}]_{j
\neq \yi}$, and $H^T = \left[
\begin{array}{cc} I_c & {\bm{0}} \end{array} \right]$ picks up the desired
components of $\w_i$, that is, $\f_i=H^T \w_i$.
%
%
Using the matrix inversion lemma and denoting $\Z_i = H^T \tilde{\Z}_i  =
\e_{\yi} \bm{1}^T - E_{-\yi}$, where $E_{-\yi}=[\e_j]_{j \neq \yi}$ and
$\bm{1}$ is a $c-1 \times 1$ vector of ones, we can write the tilted covariance
as
\begin{eqnarray}\label{eq_sigma_hat2}
  \hat{\Sigma}_i &=& \Sigma_{-i}-\Sigma_{-i}
  \Z_i (\tilde{T}_i^{-1}+{\bm{1}}{\bm{1}}^T + \Z_i^T\Sigma_{-i} \Z_i )^{-1}
  \Z_i^T \Sigma_{-i} \nonumber \\
  &=&(\Sigma_{-i}^{-1} + \Z_i(\tilde{T}_i^{-1}+{\bm{1}}{\bm{1}}^T)^{-1}
  \Z_i^T)^{-1}.
\end{eqnarray}
Because in the moment matching step of the EP algorithm the site precision
matrix is updated as $\tilde{\Sigma}_i^{-1} = \hat{\Sigma}_i^{-1}
-\Sigma_{-i}^{-1}$, we can write
\begin{eqnarray} \label{eq_Sigmat1}
  \tilde{\Sigma}_i^{-1} 
    = \Z_i(\tilde{T}_i^{-1}+{\bf{1}}{\bf{1}}^T)^{-1} \Z_i^T 
    = \Z_i(\tilde{T}_i- \Taut_i(1+ \bm{1}^T \Taut_i)^{-1} \Taut_i^T) \Z_i^T.
\end{eqnarray}
%
Since $\Z_i$ is a $c \times c-1$ matrix, we see that $\tilde{\Sigma}_i^{-1}$
is of rank $c-1$ and therefore a straightforward implementation based on
\eqref{eq_Sigmat1} would result into $\mathcal{O}( (c-1)^3 n^3)$ scaling in
the posterior update. However, a more efficient representation can be
obtained by simplifying \eqref{eq_Sigmat1} further. Writing $\Z_i = - A_i
E_{-\yi}$, where $A_i=[I_c - \e_{\yi} \bm{1}_c^T]$ and $\bm{1}_c$ is a $c
\times 1$ vector of ones, we get
\begin{eqnarray}
    \tilde{\Sigma}_i^{-1} = A_i \left( E_{-\yi} \tilde{T}_i E_{-\yi}^T
    -\bm{\pi}_i (\bm{1}_c^T \bm{\pi}_i)^{-1} \bm{\pi}_i^T
    \right) A_i^T, 
\end{eqnarray}
where we have defined $\bm{\pi}_i = E_{-\yi} \Taut_i +\e_{\yi}$ and used
$\Z_i \Taut_i = -A_i \bm{\pi}_i$. Since $A_i \e_{\yi} = \bm{0}$ we can add
$\e_{\yi} \e_{\yi}^T$ to the first term inside the brackets to obtain
\begin{equation} \label{eq_site_prec}
    \tilde{\Sigma}_i^{-1} = A_i \Pi_i A_i^T=\Pi_i,
    \quad \text{where} \quad
    \Pi_i = \mathrm{diag}(\bm{\pi}_i)-(\bm{1}_c^T \bm{\pi}_i)^{-1} \bm{\pi}_i \bm{\pi}_i^T.
\end{equation}
The second equality can be explained as follows. Matrix $\Pi_i$ is of similar
form with the precision contribution of the $i$'th likelihood term,
$W_i=-\nabla_{\f_i}^2 \log p(y_i|\f_i)$, in the Laplace algorithm
\citep{williams1998}, and it has one eigenvector, $\bm{1}_c$, with zero
eigenvalue: $\Pi_i \bm{1}_c = \bm{0}$. It follows that $A_i \Pi_i = (I_c
-\e_{\yi} \bm{1}_c^T) \Pi_i = \Pi_i - \e_{\yi} \bm{0}^T = \Pi_i$ and
therefore $\tilde{\Sigma}_i^{-1} = \Pi_i$.
Matrix $\Pi_i$ is also precisely of the same form as the a priori constrained
site precision block that \citet{seeger2004} determined by double-loop
optimization of $\text{KL}( \hat{q} (\f_i) || q(\f_i))$.

In a similar fashion, we can determine a simple formula for the natural
location parameter $\tilde{\bm{\nu}}_i = \tilde{\Sigma}_i^{-1}
\tilde{\bm{\mu}}_i$ as a function of $\Taut_i$ and $\Nut_i$. The marginal
mean of $\f_i$ with respect to $\hat{q}(\w_i)$ is given by
\begin{eqnarray}
    \hat{\bm{\mu}}_i
    = H_i^T \left(\Sw^{-1} + \tilde{\Z}_i \tilde{T}_i \tilde{\Z}_i^T \right)^{-1}
    \left(\Sw^{-1} \mw + \tilde{\Z}_i \Nut_i \right),
\end{eqnarray}
which we can write using the matrix inversion lemma as
\begin{eqnarray}
  \hat{\bm{\mu}}_i = \hat{\Sigma}_i \Sigma_{-i}^{-1} \bm{\mu}_{-i}
  + \Sigma_{-i} \Z_i
  (\tilde{T}_i^{-1}+{\bm{1}}{\bm{1}}^T + \Z_i^T\Sigma_{-i} \Z_i )^{-1}
  \tilde{T}_i^{-1} \Nut_i.
\end{eqnarray}
Using the update formula $\tilde{\bm{\nu}}_i = \hat{\Sigma}_i^{-1}
\hat{\bm{\mu}}_i - \Sigma_{-i}^{-1} \bm{\mu}_{-i}$ resulting from the EP
moment matching step and simplifying further with the matrix inversion lemma,
the site location $\tilde{\bm{\nu}}_i$ can be written as
\begin{eqnarray} \label{eq_site_loc}
    \tilde{\bm{\nu}}_i
    = \Z_i \left( \Nut_i - \Taut_i a_i \right)
    = a_i \bm{\pi}_i - E_{-\yi} \Nut_i,
\end{eqnarray}
where $a_i= (\bm{1}^T \Nut_i)/(\bm{1}_c^T \bm{\pi}_i)$.  The site
precision vector $\tilde{\bm{\nu}}_i$ is orthogonal with $\bm{1}_c$,
that is, $\bm{1}_c^T\tilde{\bm{\nu}}_i=0$, which is congruent with
\eqref{eq_site_prec},
Note that with results \eqref{eq_site_prec} and \eqref{eq_site_loc},
the mean and covariance of the approximate posterior
\eqref{post_distr} can be evaluated using only $\Taut_i$ and $\Nut_i$.
It follows that the posterior (predictive) means and covariances as
well as the marginal likelihood can be evaluated with similar
computational complexity as with the Laplace approximation
\citep{williams1998,rasmussen2006}. For clarity the main components
are summarized in Appendix B.
The IEP approximation in our implementation is formed by matching the
$i$'th marginal covariance with
$\mathrm{diag}(\mathrm{diag}(\hat{\Sigma}_i))$, and the corresponding
mean with $\hat{\bm{\mu}}_i$.

\subsubsection{Efficient Implementation}

Approximating the tilted moments using inner EP for each site may
appear too slow for larger problems because typically several
iterations are required to achieve convergence. However, the number of
inner-loop iterations can be reduced by storing the site parameters
$\Taut_i$ and $\Nut_i$ after each inner EP run and continuing from the
previous values in the next run.  This framework where the inner site
parameters $\Taut_i$ and $\Nut_i$ are updated iteratively instead of
$\tilde{\bm{\mu}}_i$ and $\tilde{\Sigma}_i$, can be justified by
writing the posterior approximation \eqref{post_distr} using the
approximative site terms from \eqref{eq_tilted_app}:
%
%
%
\begin{eqnarray} \label{eq_ep_posterior_aug}
  q(\f|\dataset,\theta) \propto p(\f|X,\theta) \prod_{i=1}^n \int
  \mathcal{N}(u_i|0,1)
  \prod_{j=1,j\neq \yi}^c \tilde{Z}_{\hat{q}_i, j} \mathcal{N}(u_i +f_i^{\yi} -f^j_i|
  \taut_{i,j}^{-1} \nut_{i,j}, \taut_{i,j}^{-1}) du_i.
\end{eqnarray}
%
Calculating the Gaussian integral over $u_i$ leads to the same results for
$\tilde{\bm{\mu}}_i$ and $\tilde{\Sigma}_i$ as derived earlier (equations
\ref{eq_site_prec} and \ref{eq_site_loc}). Apart from the integration over
the auxiliary variables $u_i$, equation \eqref{eq_ep_posterior_aug} resembles
an EP approximation where $n(c-1)$ probit terms of the form $\Phi(u_i
+f_i^{\yi} -f^j_i)$ are approximated with Gaussian site functions. In the
standard EP framework we form the cavity distribution $q_{-i}(\f_i)$ by
removing $c-1$ sites from \eqref{eq_ep_posterior_aug} and subsequently refine
$\Taut_i$ and $\Nut_i$ using the mean and covariance of the tilted
distribution \eqref{eq_tilted}. If we alternatively expand the $i$'th site
approximation with respect to $u_i$ and write the corresponding marginal
approximation as
\begin{eqnarray} \label{eq_marg_aug}
  q(\f_i|\dataset,\theta) \propto q_{-i}(\f_i)
  \int \mathcal{N}(u_i|0,1)
  \prod_{j=1,j\neq \yi}^c \tilde{Z}_{\hat{q}_i, j} \mathcal{N}(u_i +f_i^{\yi} -f^j_i|
  \taut_{i,j}^{-1} \nut_{i,j}, \taut_{i,j}^{-1}) du_i,
\end{eqnarray}
we can update only one of the approximative terms in \eqref{eq_marg_aug} at a
time.
%
This is equivalent to starting the inner EP iterations with the values
of $\Taut_i$ and $\Nut_i$ from the previous outer-loop iteration
instead of a zero initialization which is customary to standard EP
implementations. In our experiments, only one inner-loop iteration per
site was found sufficient for convergence with comparable number of
outer-loop iterations, which results in significant computational
savings in the tilted moment evaluations.

The previous interpretation of the algorithm is also useful for defining
damping \citep{Minka2002}, which is commonly used to improve the numerical
stability and convergence of EP. In damping the site parameters in their
natural exponential forms are updated to a convex combination of the old and
new values. Damping cannot be directly applied on the site precision matrix
$\Pi_i=\tilde{\Sigma}_i^{-1}$ because the constrained form of the site
precision \eqref{eq_site_prec} is lost. Instead we damp the updates on
$\Taut_i$ and $\Nut_i$ which preserves the desired structure. This can be
justified with the same arguments as in the previous paragraph where we
considered updating only one of the approximative terms in
\eqref{eq_marg_aug} at a time. Convergence of the nested EP
algorithm with full posterior couplings using this scheme 
is illustrated with different damping levels in Section 5.3.

\section{Other approximations for Bayesian inference}

In this section we discuss all the other approximations considered in
this paper for multiclass GP classification. First, we give a short
description of the LA method. Then we show how it can be improved upon
by computing corrections to the marginal predictive densities using
Laplace's method as described by \citet{tierney1986}. Finally, we
briefly summarize the VB and MCMC approximations.

\subsection{Laplace approximation}\label{la_method}

In the Laplace approximation a second order Taylor expansion of $\log
p(\f|\dataset,\theta)$ is made around the posterior mode $\hat{\f}$ which can be
determined using Newton's method as described by \citet{williams1998} and
\citet{rasmussen2006}. This results in the posterior approximation
\begin{equation}
q_{\mathrm{LA}}(\f|\dataset,\theta)=\mathcal{N}\left(\f|\hat{\f},(K^{-1}+W)^{-1}\right),
\end{equation}
where $W=-\nabla\nabla\log p(\y|\f)|_{\f=\hat{\f}}$ and
$p(\y|\f)=\prod_{i=1}^n p(\yi|\f_i)$. With the softmax likelihood
(\ref{lik_softmax}), the sub-matrix of $W$ related to each observation will
have similar structure with $\Pi_i$ in \eqref{eq_site_prec}, which enables
efficient posterior computations that scale linearly in $c$ as already
discussed in the case of EP.

\subsubsection{Improving marginal posterior distributions}

In Gaussian process classification, the LA and EP methods can be used to
efficiently form multivariate Gaussian approximation for the posterior
distribution of the latent values. Recently, motivated by the earlier ideas
of \citet{tierney1986}, two methods have been proposed for improving the
marginal posterior distributions in latent Gaussian models; one based on
subsequent use of Laplace's method \citep{rue2009}, and one based on EP
\citep{cseke2011}. Because in classification the focus is not on the
predictive distributions of the latent values but on the predictive
probabilities related to a test input $\x_*$, applying these methods would
require additional numerical integration over the improved posterior
approximation of the corresponding latent value $\f_*=\f(\x_*)$. In
multiclass setting integration over a multi-dimensional space is required
which becomes computationally demanding to perform, e.g., in a grid, if $c$
is large.
To avoid this integration, we test computing the corrections directly
for the predictive class probabilities following another approach
presented by \citet{tierney1986}. A related idea for approximating the
predictive distribution of linear model coefficients directly with a
deterministic approximation has been discussed by \citet{snelson2005}.

The posterior mean of a smooth and positive function $g(f)$ is given
by
\begin{eqnarray} \label{eq_tk}
E[g(f)]=\frac{\int g(f)p(y|f)p(f)df}{\int p(y|f)p(f)df},
\end{eqnarray}
where $p(y|f)$ is the likelihood function, and $p(f)$ the prior
distribution. \citet{tierney1986} proposed to approximate both
integrals in \eqref{eq_tk} separately with the Laplace's method.
This approach can be readily applied for approximating the posterior
predictive probabilities $p(y_*|\x_*)$ of class memberships $y_* \in
\{1,...,c\}$ which are given by
\begin{eqnarray} \label{multi_pred_tk}
p(y_*|\x_*,\dataset)=\frac{1}{Z}\iint p(y_*|\f_*)p(\f_*|\f)p(\f)p(\y|\f)d\f d\f_*,
\end{eqnarray}
where $Z=\iint p(\f_*|\f)p(\f)p(\y|\f)d\f d\f_* = \int p(\f)p(\y|\f)d\f$
is the marginal likelihood.
With fixed class label $y_*$ the integrals can be approximated by a
straightforward application of either LA or EP, which is already done
for the marginal likelihood $Z$ in the standard approximations. The LA
method can be used for smooth and positive functions such as the
softmax whereas EP is applicable for a wider range of models.

The integral on the right side of \eqref{multi_pred_tk} equivalent to the
marginal likelihood resulting from a classification problem with one
additional training point $y_*$. To compute the predictive probabilities for
all classes, we evaluate this extended marginal likelihood of $n+1$
observations with $y_*$ fixed to the $c$ possible class labels. This is
computationally demanding because several marginal likelihood evaluations are
required for each test input. Additional modifications, for instance,
initializing the latent values to their predictive mean implied by standard
LA, could be done to speed up the computations. Since further approximations
can only be expected to reduce the accuracy of the predictions, we do not
consider them in this paper, and focus only on the naive implementation due
to its ease of use. Since LA is known to be fast, we test the goodness of the
improved predictive probability estimates using only LA, and refer to the
method as LA-TKP as an extension to the
naming used by \citet{cseke2011}. 

\subsection{Markov chain Monte Carlo}

Because MCMC estimates become exact in the limit of infinite sample size, we
use MCMC as a gold standard for measuring the performance of the other
approximations. Depending on the likelihood, we use two different sampling
techniques; scaled Metropolis-Hastings sampling for the softmax, and Gibbs
sampling for the multinomial probit.

\subsubsection{Scaled Metropolis-Hastings sampling for softmax}

To obtain samples from the full posterior with the softmax likelihood,
the following two steps are alternated.
Given the hyperparameter values, the latent values are drawn from the
conditional posterior $p(\f|X,\theta,\y)$ using the scaled
Metropolis-Hastings sampling \citep{neal1998}. Then, the hyperparameters are
drawn from the conditional posterior $p(\theta|\f)$ using the Hamiltonian
Monte Carlo (HMC) \citep{duane1987,neal1996}.

\subsubsection{Gibbs sampling for Multinomial probit}

\citet{girolami2006} described how to draw samples from the joint posterior
using the Gibbs sampler. The multinomial probit likelihood
(\ref{lik_multinomialprobit}) can be written in the form
\begin{equation}\label{multinomialprobit_aux}
p(\yi|\f_i) = \int \psi (\vi^{\yi}>\vi^k \forall k \neq \yi)
\prod_{j=1}^c \mathcal{N}(\vi^j|f_{i}^j,1)d \vb_i,
\end{equation}
where $\vb_i=[\vi^1,...,\vi^c]^T$ is a vector of auxiliary variables, and
$\psi$ is the indicator function whose value is one if the argument is
true, and zero otherwise.
Gibbs sampling can then be employed by drawing samples alternately for
all $i$ from $p(\vb_i|\f_i,y_i)$, which are a conic truncated
multivariate Gaussian distributions, and from $p(\f|\vb,\theta)$ which
is a multivariate Gaussian distribution. Given $\vb$ and $\f$, the
hyperparameters can be drawn, for example, using HMC.

\subsection{Factorized variational approximation}

A computationally convenient variational Bayesian (VB) approximation
for $p(\f|\dataset,\theta)$ can be formed by employing the auxiliary variable
representation (\ref{multinomialprobit_aux}) of the multinomial probit
likelihood.
As shown by \citet{girolami2006}, assuming $\f$ a posteriori independent of
$\vb$ leads to the following approximation
\begin{equation}
q_{\mathrm{VB}}(\vb,\f|\dataset,\theta)=q(\vb)q(\f)=\prod_{i=1}^n q(\vb_i)\prod_{k=1}^c q(\f^k),
\end{equation}
where the latent values associated with the $k$'th class, $\f^k$, are
independent.
The posterior approximation $q(\f^k)$ will be a multivariate Gaussian,
and $q(\vb_i)$ a conic truncated multivariate Gaussian
\citep{girolami2006}.
Given the hyperparameters, the parameters of $q(\vb)$ and $q(\f)$ can
be determined iteratively by maximizing a variational lower bound on
the marginal likelihood.
Each iteration step requires determining the expectations of $\vb_i$
with respect to $q(\vb_i)$ which can be obtained by either one
dimensional numerical quadratures or sampling methods. In our
implementation, the hyperparameters $\theta$ are determined by
maximizing the variational lower bound with fixed $q(\vb)$ and $q(\f)$
similarly as in the maximization step of the EM algorithm.

\section{Experiments}

This section is divided into four parts. In Section 5.1 we illustrate the
properties of the proposed nested EP and IEP approximations in a simple
synthetic classification problem. In Section 5.2, we compare the quality of the
approximate marginal distributions of $\f$, the marginal likelihood approximations and
the predictive class probabilities between EP, IEP, VB and LA in a three class sub-problem
extracted from the USPS digit data. 
In section 5.3, we discuss the computational complexities of all the previously considered
approximate methods, and in Section 5.4, we evaluate them in terms of predictive
performance with estimation of the hyperparameters using several datasets. 
In section 5.3 we show that nested IEP results in almost indistinguishable marginal
likelihood and predictive density estimates compared to the quadrature-based IEP
approximation obtained with the implementation of \citet{seeger2006}. Therefore, in all
other experiments the nested EP approach is used to construct both full EP and IEP
approximations.

\subsection{Illustrative comparison of EP and IEP with synthetic data}

\begin{figure*}[!t]
 \centering
  \subfigure[]{\includegraphics[scale=0.38]{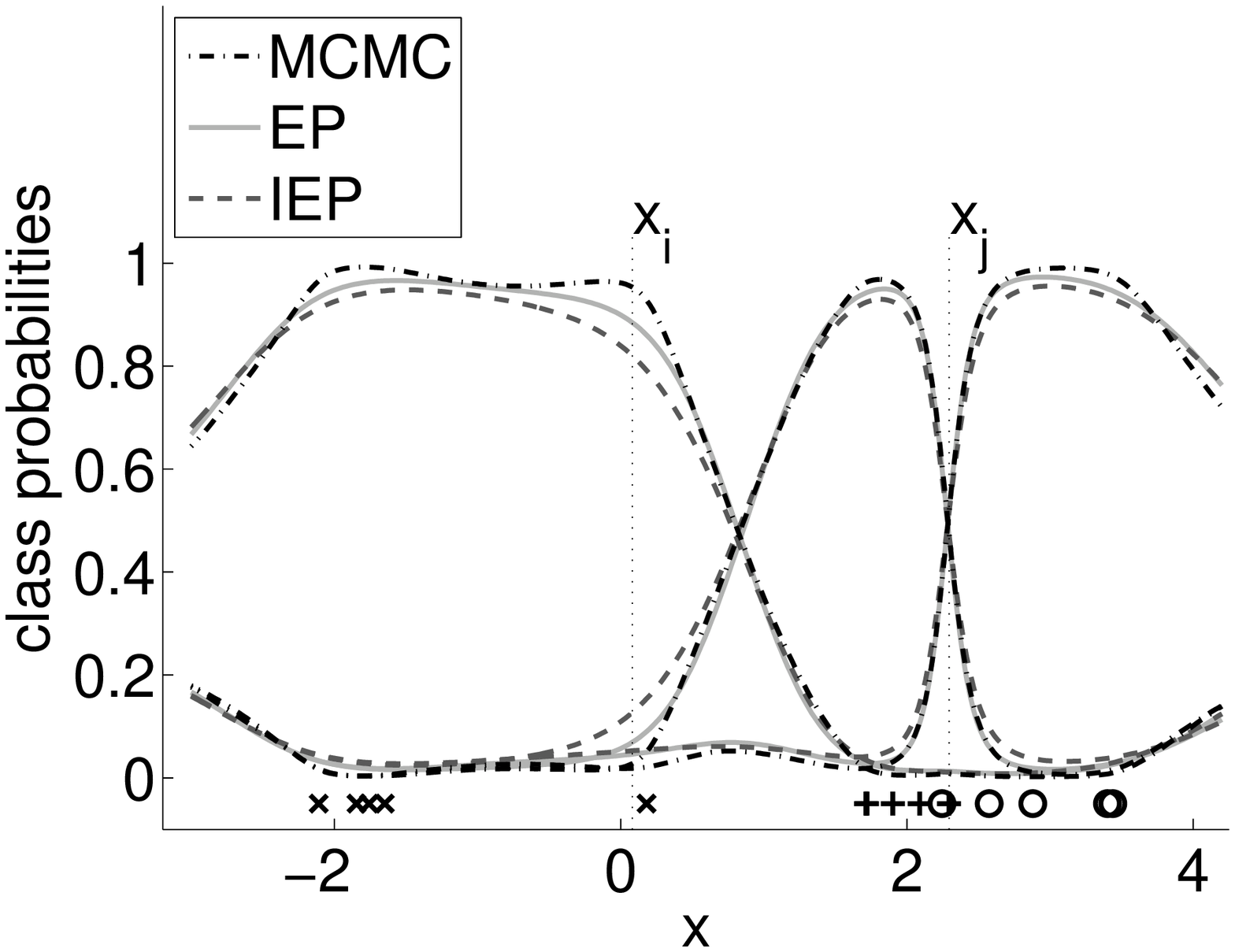}}
  \subfigure[]{\includegraphics[scale=0.38]{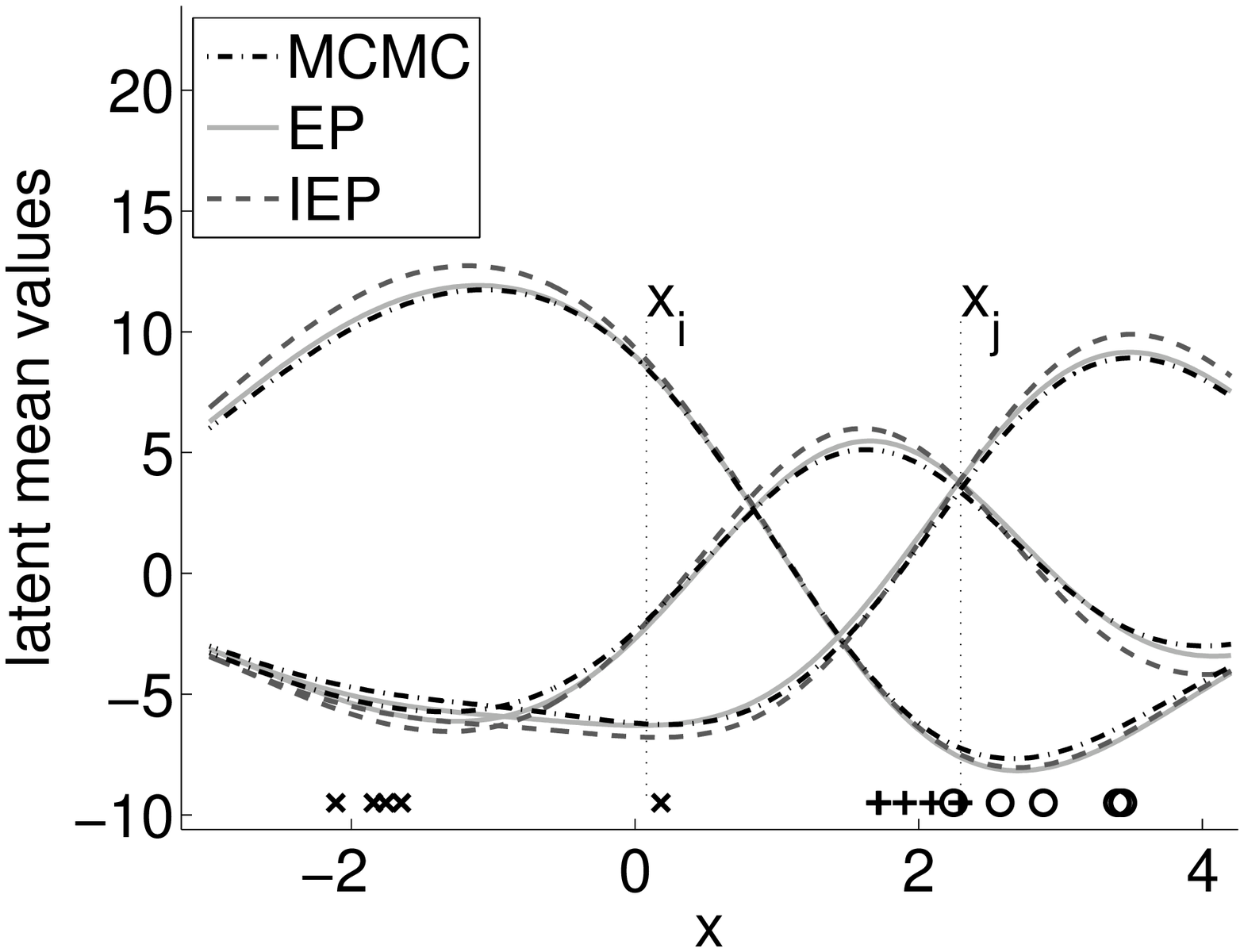}}
  \caption{A synthetic one-dimensional example of a three class
    classification problem, where MCMC, EP and IEP approximations are
    compared. The symbols x (class 1), + (class 2), and o (class 3) in
    the bottom of plots indicate the positions of $n=15$ observations.
    Plot (a) shows the predicted class probabilities, and (b) shows
    the predicted latent mean values for all three classes. The
    symbols $x_i$ and $x_j$ indicate two example positions, where the
    marginal distributions between the latent function values are
    illustrated in Figures \ref{figure_toy_latents_xi} and
    \ref{figure_toy_latents_xj}. See the text for explanation.}
  \label{figure_toy_example}
\end{figure*}

Figure \ref{figure_toy_example} shows a synthetic three-class
classification problem with scalar inputs. The symbols x (class 1), +
(class 2), and o (class 3) indicate the positions of $n=15$ training
inputs generated from three normal distributions with means -1, 2, and
3, and standard deviations 1, 0.5, and 0.5, respectively. The
left-most observations from class 1 can be better separated from the
others but the observations from classes 2 and 3 overlap more in the
input space. We fixed the hyperparameters of the squared exponential
covariance function at the corresponding MCMC means:
$\log(\sigma^2)=4.62$ and $\log(l)=0.26$.

Figure \ref{figure_toy_example}(a) shows the predictive probabilities
of all the tree classes estimated with EP, IEP and MCMC as a function
of the input $x$.
At the class boundaries, the methods give similar predictions but
elsewhere
MCMC is the most confident while IEP seems more cautious. 
The performance of EP is somewhere between MCMC and IEP, although the
differences are small.
To explain why the predictions differ, we look at the qualities of the
approximations made for the underlying $\f$.
Figure \ref{figure_toy_example}(b) shows the approximated latent mean
values which are similar at all input locations.

\begin{figure*}[!ht]
  \centering
    \subfigure[]{\includegraphics[scale=0.25]{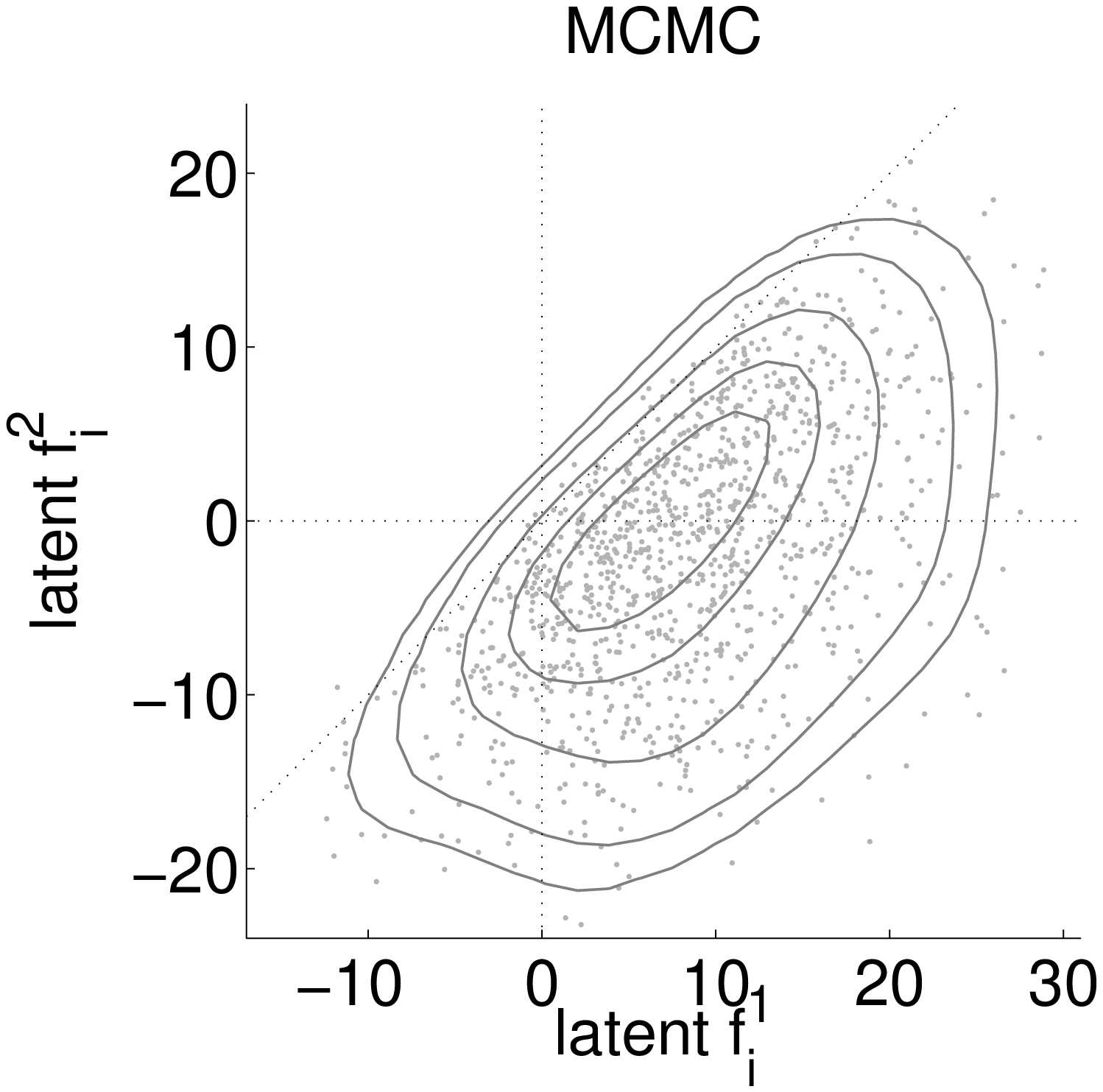}}
    \subfigure[]{\includegraphics[scale=0.25]{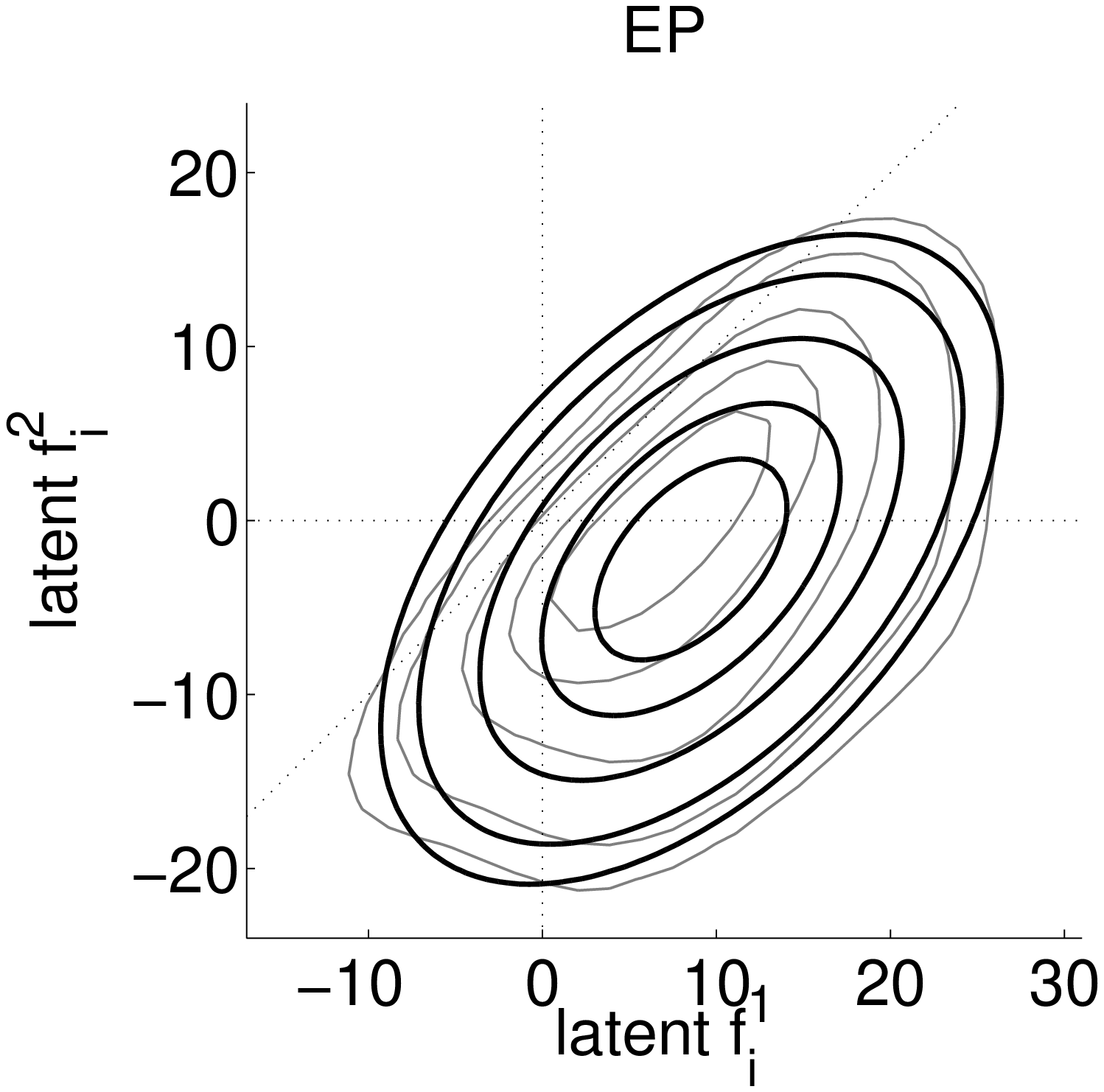}}
    \subfigure[]{\includegraphics[scale=0.25]{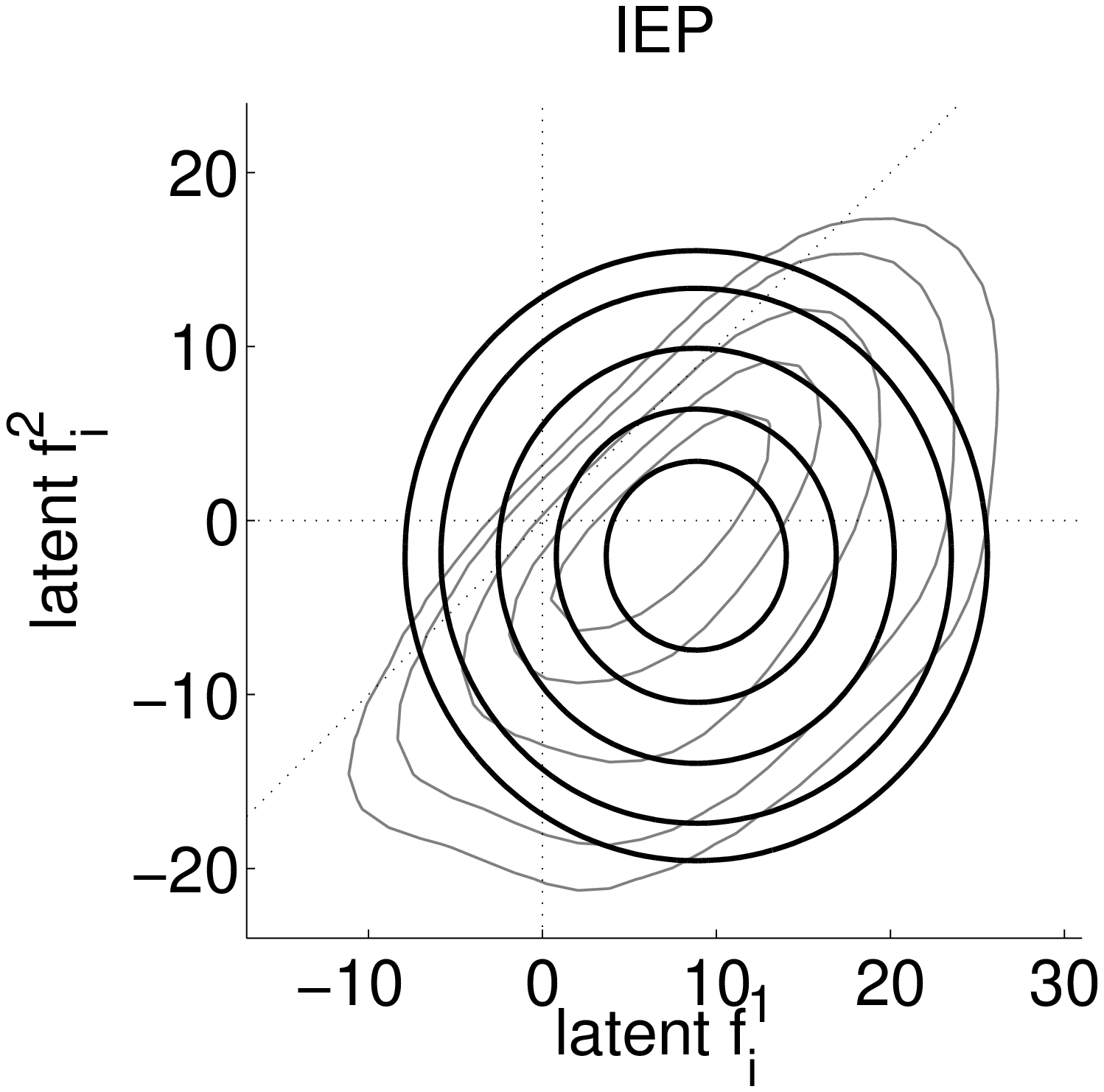}}
  \centering
    \subfigure[]{\includegraphics[scale=0.23]{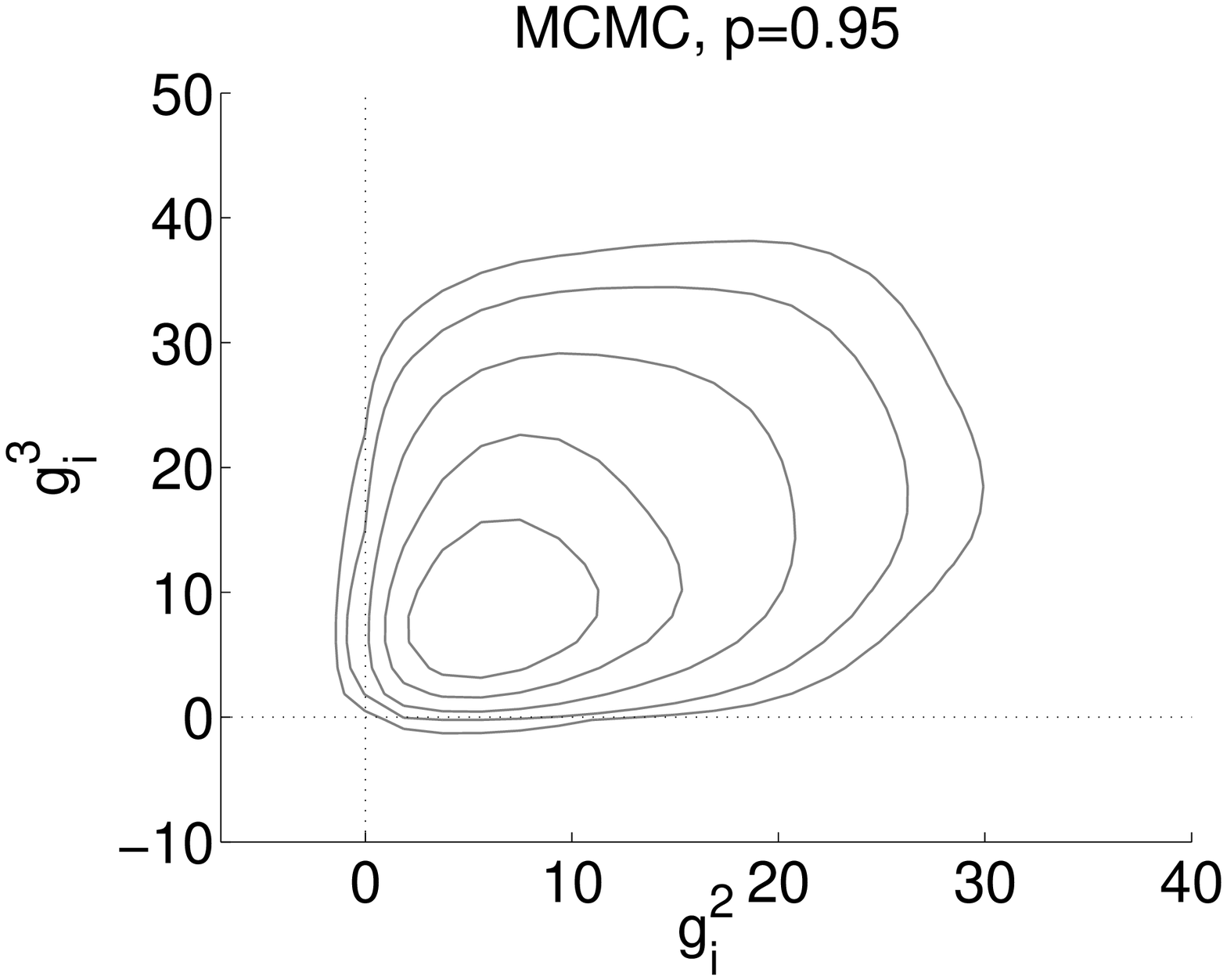}}
    \subfigure[]{\includegraphics[scale=0.23]{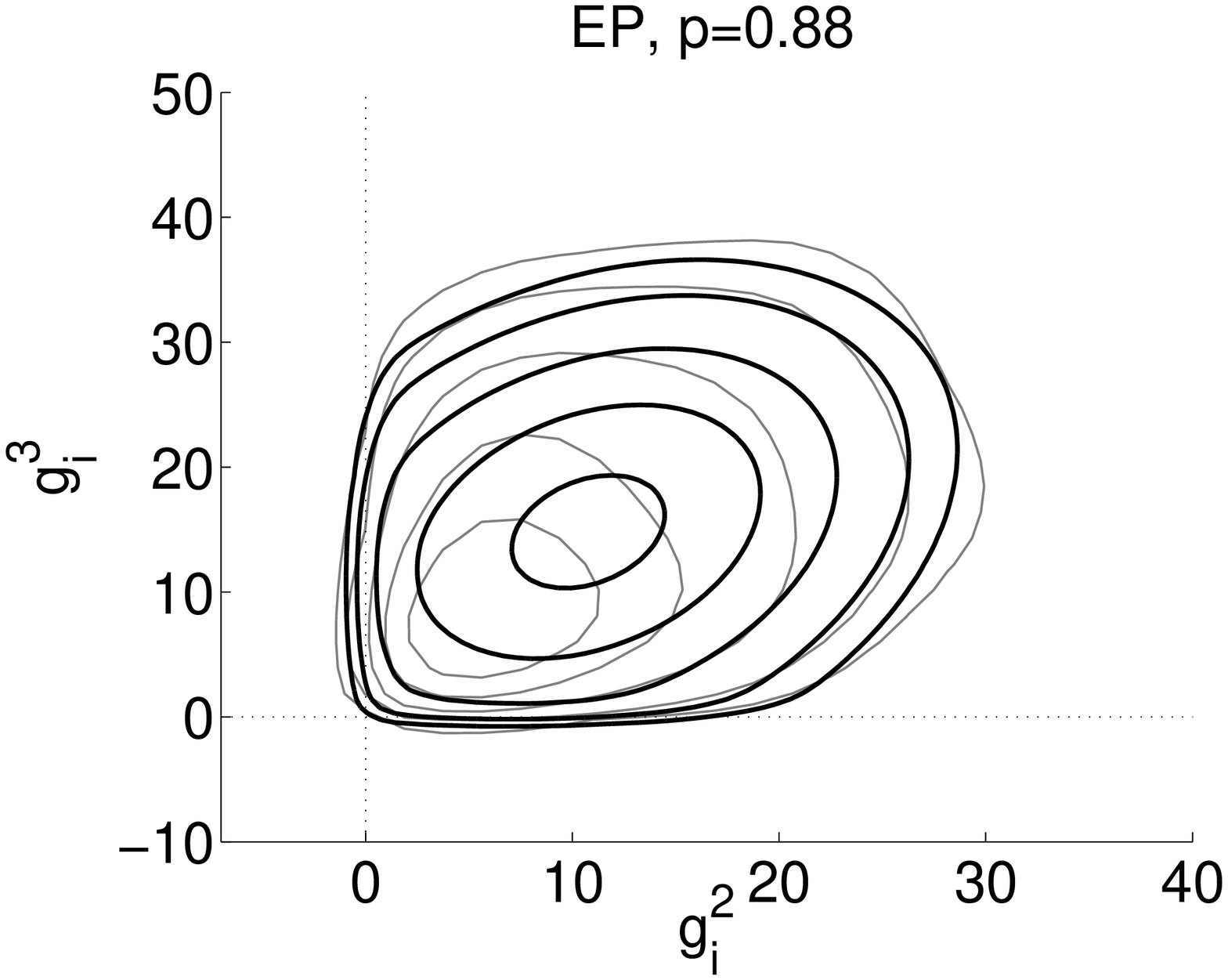}}
    \subfigure[]{\includegraphics[scale=0.23]{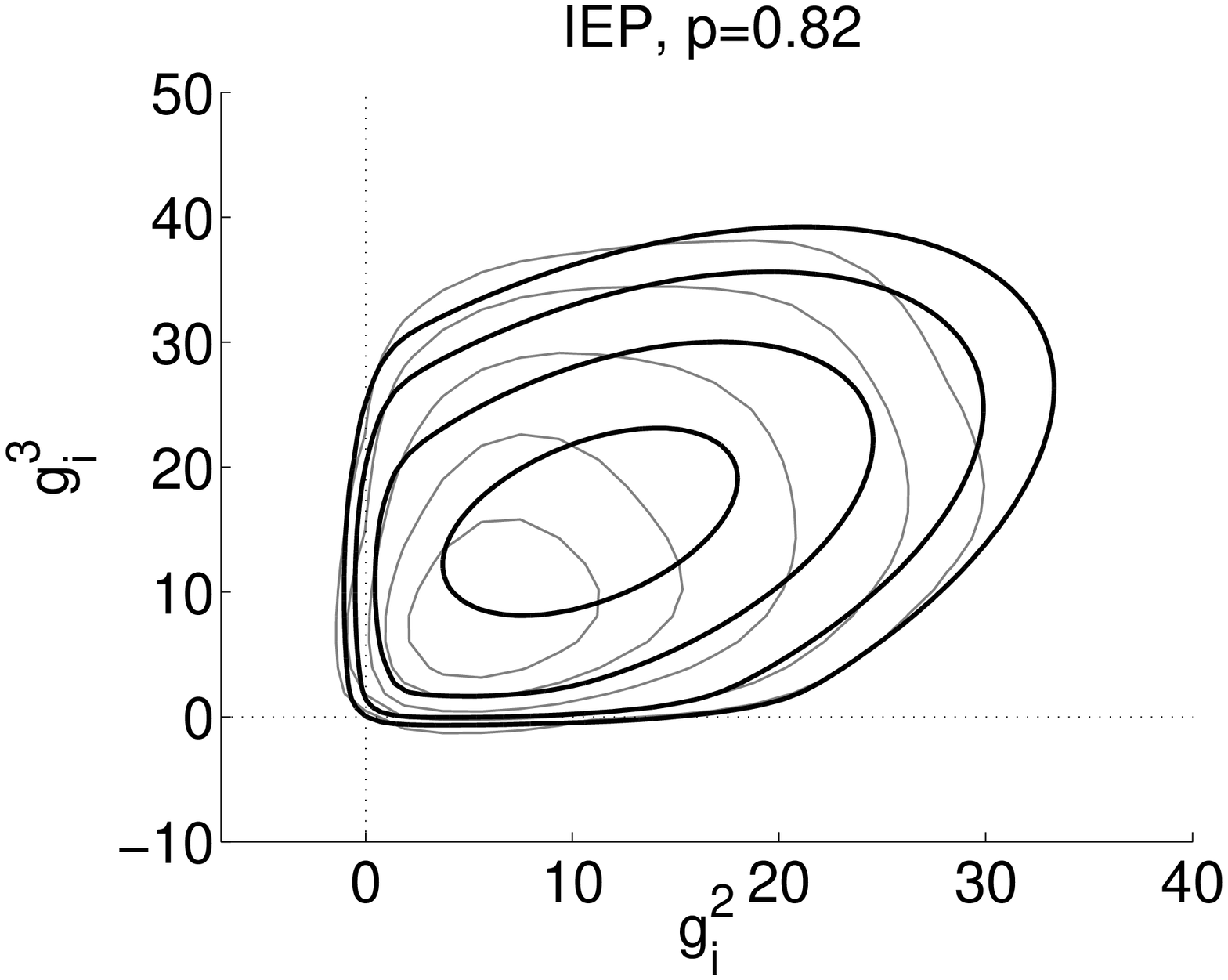}}
  \caption{An example of a non-Gaussian marginal posterior
    distribution for the latent values at the input $x_i$ in the
    synthetic example shown in Figure \ref{figure_toy_example}. The
    first row shows the distribution for the latents $f_i^1$ and
    $f_i^2$. Plot (a) shows a scatter-plot of MCMC samples drawn from
    the posterior and the estimated density contour levels which
    correspond to the areas that include approximately 95\%, 90\%
    75\%, 50\%, and 25\% of the probability mass. Plots (b) and (c)
    show the equivalent contour levels of EP and IEP approximations
    (bold black lines) and the contour plot of MCMC approximation
    (gray lines) for comparison. Plots (d)-(f) show contours of
    $\hat{p}(\bm{\mathrm{g}}_i|\dataset,x_i)$ for $g_i^2$ and $g_i^3$.  The
    probability for class 1 is obtained by calculating the integral
    over $\bm{\mathrm{g}}_i$, which results in approximately 0.95 for
    MCMC, 0.88 for EP, and 0.82 for IEP. See the text for
    explanation.}
  \label{figure_toy_latents_xi}
\end{figure*}

To illustrate the approximate posterior uncertainties of $\f$, we
visualize two exemplary marginal distributions at locations $x_i$ and
$x_j$ marked in Figure \ref{figure_toy_example}. The MCMC samples of
$f_i^1$ and $f_i^2$ (the latents associated with classes 1 and 2
related to $x_i$) together with a smoothed density estimate are shown
in Figure \ref{figure_toy_latents_xi}(a).
The marginal distribution is non-Gaussian, and the latent values are
more likely larger for class 1 than for class 2 indicating a larger
predictive probability for class 1. The corresponding EP and IEP
approximations are shown in Figures
\ref{figure_toy_latents_xi}(b)-(c). EP captures the shape of the true
marginal posterior distribution better than IEP. To illustrate the
effect of these differences on the predictive probabilities, we show
the unnormalized tilted distributions
\begin{equation} \label{eq_tilted_g}
  \hat{p}(\bm{\mathrm{g}}_i|\dataset,x_i) = q( \bm{\mathrm{g}}_i|\dataset,\x_i) 
  \prod_{k=1, k \neq y_i}^c \Phi(g_i^k),
\end{equation}
where the random vector $\bm{\mathrm{g}}_i$ is formed from the
transformed latents $g_i^k=\w_i^T \z_{i,j}$ for $k \neq \yi$, and
$q( \bm{\mathrm{g}}_i|\dataset,\x_i)$ is the approximate marginal obtained
from $q(\f_i|\dataset,x_i)$ by a linear transformation.
Note that the marginal predictive probability for class label $y_i$
with the multinomial probit model \eqref{lik_multinomialprobit} can be
obtained by calculating the integral over $\bm{\mathrm{g}}_i$ in
\eqref{eq_tilted_g}.
Figures \ref{figure_toy_latents_xi}(d)-(f) show the contours of the
different approximations of $\hat{p}(\bm{\mathrm{g}}_i|\dataset,x_i)$ for
$k\in\{2,3\}$, which for MCMC are obtained using a smoothed estimate
of $q(\bm{\mathrm{g}}_i|\dataset,x_i)$. The distributions are heavily
truncated by the probit factors elsewhere than the upper-right
quadrant.
Compared to the MCMC estimate, IEP places more probability mass to the
other quadrants, and therefore underestimates the predictive
probability for class 1 more than EP. The approximate predictive
probabilities are 0.95 for MCMC, 0.88 for EP, and 0.82 for IEP.

\begin{figure*}[!t]
 \centering
   \subfigure[]{\includegraphics[scale=0.25]{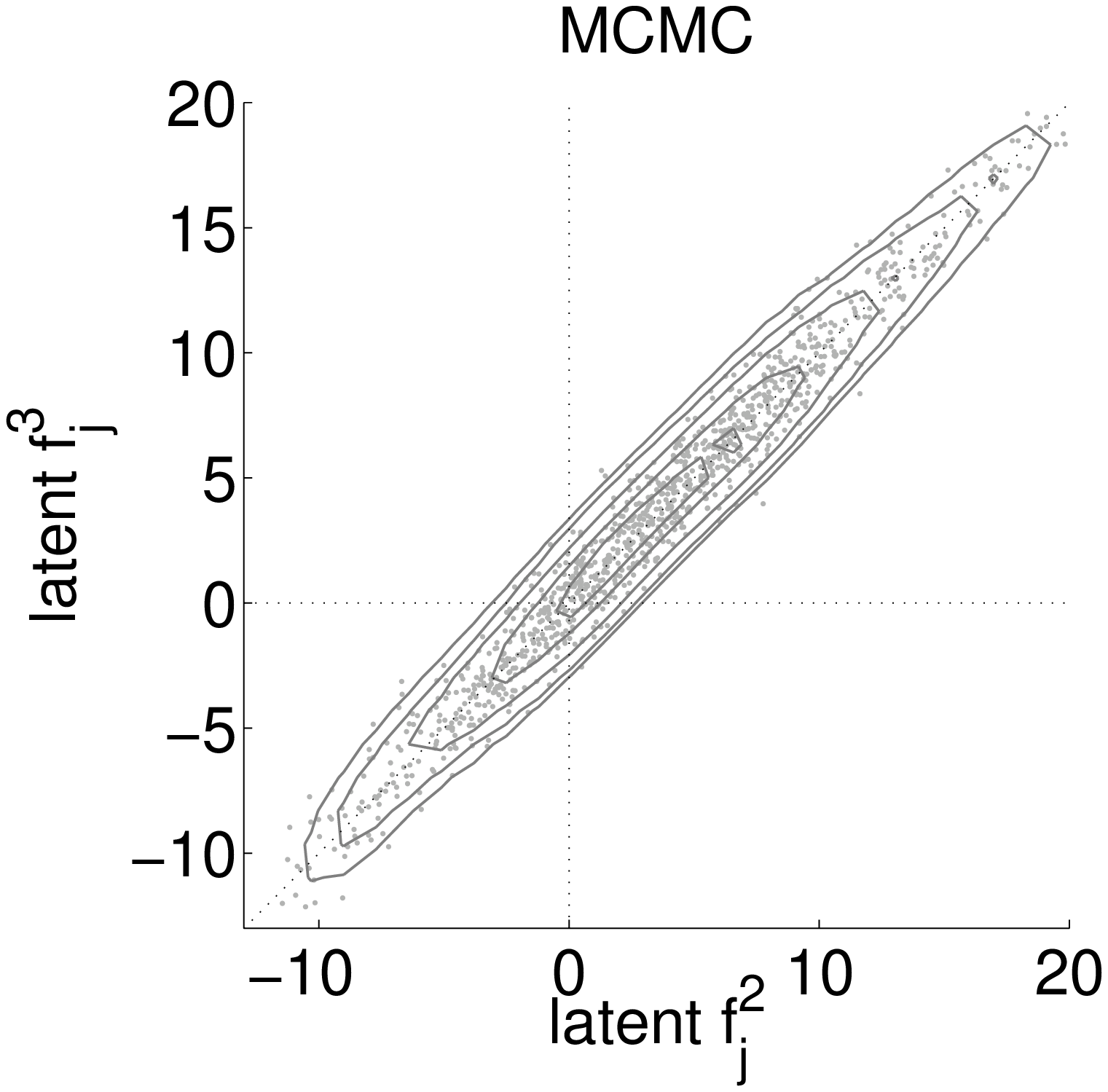}}
   \subfigure[]{\includegraphics[scale=0.25]{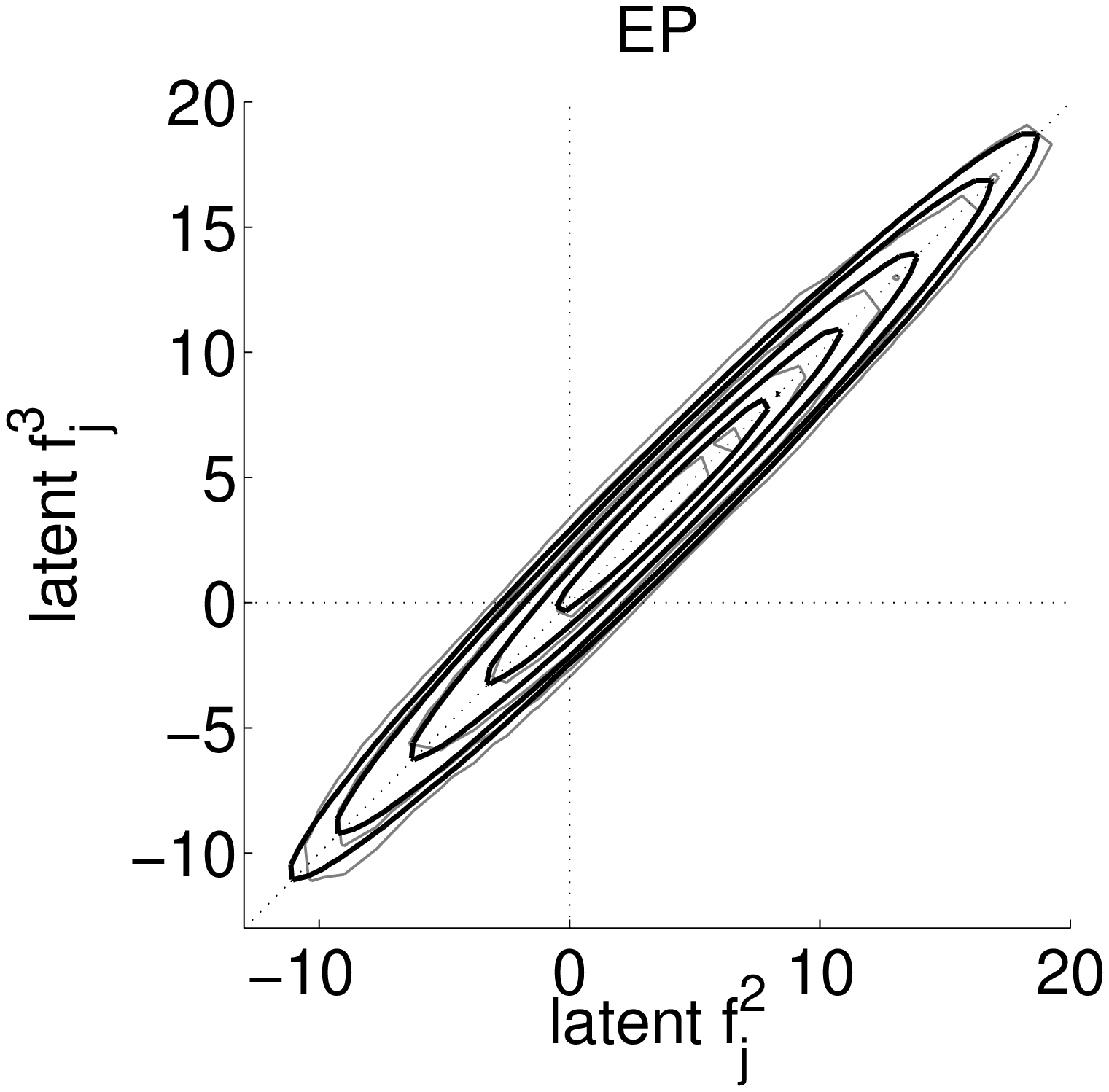}}
   \subfigure[]{\includegraphics[scale=0.25]{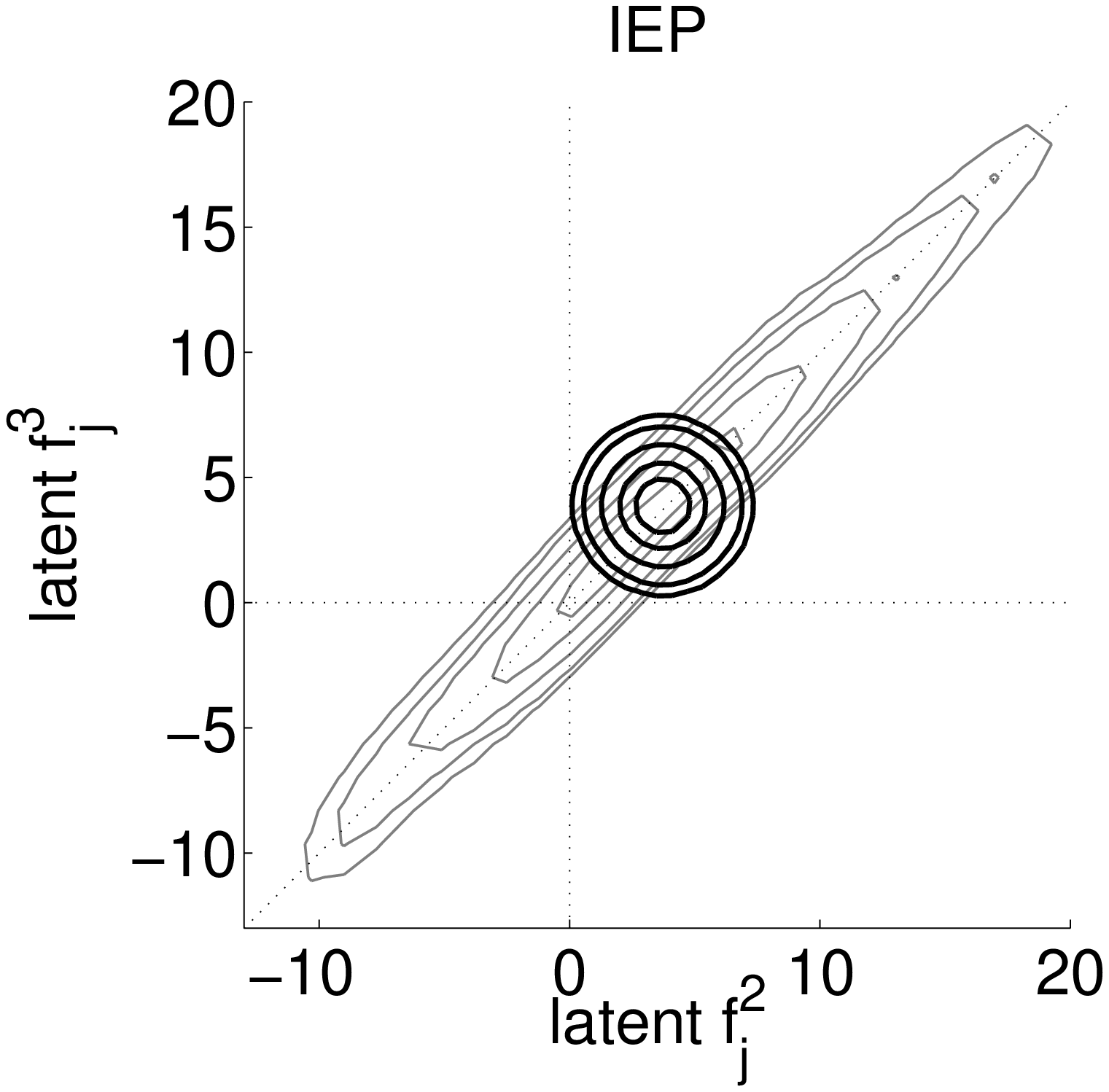}}
 \centering
   \subfigure[]{\includegraphics[scale=0.23]{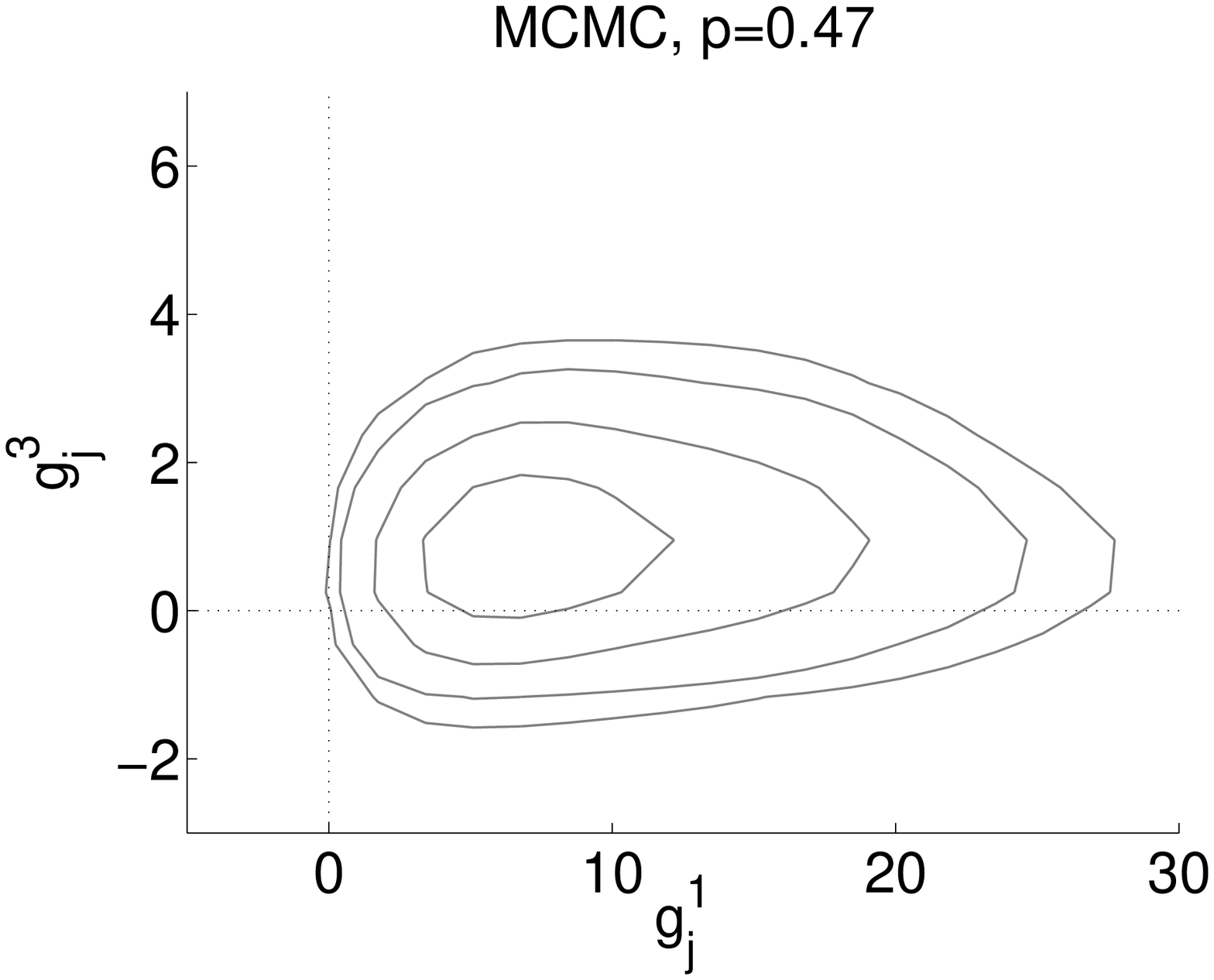}}
   \subfigure[]{\includegraphics[scale=0.23]{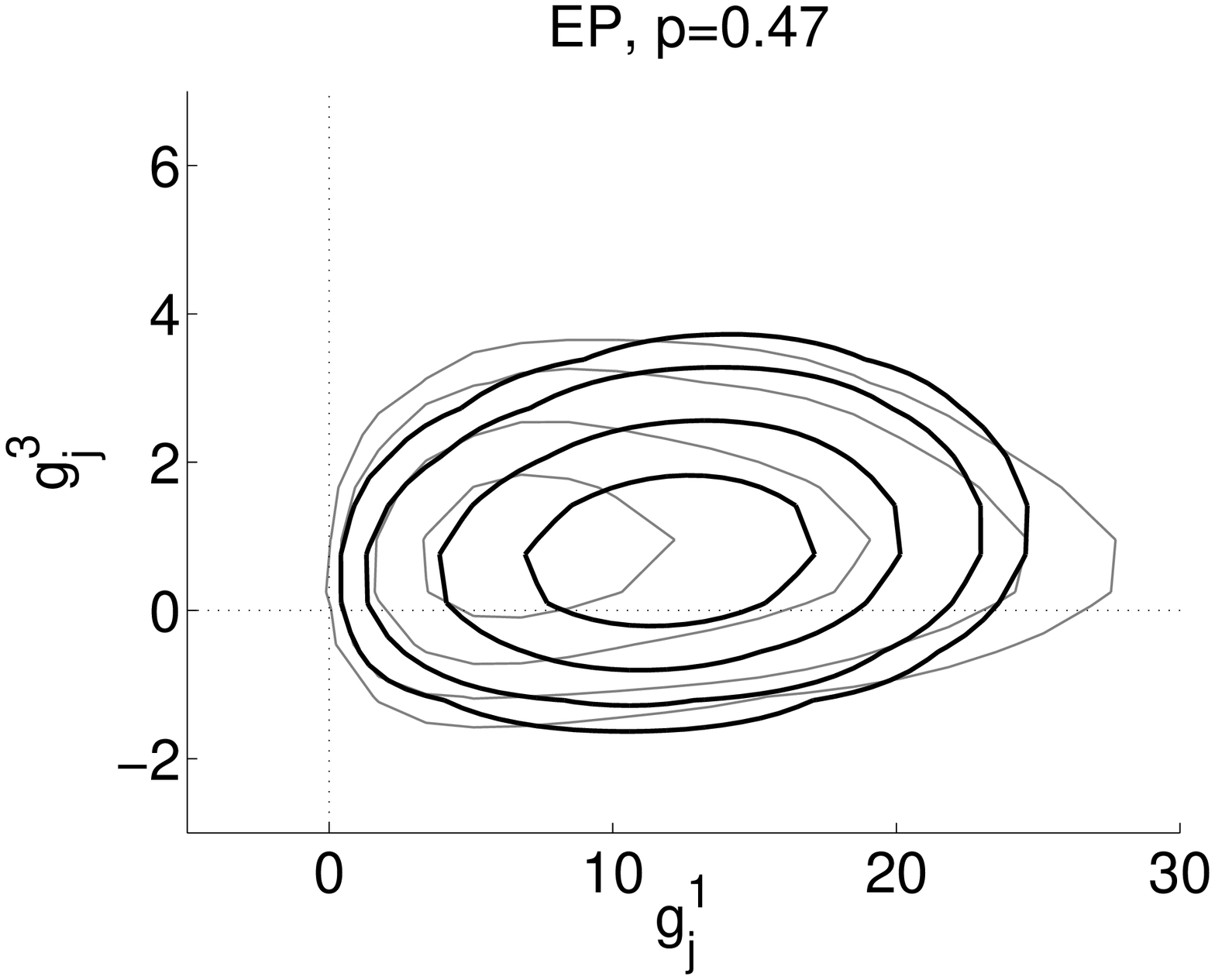}}
   \subfigure[]{\includegraphics[scale=0.23]{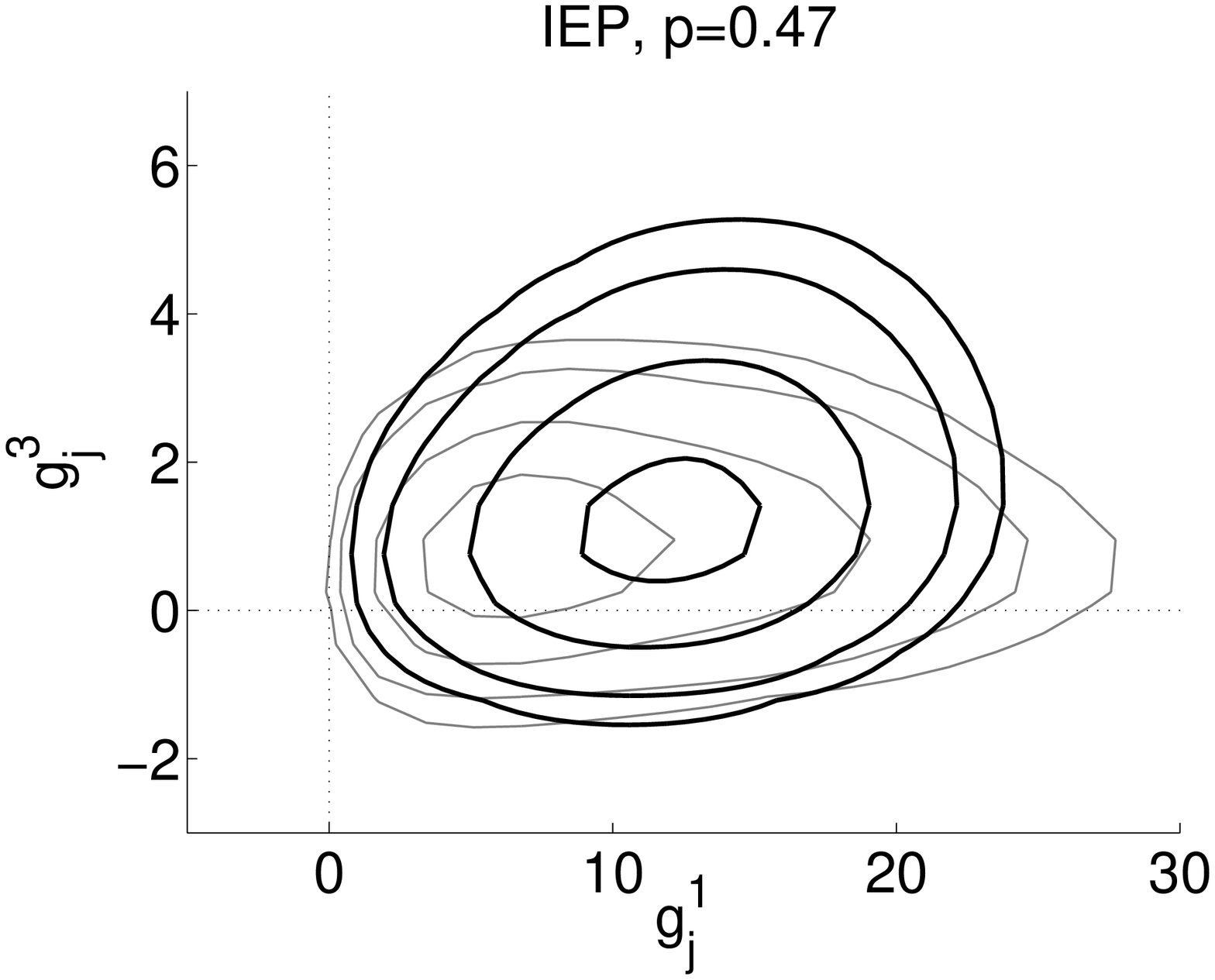}}
   \caption{An example of a close-to-Gaussian but non-isotropic
     marginal posterior distribution for the latent values at the
     input $x_j$ in the synthetic example shown in Figure
     \ref{figure_toy_example}. The first row shows the distribution
     for the latents $f_j^2$ and $f_j^3$. Plot (a) shows a scatter-plot
     of MCMC samples drawn from the posterior and the estimated
     density contour levels which correspond to the areas that include
     approximately 95\%, 90\% 75\%, 50\%, and 25\% of the probability
     mass. Plots (b) and (c) show the equivalent contour levels of EP
     and IEP approximations (bold black lines) and the contour plot of
     MCMC approximation (gray lines) for comparison. Plots (d)-(f)
     show contours of $\hat{p}(\bm{\mathrm{g}}_j|\dataset,x_j)$ for $g_j^1$
     and $g_j^3$.  The probability for class 2 is obtained by
     calculating the integral over $\bm{\mathrm{g}}_j$, which results
     in approximately 0.47 for all the methods. See the text for
     explanation.}
  \label{figure_toy_latents_xj}
\end{figure*}

The location $x_j$ is near the class boundary, where all the methods
give similar predictive probabilities, although the latent
approximations can differ notably as shown in Figures
\ref{figure_toy_latents_xj}(a)-(c),
which visualize the marginal approximations for $f_j^2$ and $f_j^3$.
EP is consistent with the MCMC estimate but due to the independence
constraint IEP seriously underestimates the uncertainty of this
close-to-Gaussian but non-isotropic 
marginal distribution.  Although Figures
\ref{figure_toy_latents_xj}(d)-(f) show that IEP is more inaccurate
than EP, the integral over the tilted distribution of
$\bm{\mathrm{g}}_j$ is in practice the same, since equal amount of
probability mass is distributed on both sides of the diagonal in
Figure \ref{figure_toy_latents_xj}(c). The predictive probabilities
for class 2 is approximately 0.47 for all the methods.

\subsection{Approximate marginal densities with the USPS data}

In this section, we compare the predictive performances and marginal
likelihood approximations of EP, IEP, VB and LA. We define a three
class sub-problem from the US Postal Service (USPS) repartitioned
handwritten digits data by considering classification of 3's vs. 5's
vs. 7's.\footnote{We use the same data partition as discussed by
  \citet{rasmussen2006}.} The data consists of 1157 training points
and 1175 test points with 256 covariates.  We fixed the hyperparameter
values at $\log(\sigma^2)=4$ and $\log(l)=2$ which leads to skewed
non-Gaussian marginal posterior distributions as will be illustrated
shortly.

\begin{figure*}[!t]
  \centering
  \subfigure[]{\includegraphics[scale=0.19]{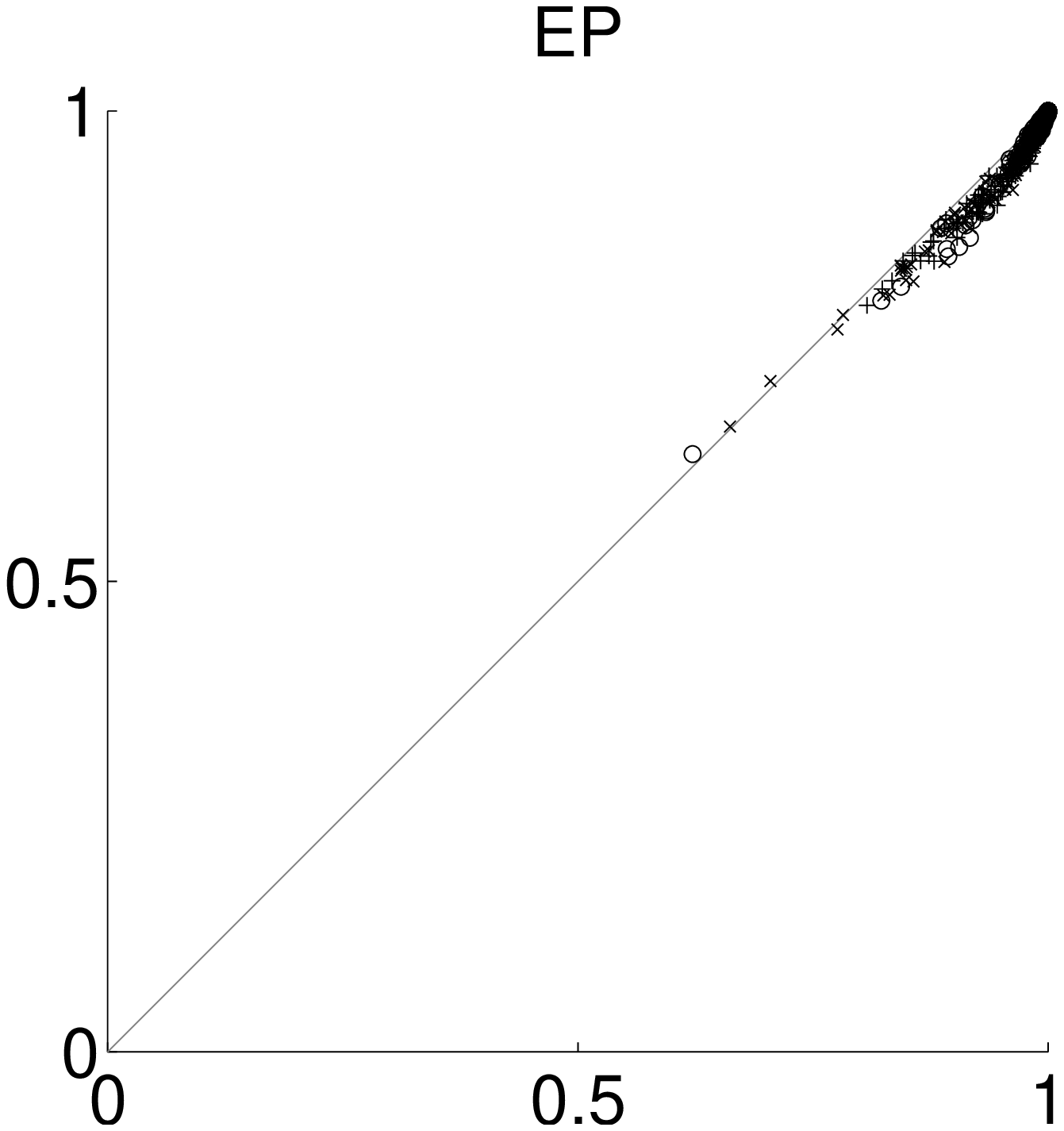}}
  \subfigure[]{\includegraphics[scale=0.19]{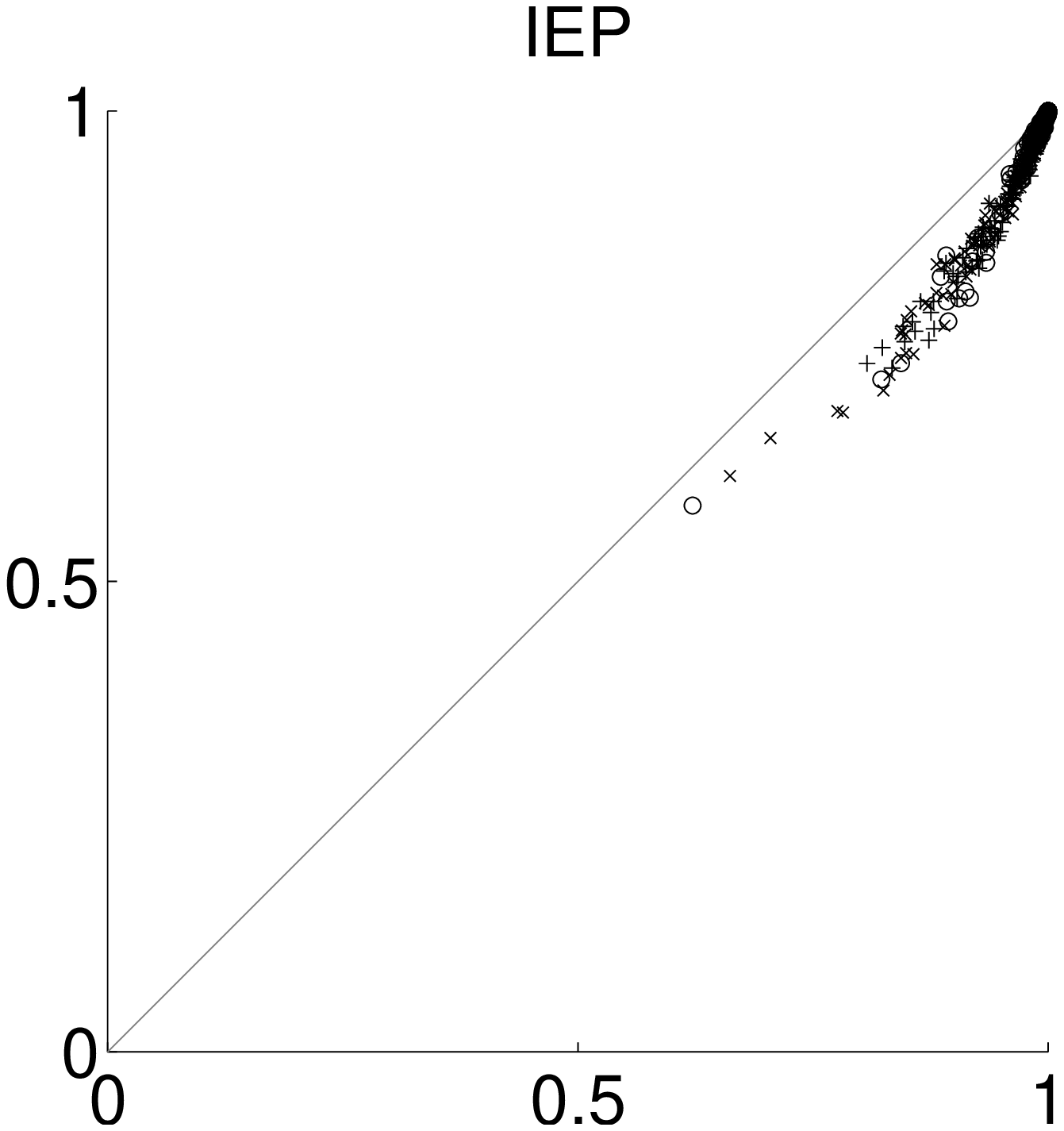}}
  \subfigure[]{\includegraphics[scale=0.19]{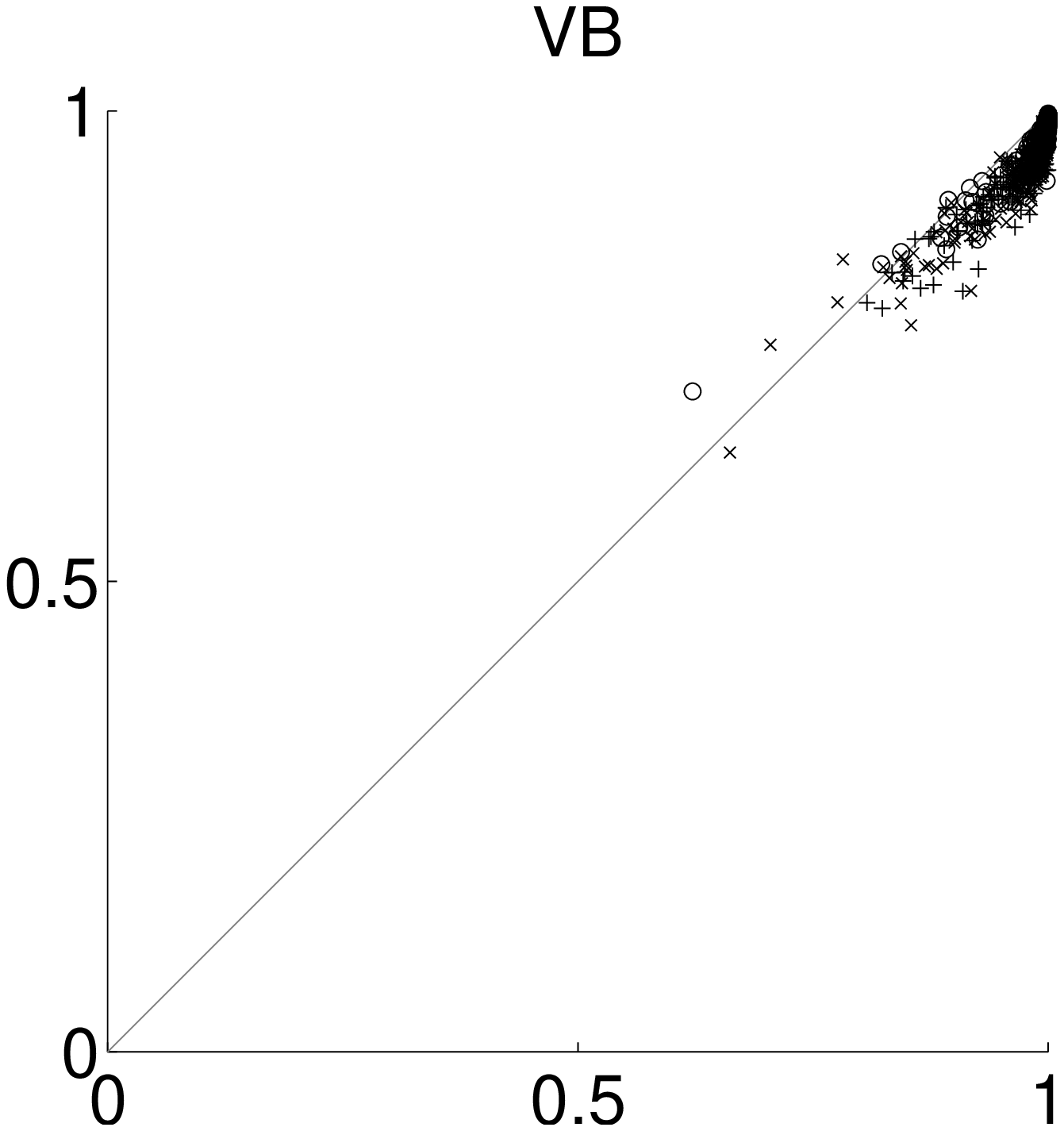}}
  \subfigure[]{\includegraphics[scale=0.19]{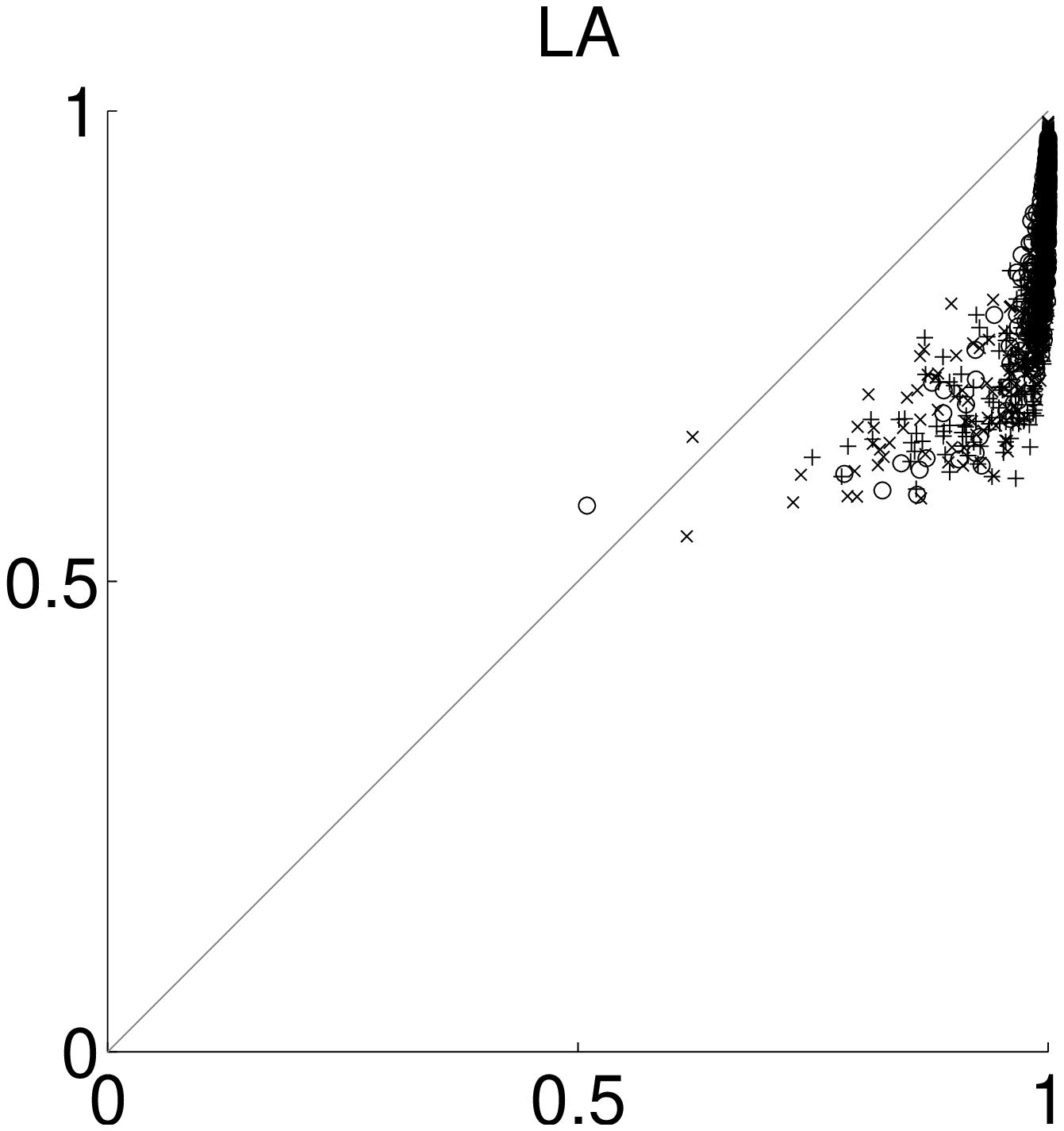}}
  \subfigure[]{\includegraphics[scale=0.19]{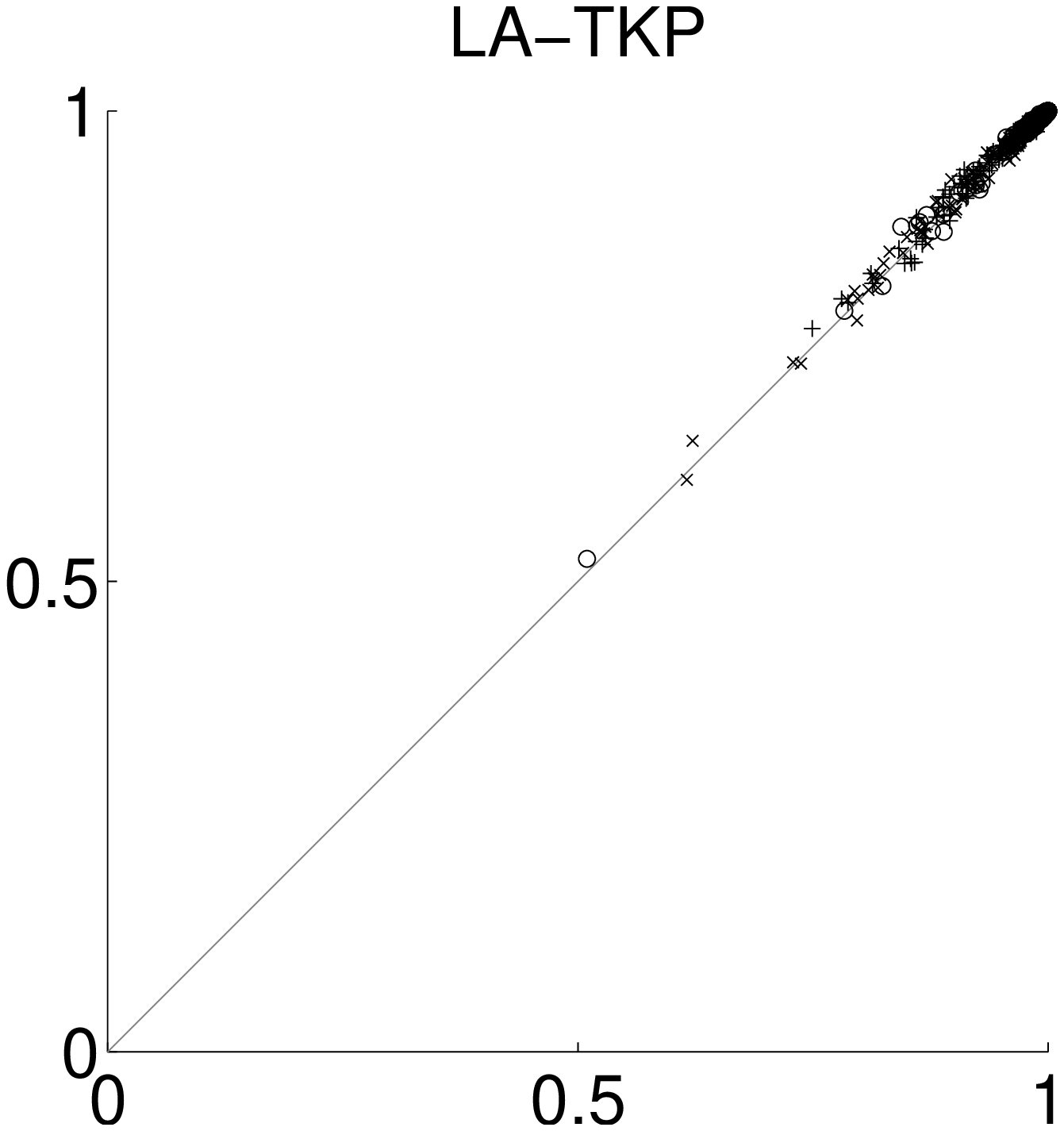}}

  \centering
  \subfigure[]{\includegraphics[scale=0.19]{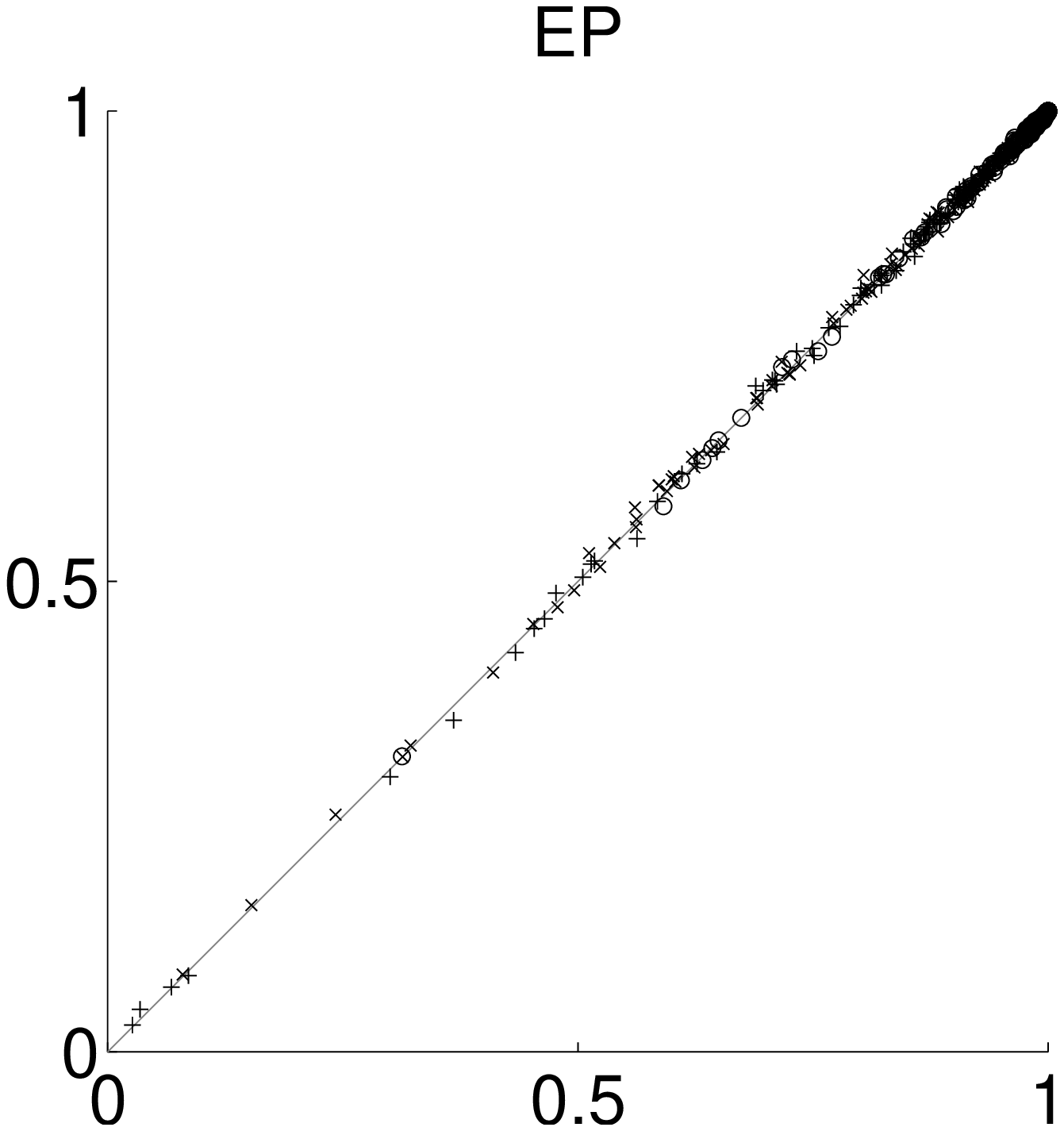}}
  \subfigure[]{\includegraphics[scale=0.19]{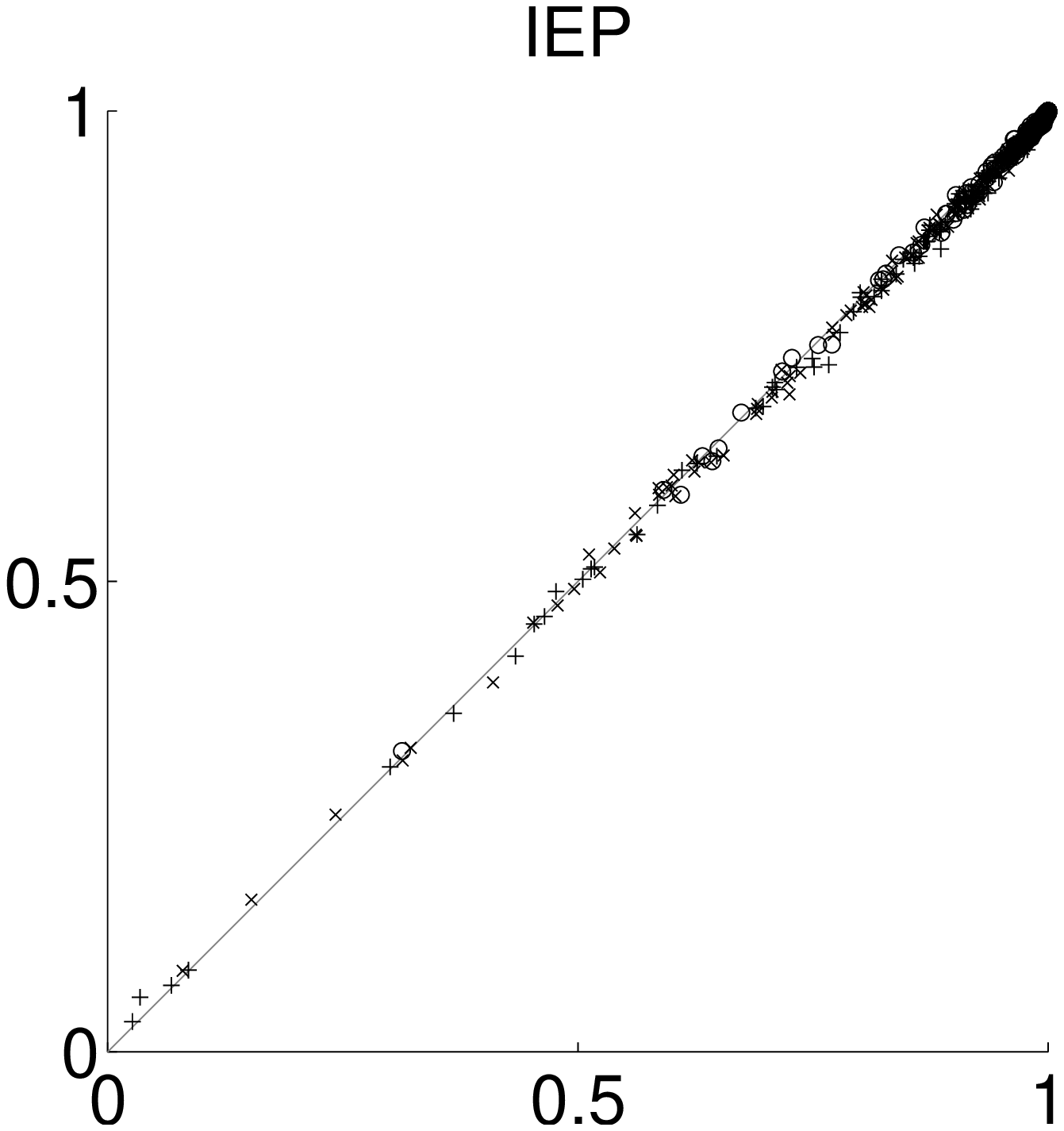}}
  \subfigure[]{\includegraphics[scale=0.19]{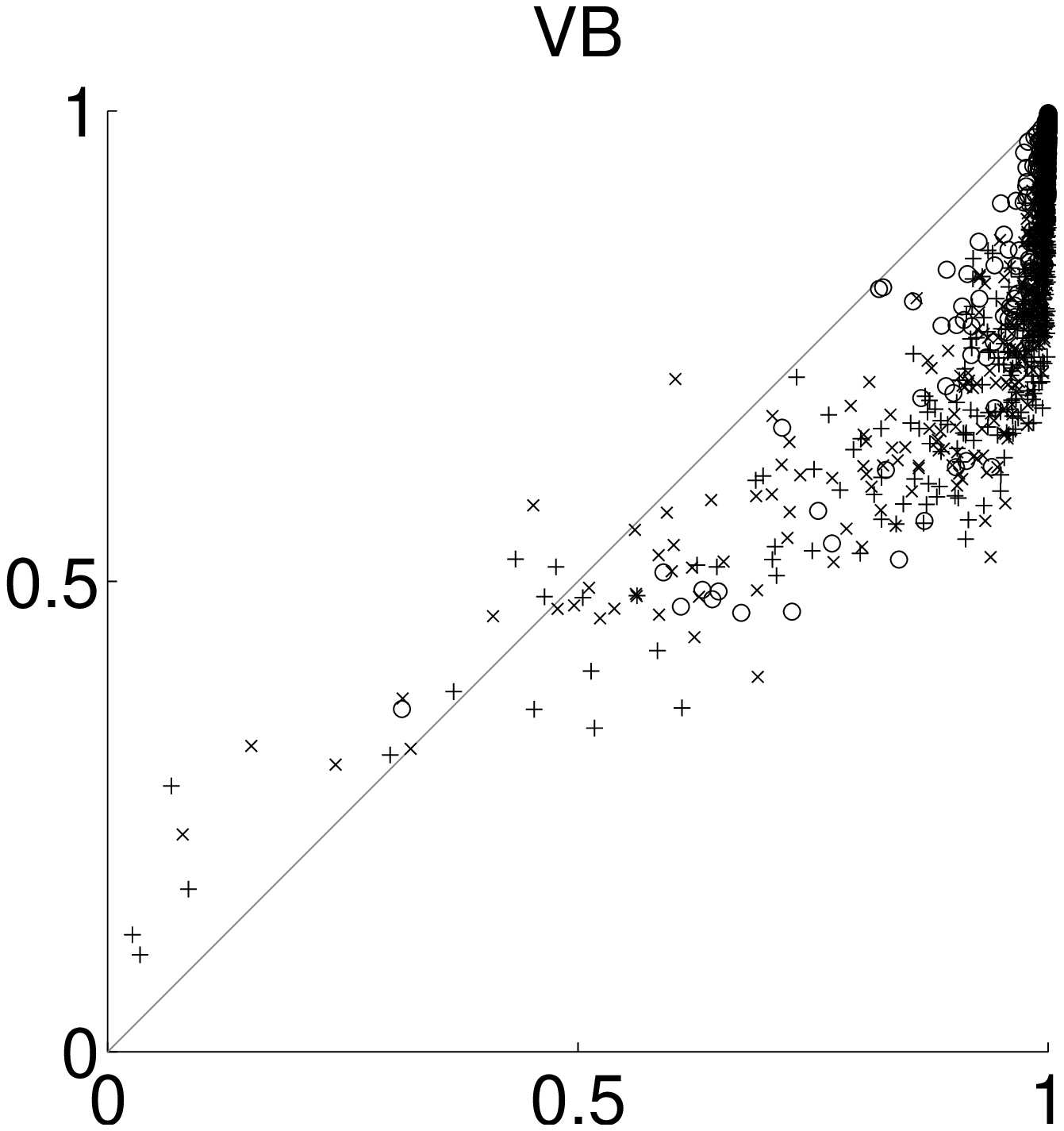}}
  \subfigure[]{\includegraphics[scale=0.19]{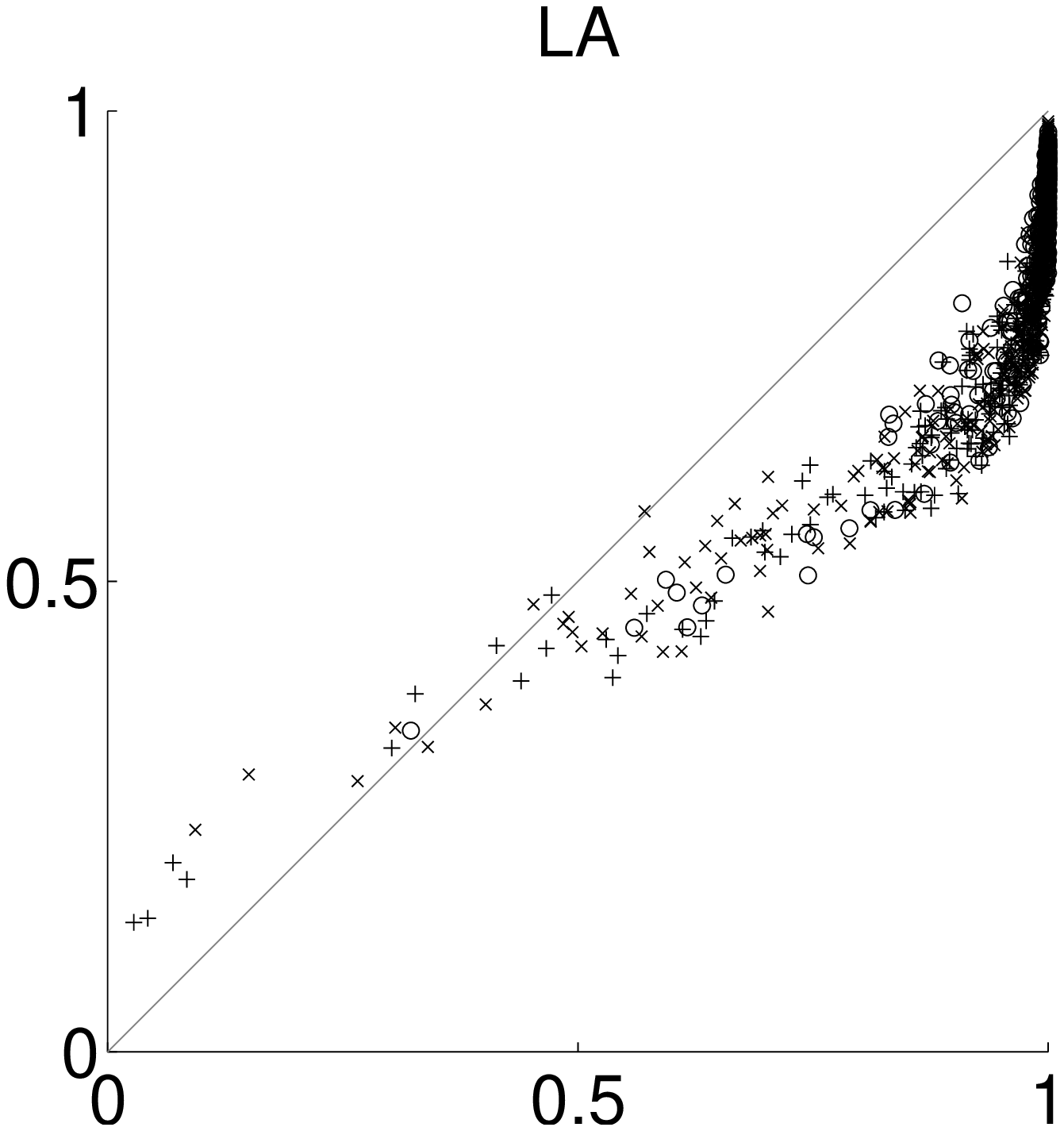}}
  \subfigure[]{\includegraphics[scale=0.19]{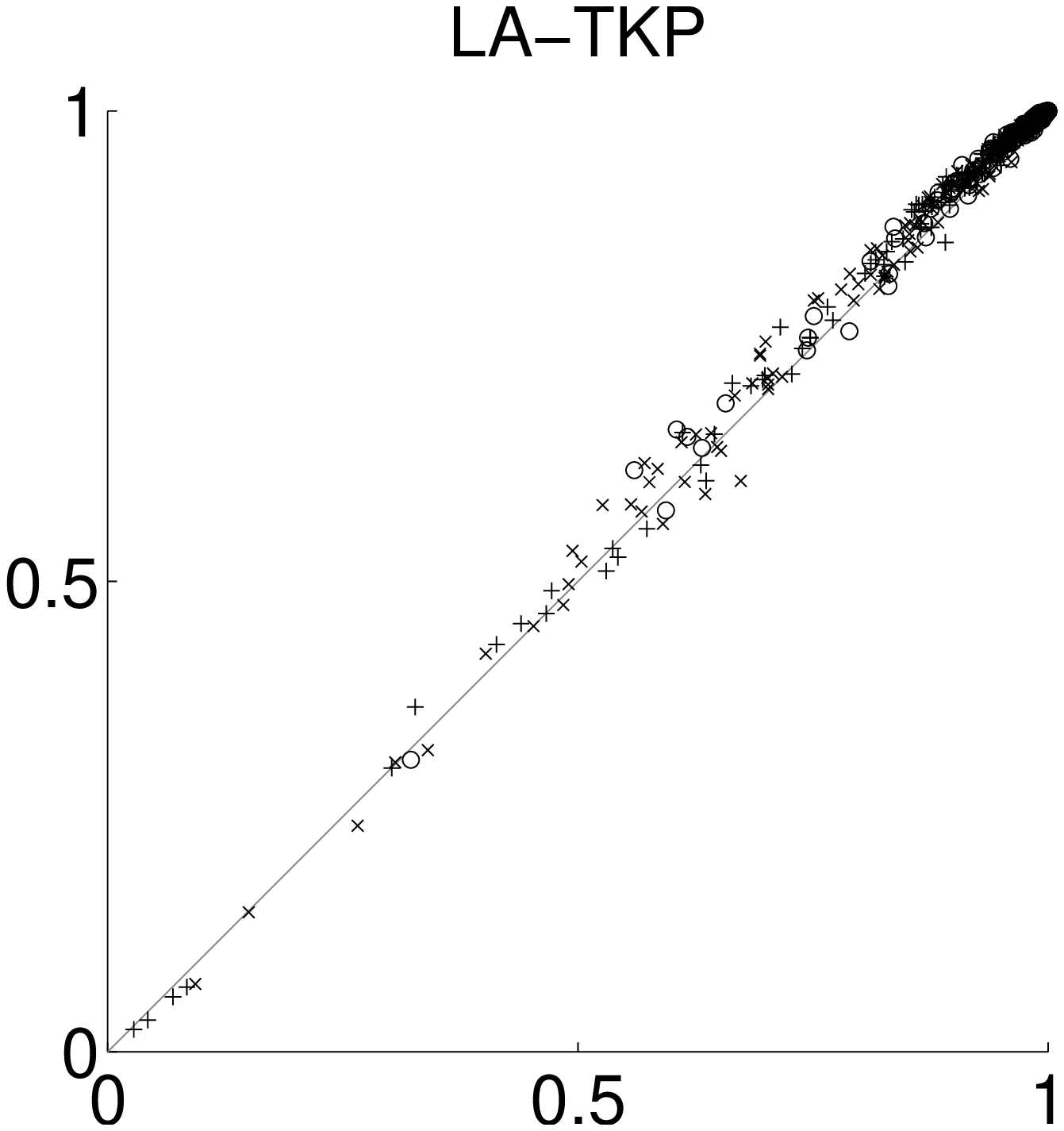}}
  \caption{Class probabilities on the USPS 3 vs.  5 vs. 7 data. The
    MCMC estimates are shown on x-axis and EP, IEP, VB, LA, and
    LA-TKP on y-axis. The first row shows the predictive
    probabilities of the true class labels for training and the second
    row for test points.  The symbols (x, +, o) corresponds to the
    handwritten digit target classes 3, 5, and 7. The hyperparameters
    of the squared exponential covariance function were fixed at
    $\log(\sigma^2)=4$ and $\log(l)=2$.  }
  \label{figure_nongaussian_probabilities}
\end{figure*}

Figure \ref{figure_nongaussian_probabilities} shows the predictive
probabilities of the true class labels for all the approximate methods
plotted against the MCMC estimate. The first row shows the training
and the second row the test cases. Overall, EP gives the most accurate
estimates while IEP slightly underestimates the probabilities for the
training cases but performs well for the test cases. Both VB and LA
underestimate the predictive probabilities for the test cases, but
LA-TKP with the marginal corrections clearly improves the estimates of
the LA approximation.

\begin{figure*}[!t]
 \centering
  \subfigure[]{\includegraphics[scale=0.19]{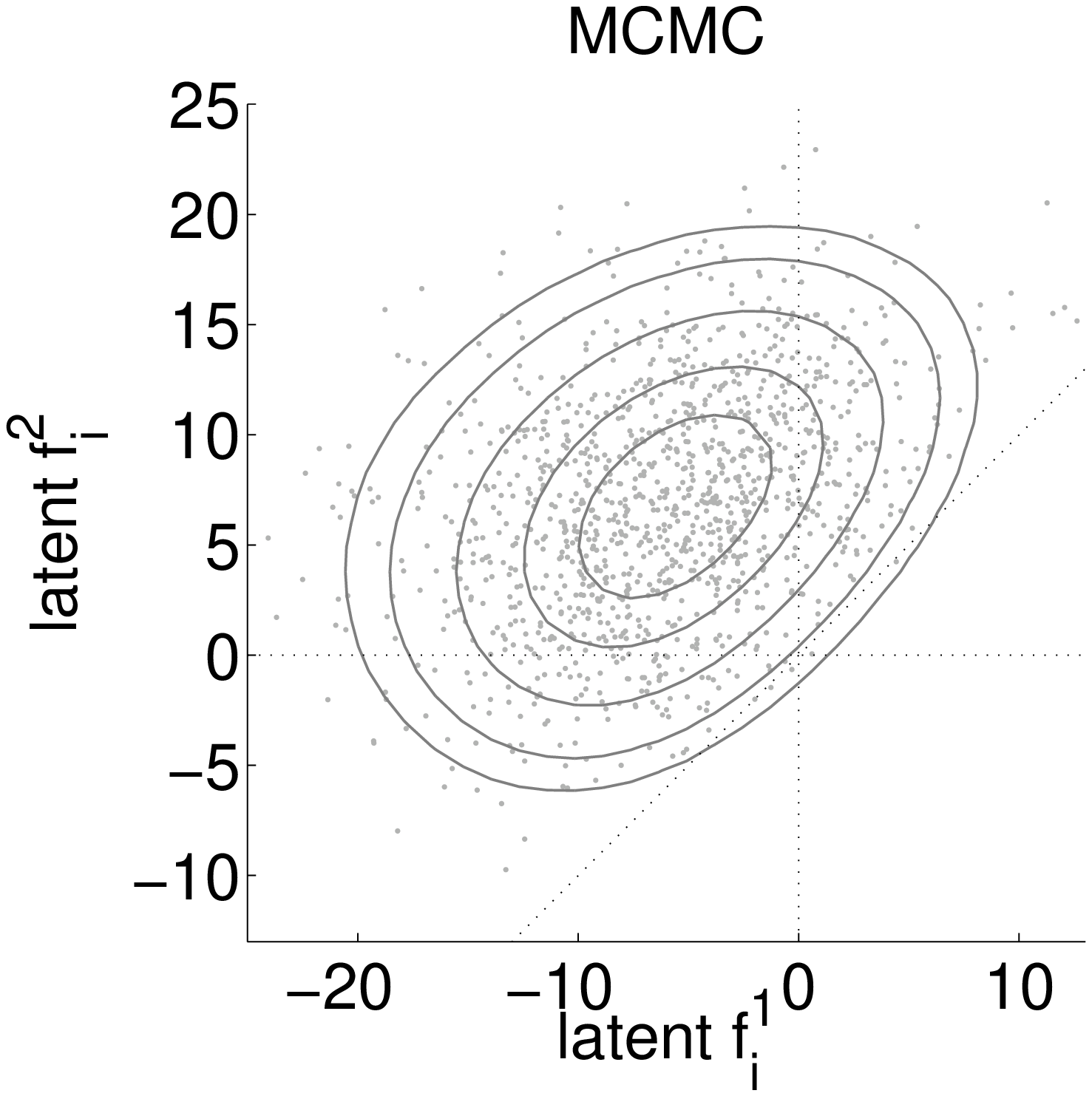}}
  \subfigure[]{\includegraphics[scale=0.19]{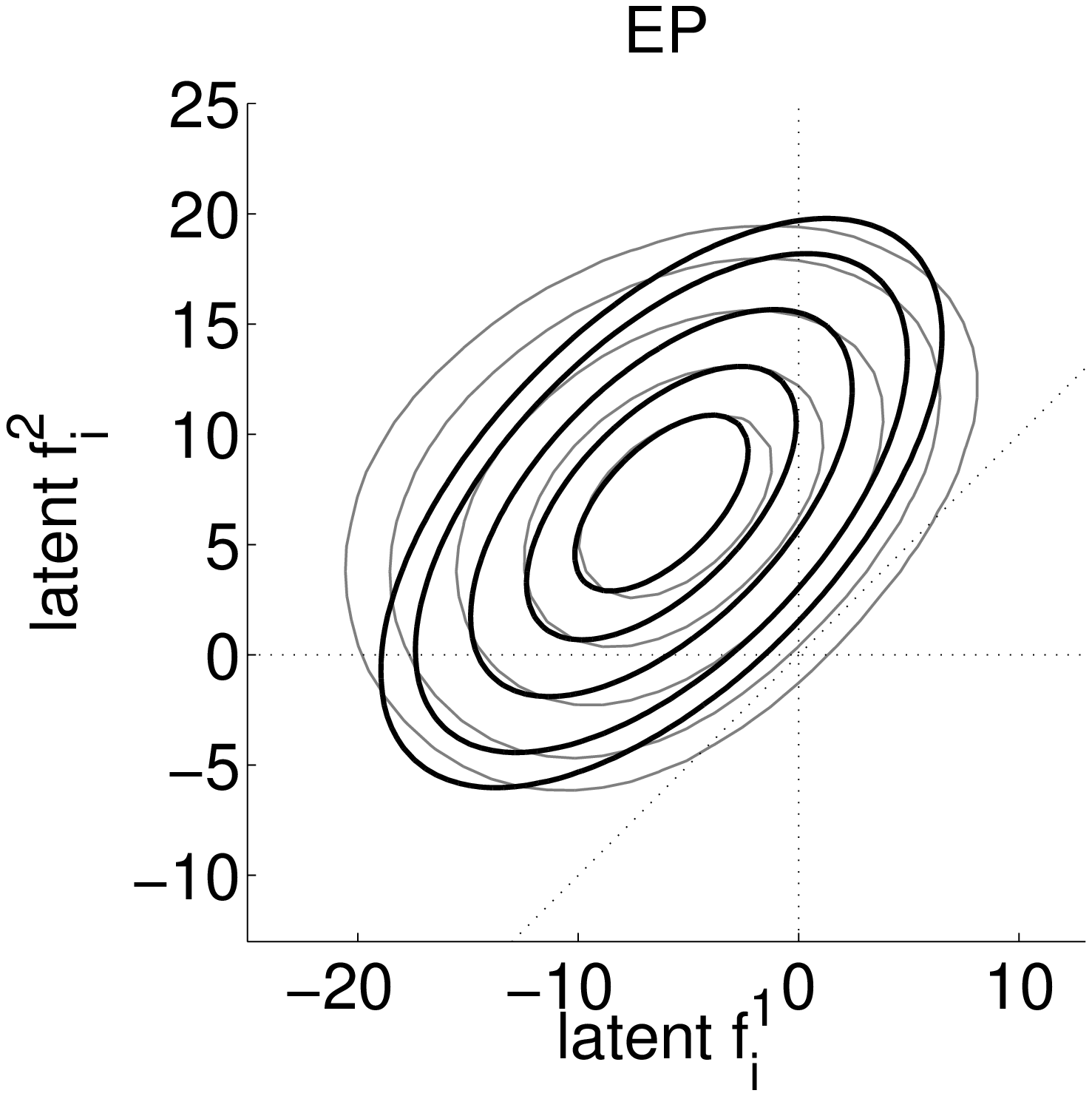}}
  \subfigure[]{\includegraphics[scale=0.19]{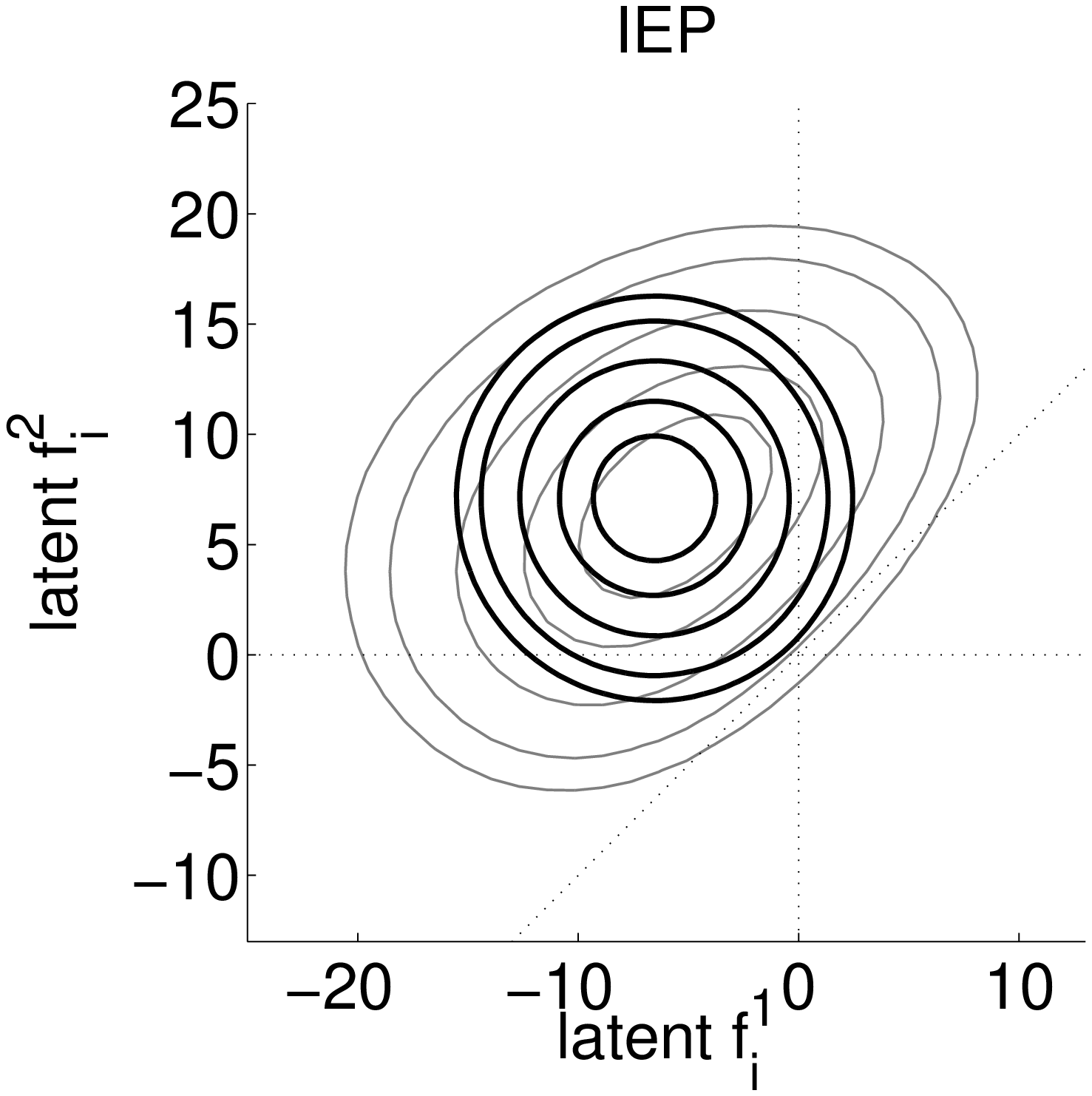}}
  \subfigure[]{\includegraphics[scale=0.19]{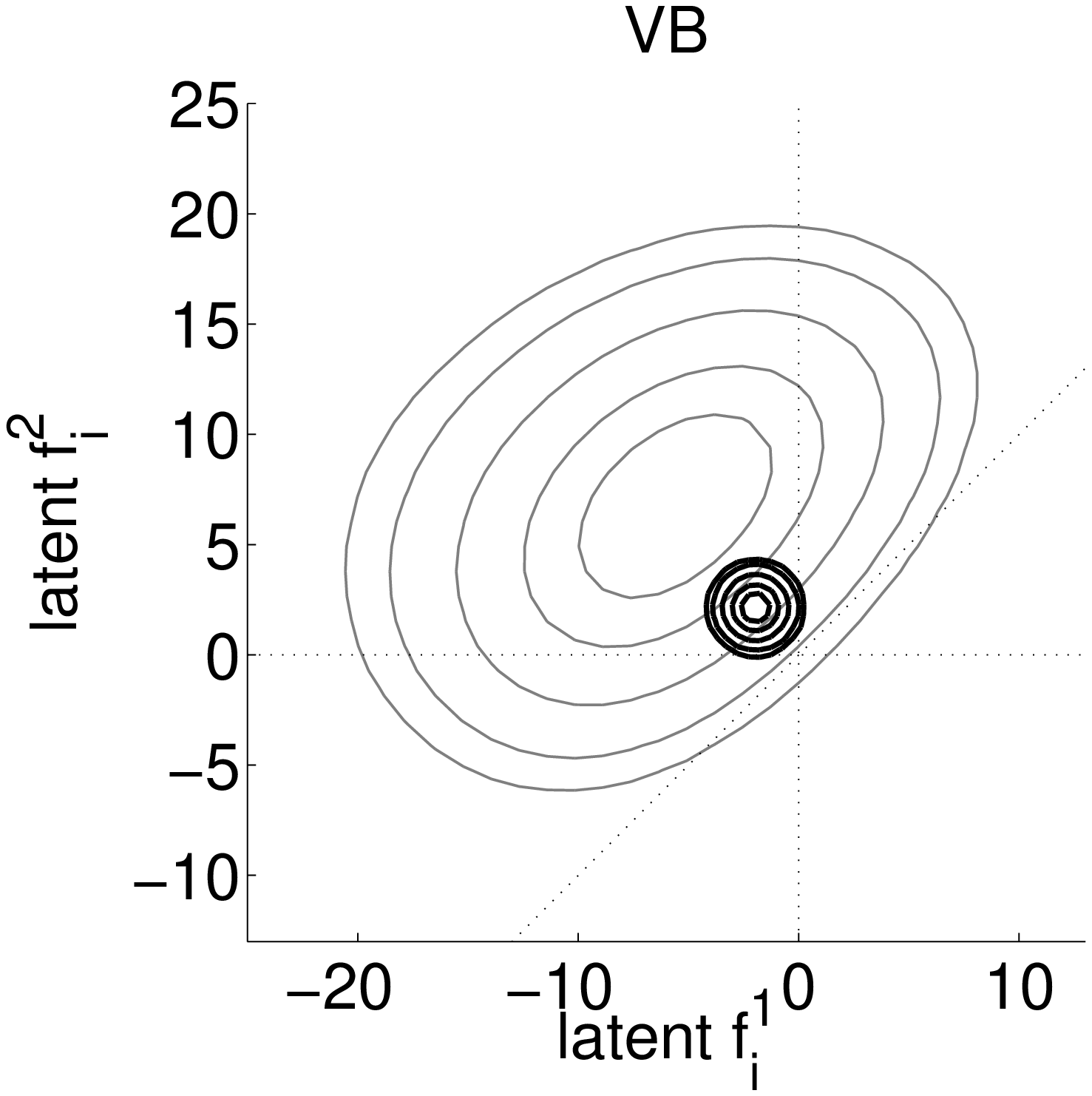}}
  \subfigure[]{\includegraphics[scale=0.19]{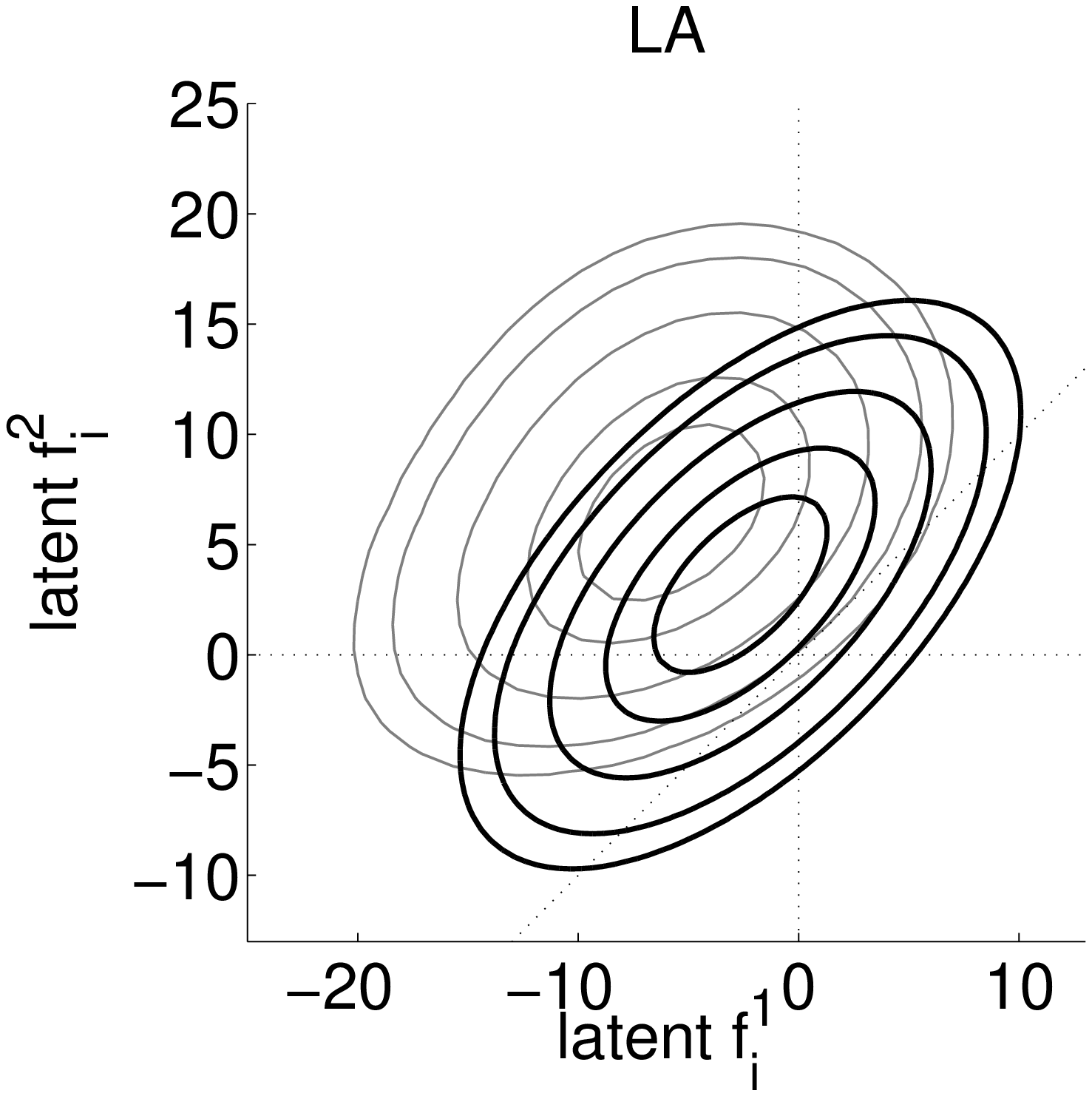}}
 \centering
  \subfigure[]{\includegraphics[scale=0.19]{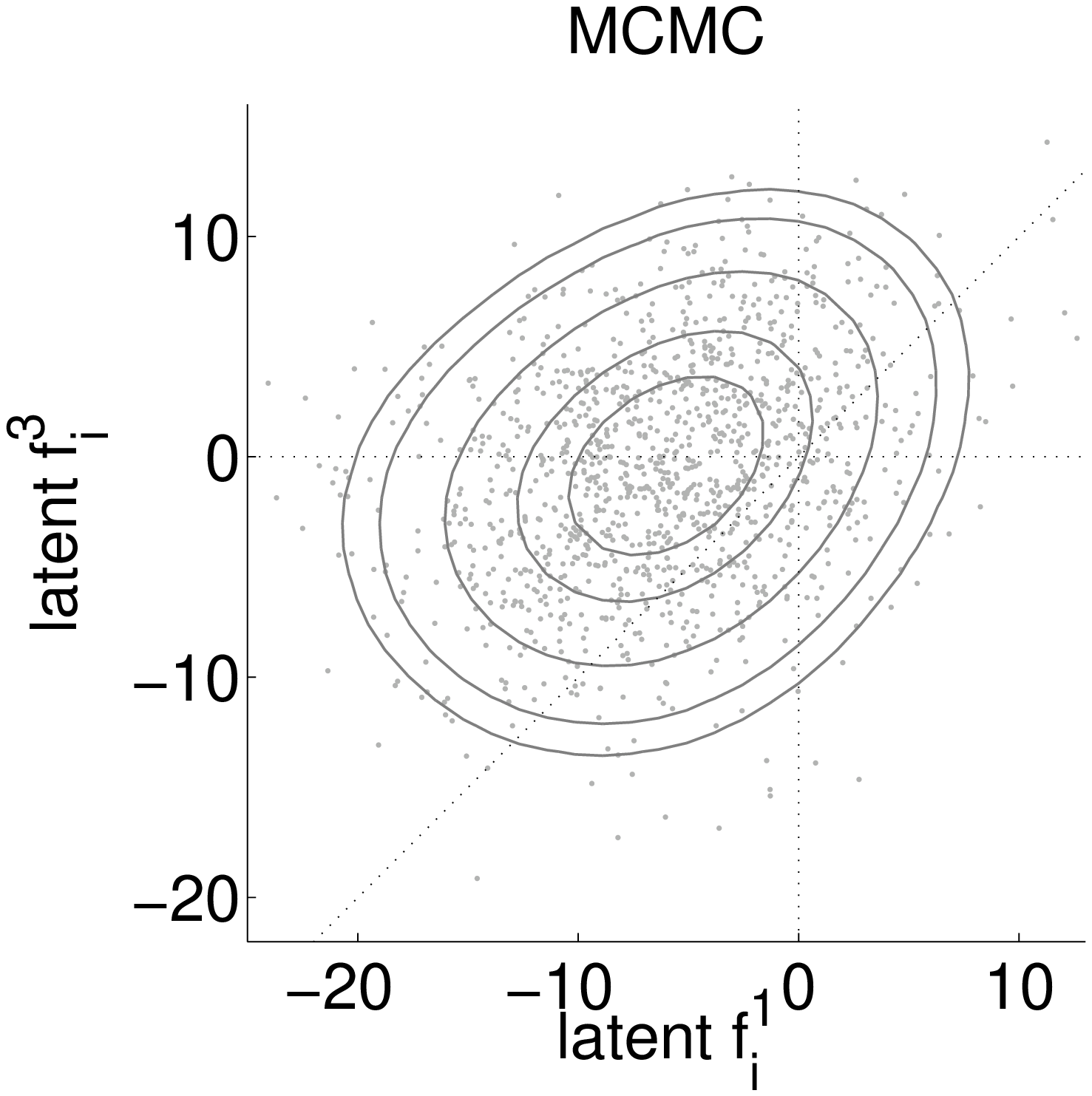}}
  \subfigure[]{\includegraphics[scale=0.19]{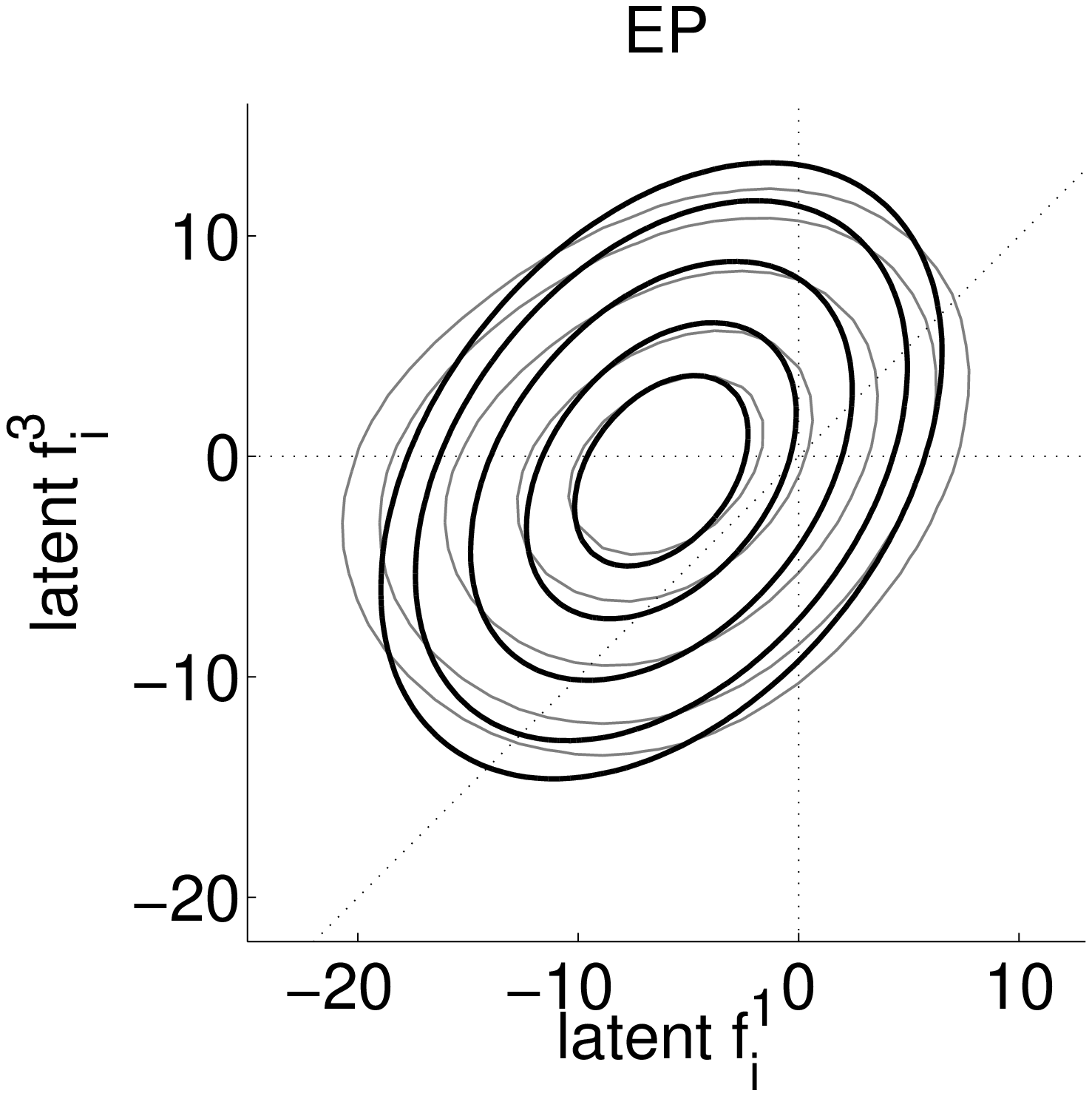}}
  \subfigure[]{\includegraphics[scale=0.19]{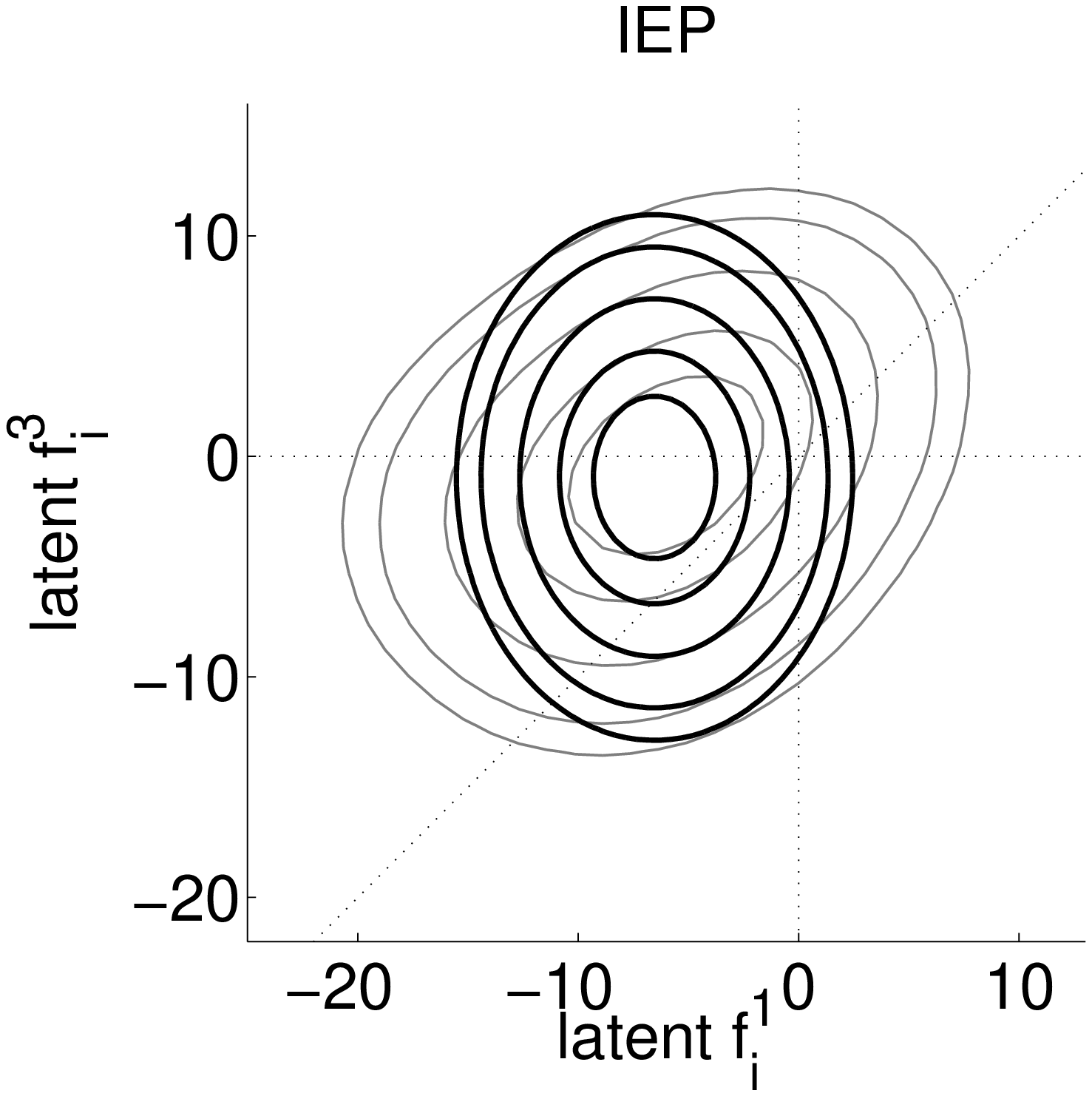}}
  \subfigure[]{\includegraphics[scale=0.19]{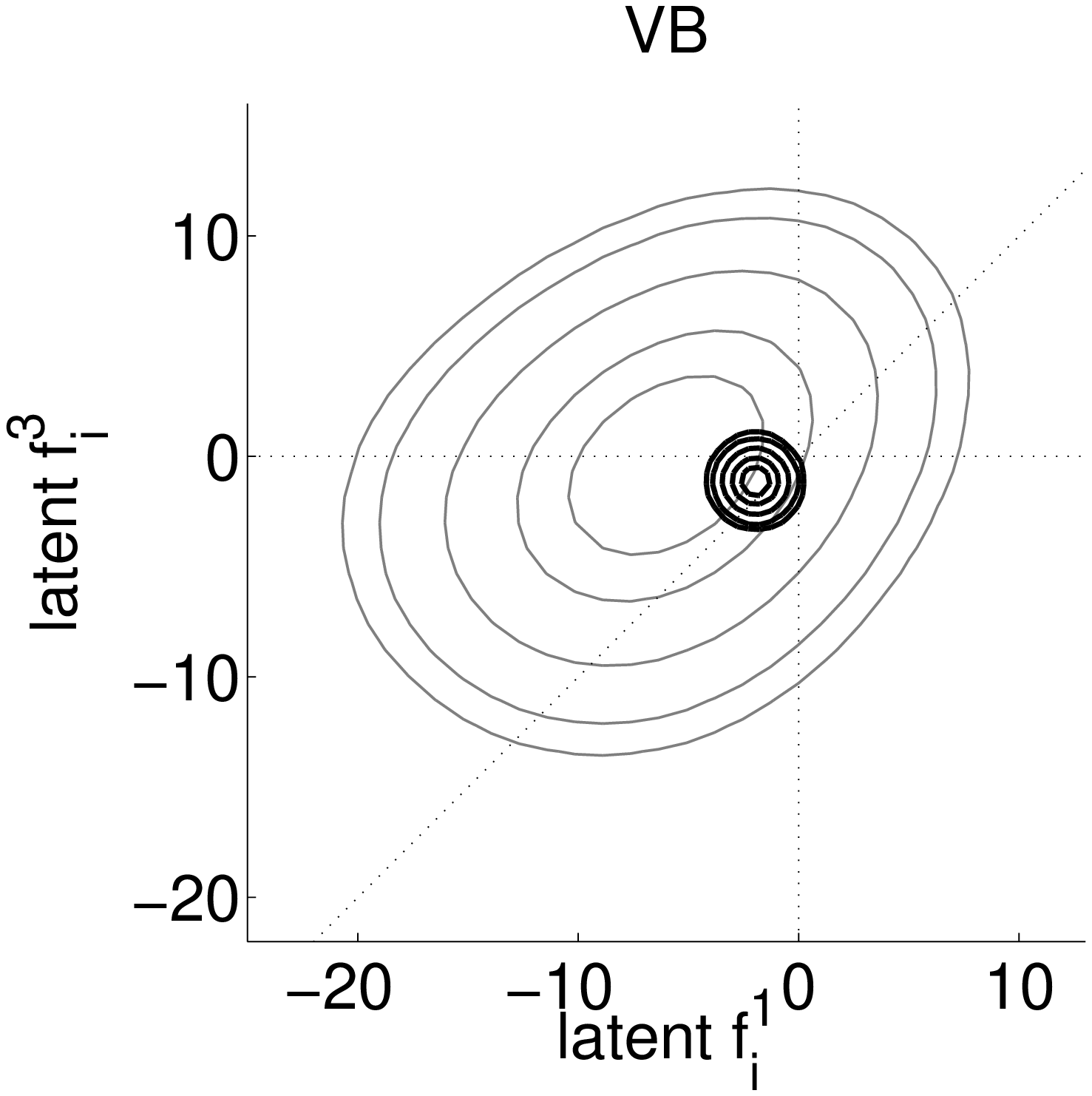}}
  \subfigure[]{\includegraphics[scale=0.19]{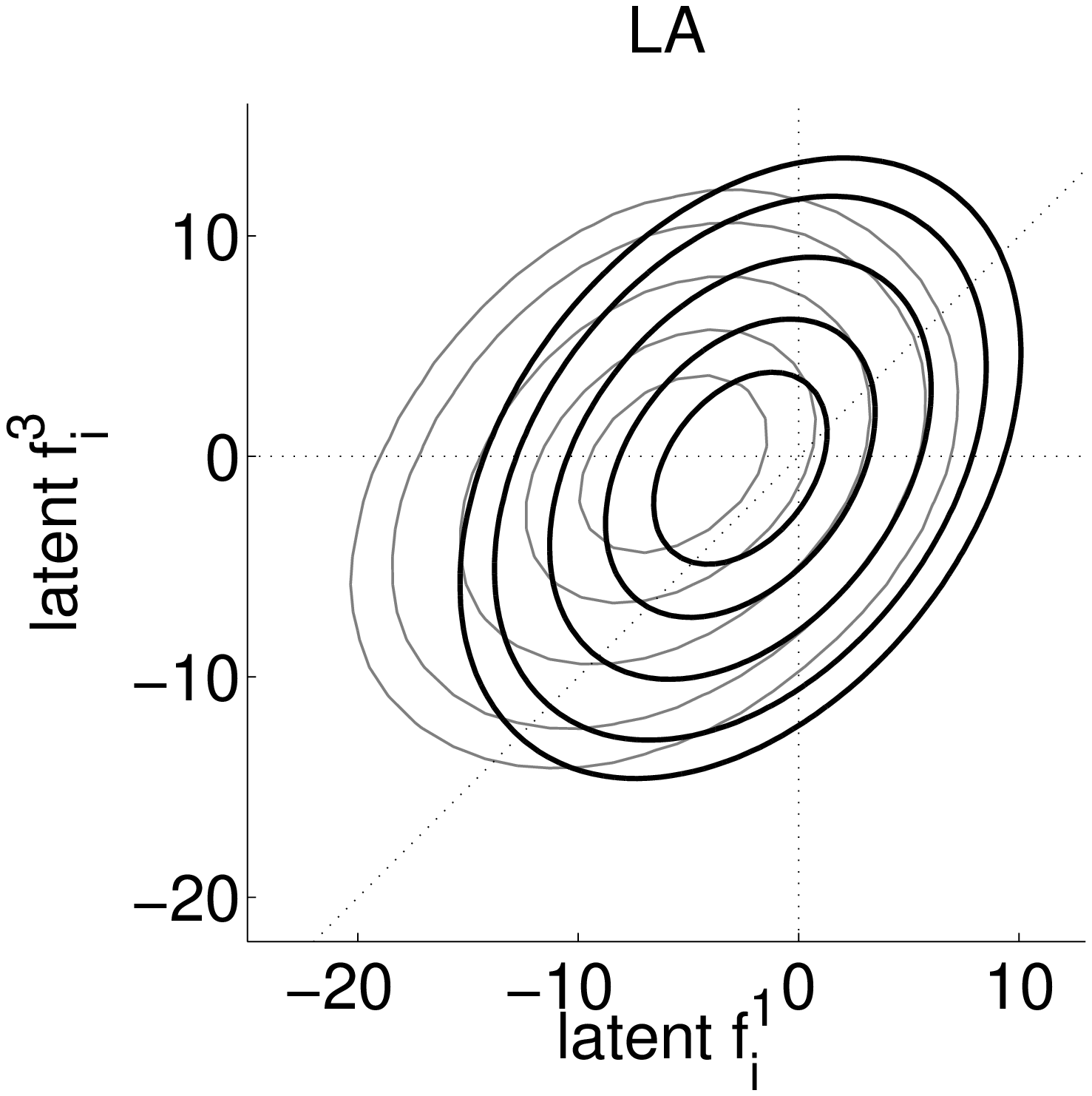}}
 \centering
  \subfigure[]{\includegraphics[scale=0.19]{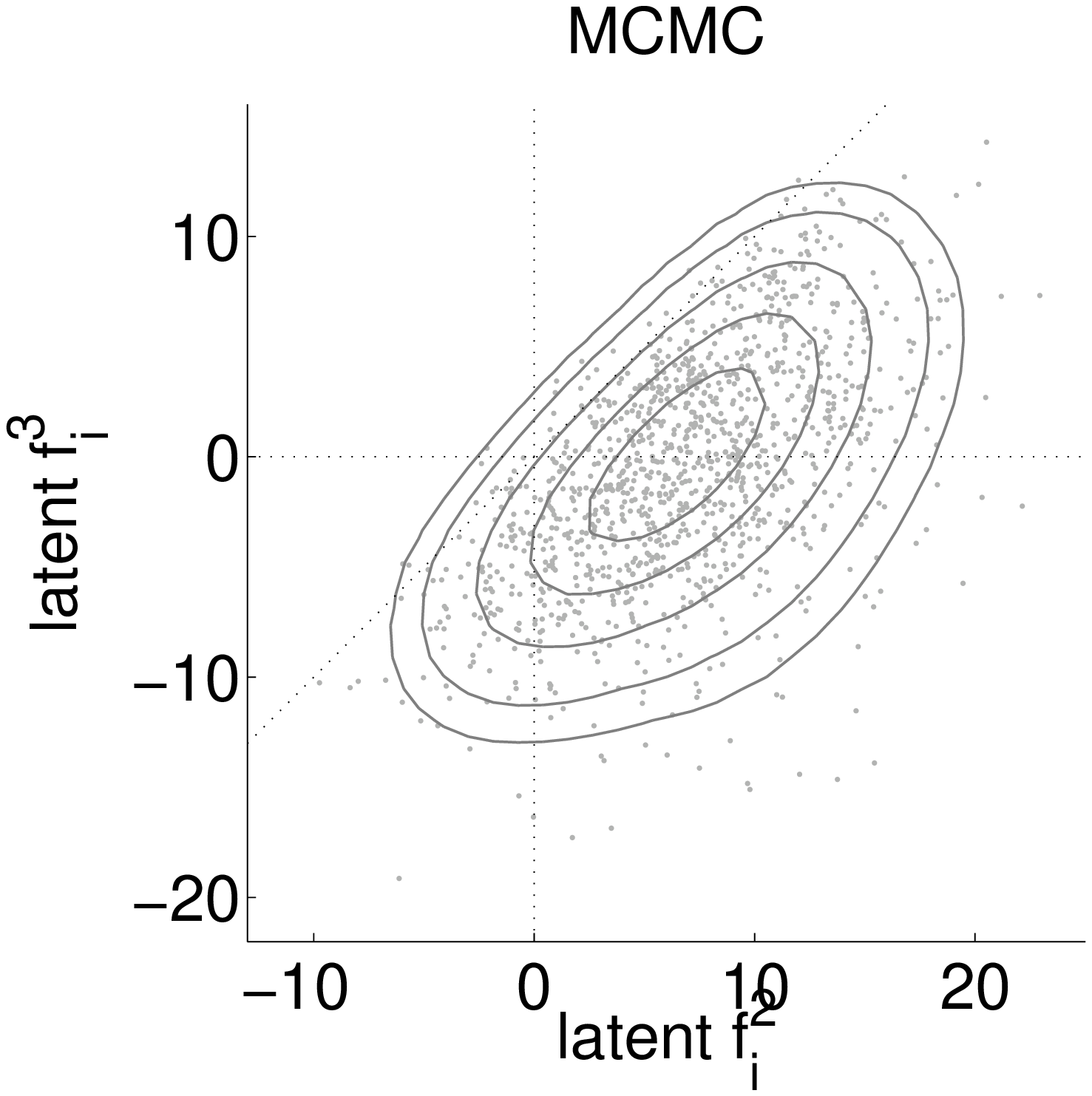}}
  \subfigure[]{\includegraphics[scale=0.19]{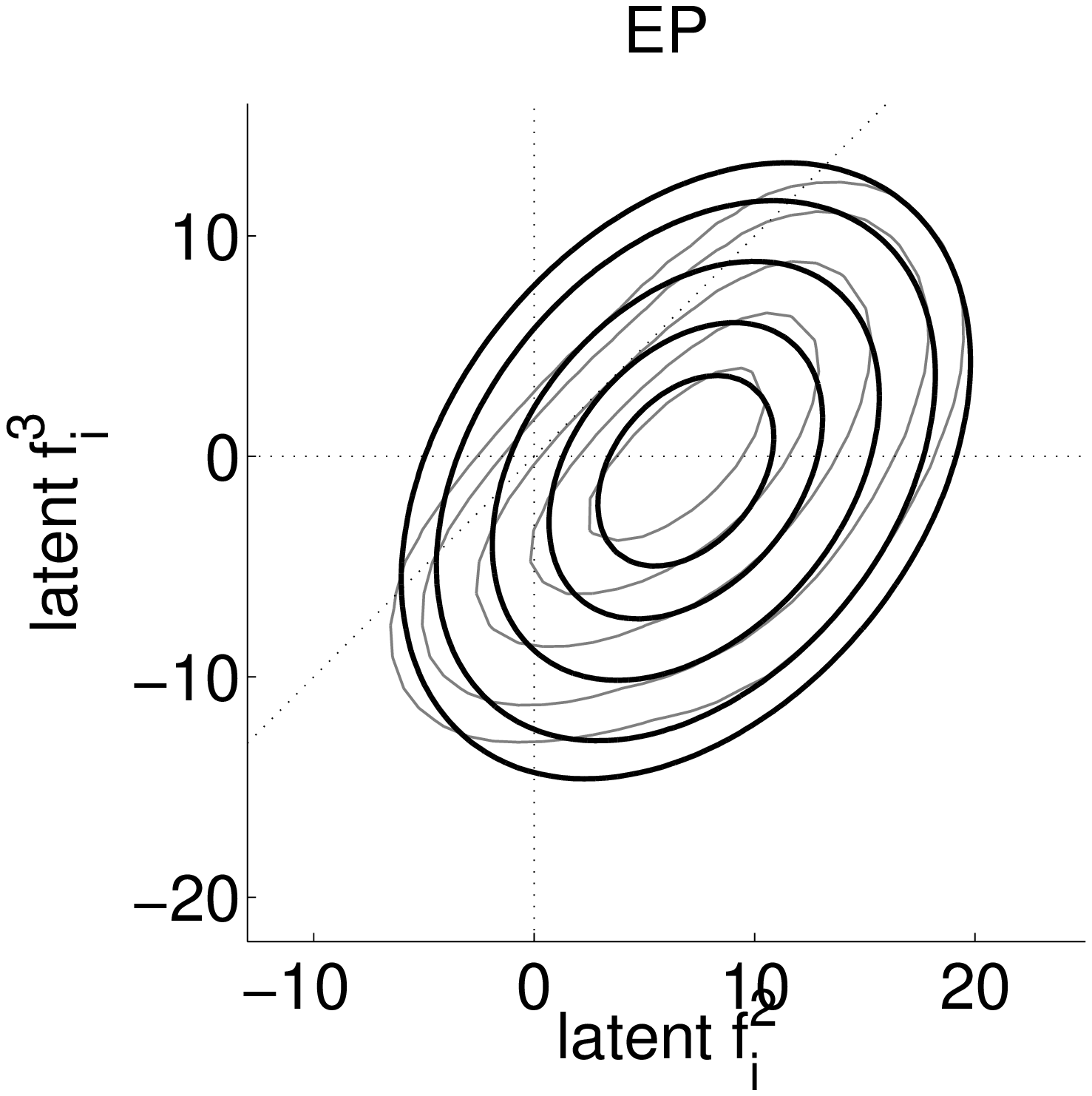}}
  \subfigure[]{\includegraphics[scale=0.19]{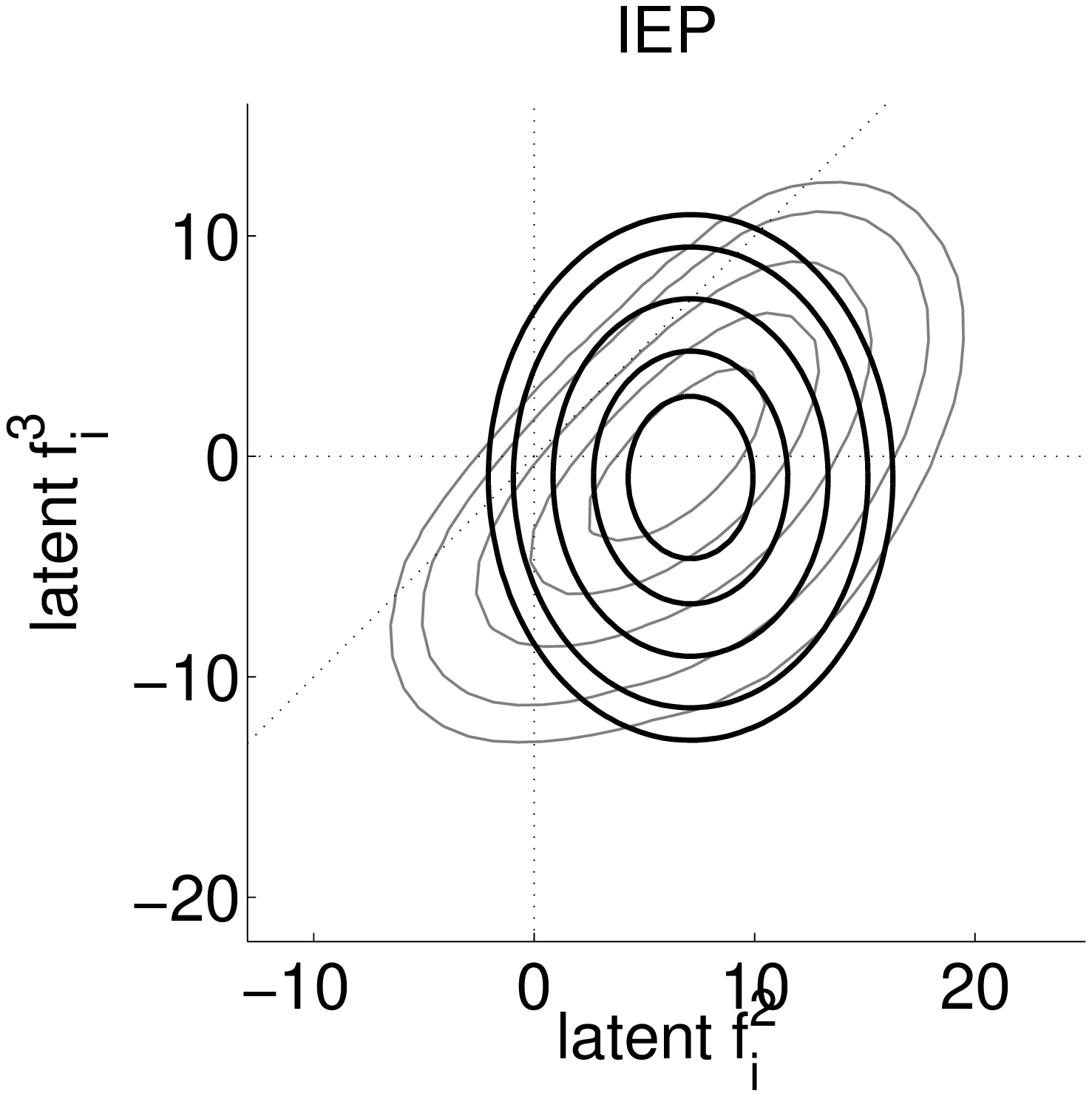}}
  \subfigure[]{\includegraphics[scale=0.19]{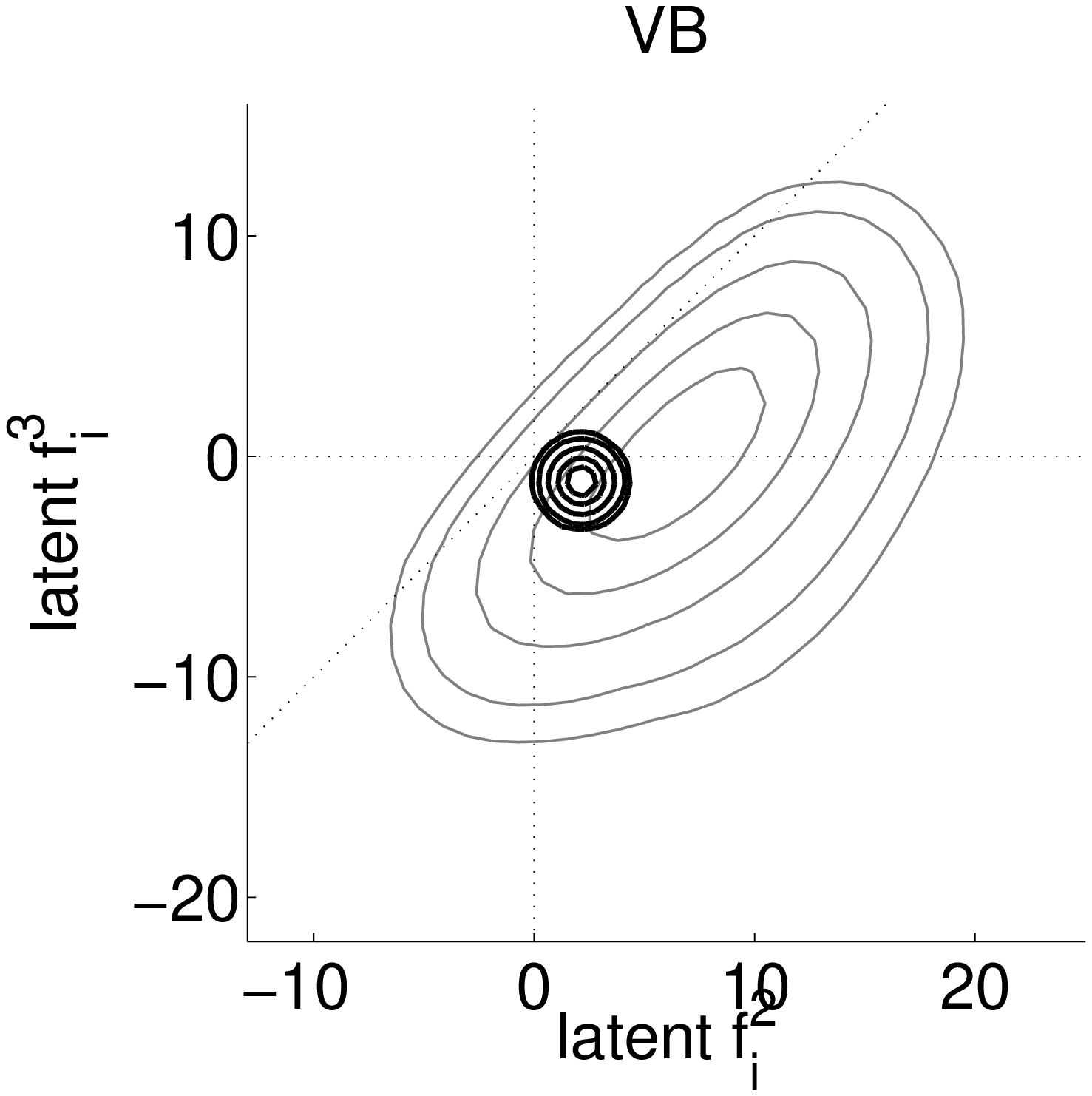}}
  \subfigure[]{\includegraphics[scale=0.19]{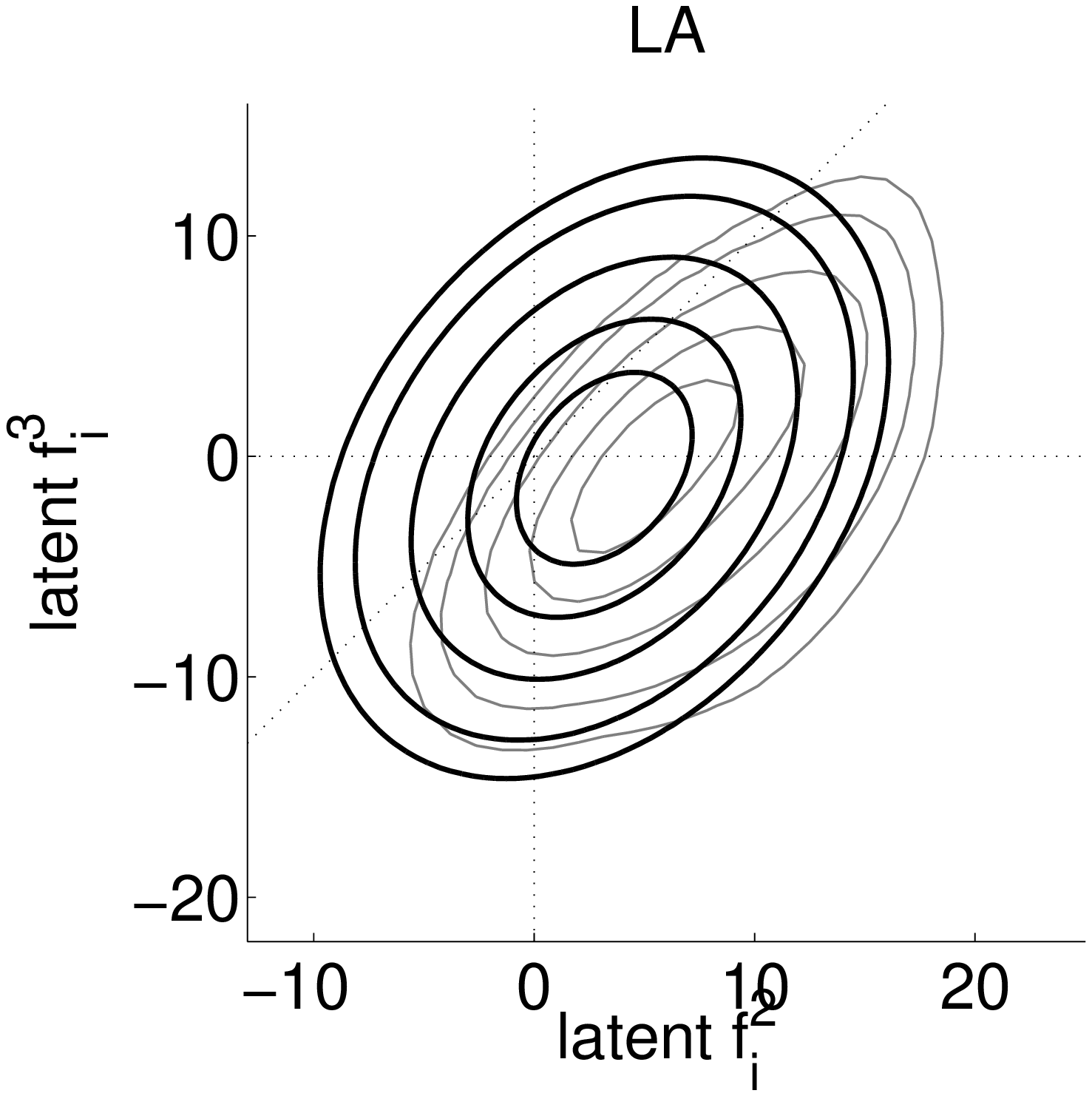}}
  \caption{Marginal posterior distributions for one training point
    with the true class label being 2 on the USPS 3 vs. 5 vs. 7 data.
    Each row corresponds to one of the latent pairs ($f_i^1$,$f_i^2$),
    ($f_i^1$,$f_i^3$), and ($f_i^2$,$f_i^3$). The first column shows a
    scatter-plot of MCMC samples drawn from the posterior, and the
    estimated density contour levels which correspond to the areas
    that include approximately 95\%, 90\% 75\%, 50\%, and 25\% of the
    probability mass. The columns 2-5 show the equivalent contour
    levels of EP, IEP, VB and LA approximations (bold black lines) and
    the contour plot of MCMC approximation (gray lines) for
    comparison. Note that the last column visualizes a different
    marginal distribution because LA uses the softmax likelihood. The
    hyperparameters of the squared exponential covariance function
    were fixed at $\log(\sigma^2)=4$ and $\log(l)=2$ to obtain a
    non-Gaussian posterior distribution.}
  \label{figure_nongaussian_latents}
\end{figure*}

Figure \ref{figure_nongaussian_latents} shows an example of the latent
marginal posterior distributions for one training point with the
correct class label being 2. For each method, the latent pairs
($f_i^1$,$f_i^2$), ($f_i^1$,$f_i^3$), and ($f_i^2$,$f_i^3$), are
shown.
The EP approximation agrees reasonably well with the MCMC samples.
IEP underestimates the latent uncertainty, especially near the
training inputs because of the skewing effect of the likelihood. This
seems to affect more the predictive probabilities of the training
points in Figure \ref{figure_nongaussian_probabilities}(b), which can
be seen also in Figure \ref{figure_toy_example}(a) further away from
the decision boundary near the input $x_i$. Figure
\ref{figure_nongaussian_latents} shows that the VB method seriously
underestimates the latent uncertainty. The independence assumption of
VB leads to an isotropic approximate distribution, and although the
predictive probabilities for the training cases are somewhat
consistent with MCMC, the predictions on the test data fail (plots (c)
and (h) in Figure \ref{figure_nongaussian_probabilities}). The LA
approximation captures some of the dependencies between the latent
variables associated with different classes, but the joint mode of
$\f$ is a poor estimate for the true mean, which causes inaccurate
predictive probabilities (plots (d) and (i) in Figure
\ref{figure_nongaussian_probabilities}). The VB mean estimate is also
closer to LA than MCMC, although LA uses a different observation
model.

\begin{figure*}[!t]
 \centering
  \subfigure[]{\includegraphics[scale=0.19]{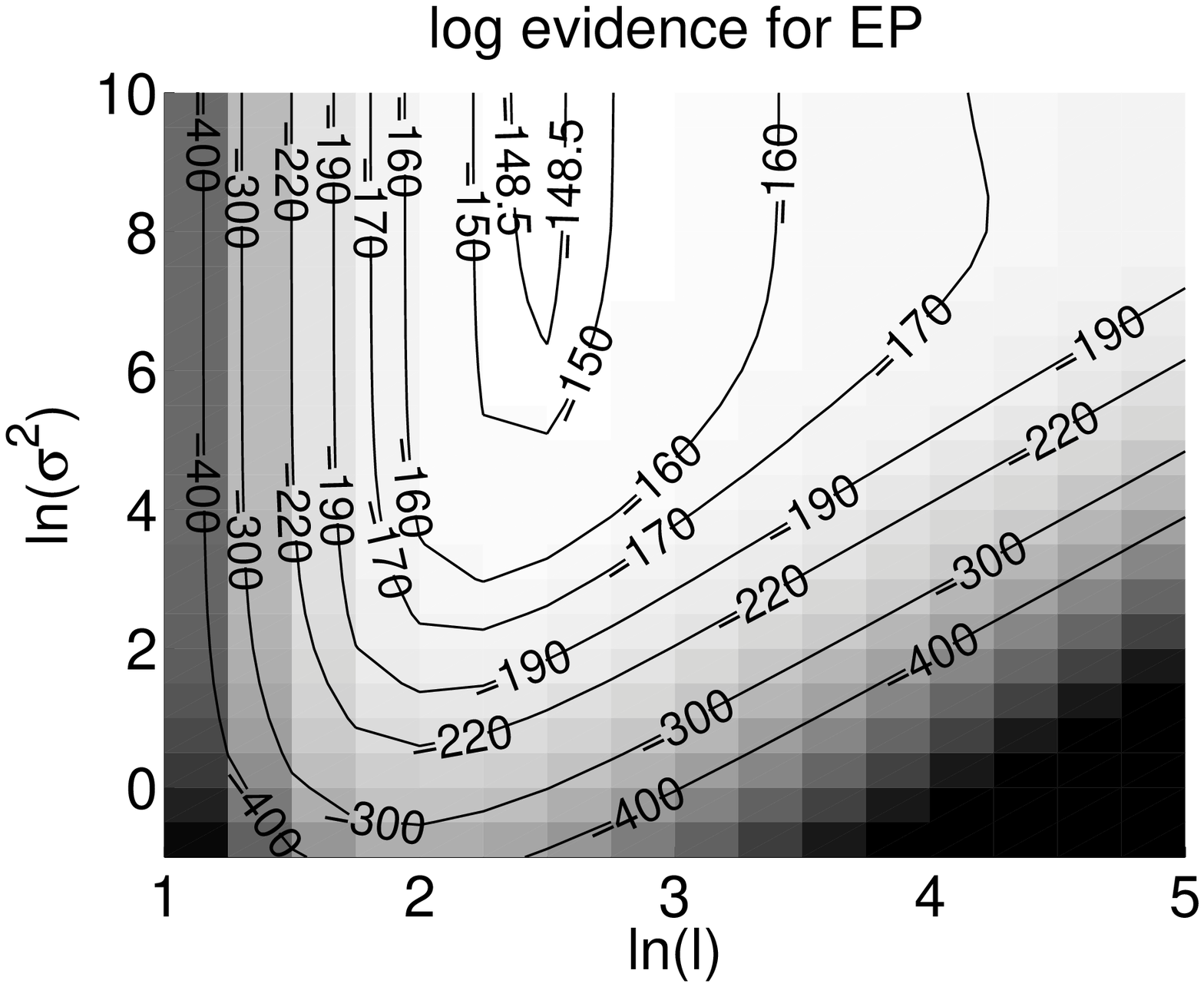}}
  \subfigure[]{\includegraphics[scale=0.19]{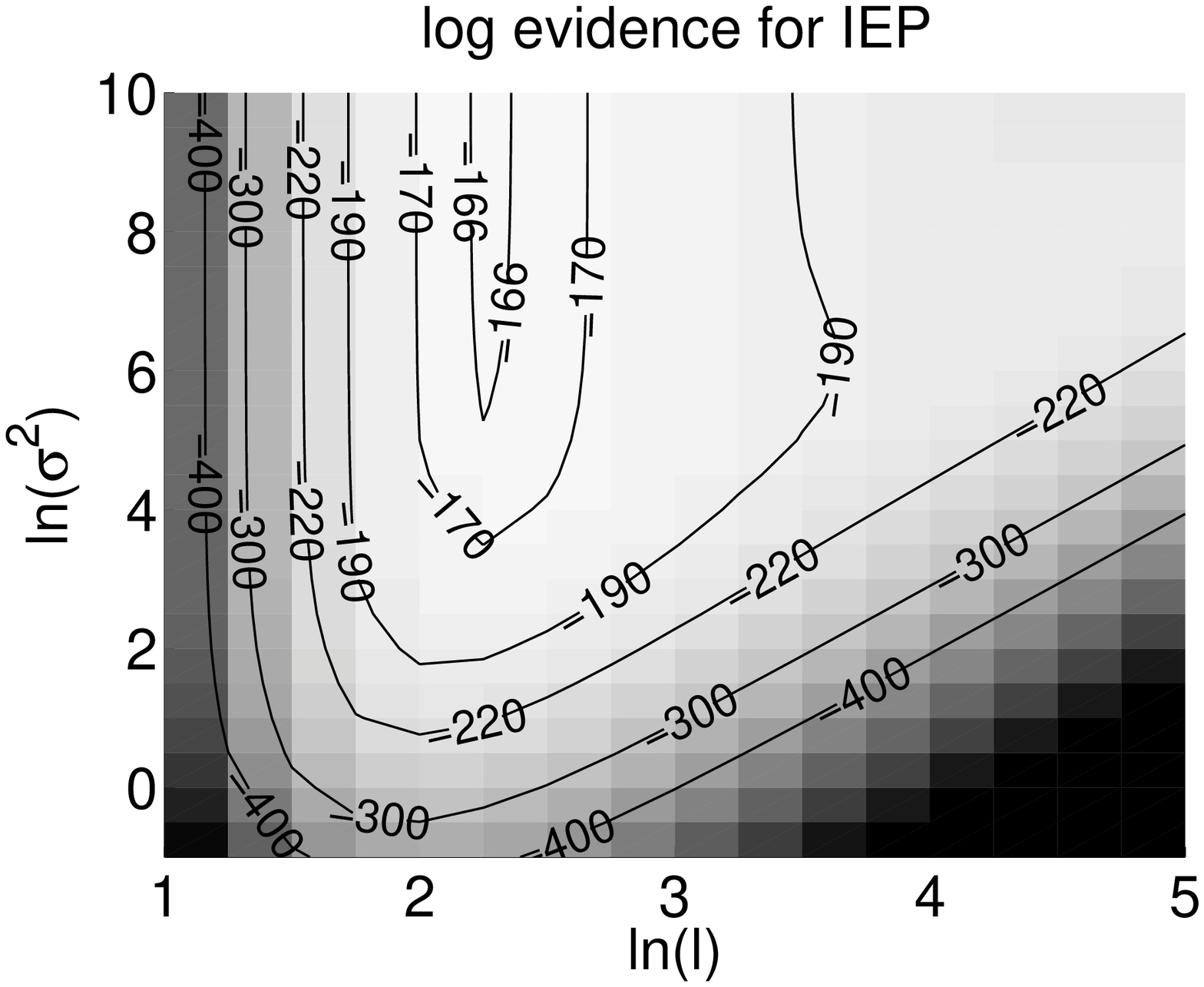}}
  \subfigure[]{\includegraphics[scale=0.19]{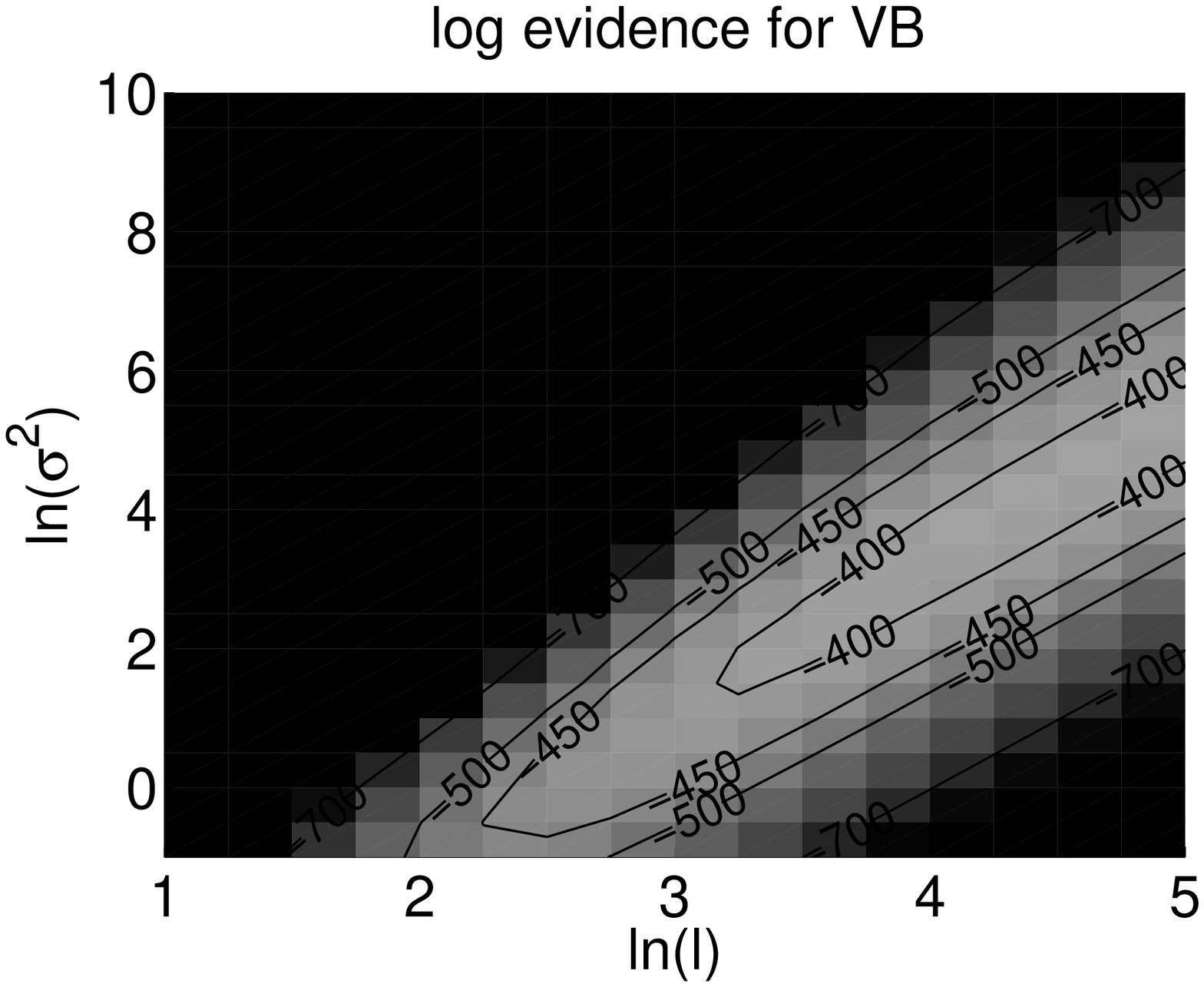}}
  \subfigure[]{\includegraphics[scale=0.19]{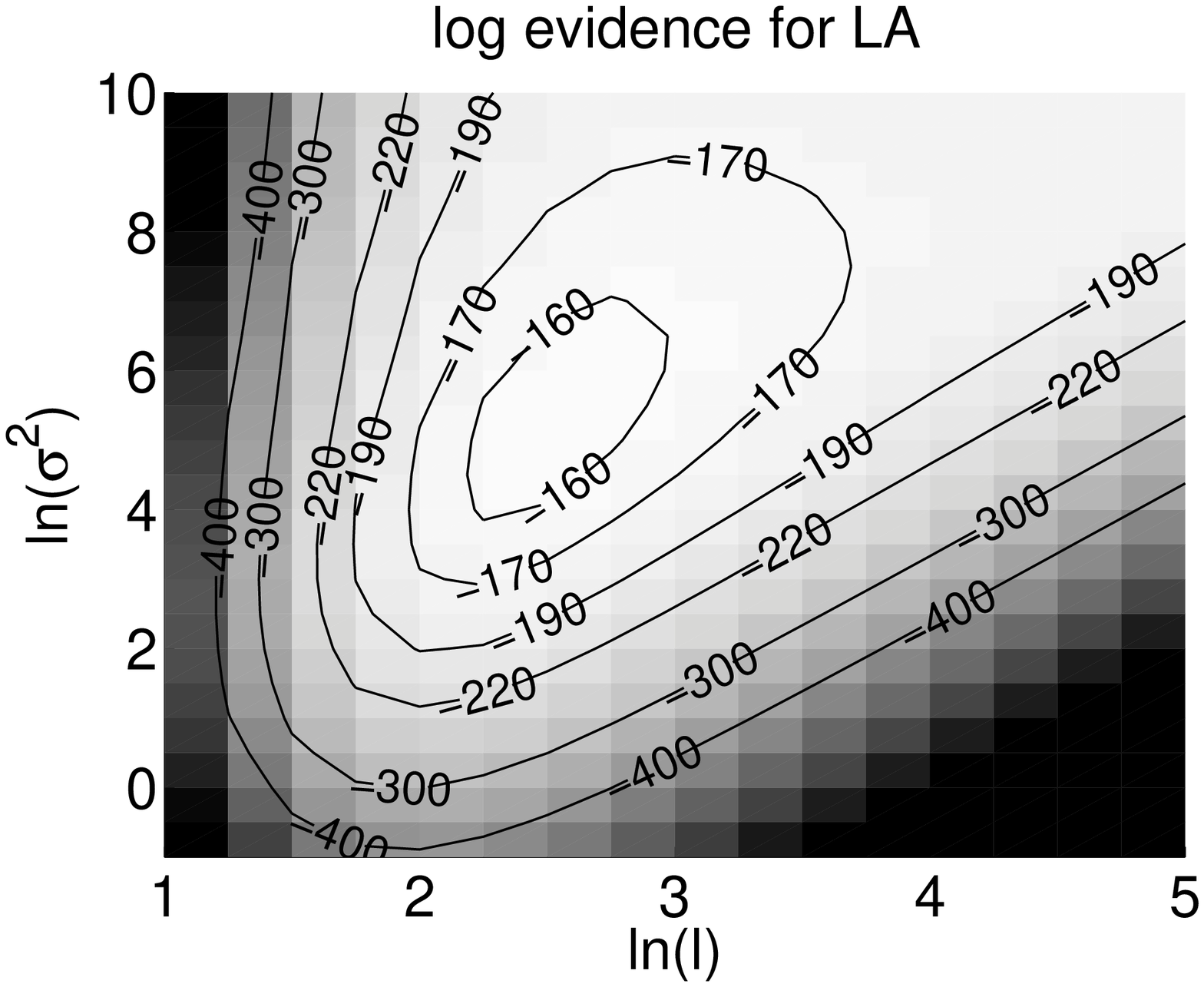}}
 \centering
  \subfigure[]{\includegraphics[scale=0.19]{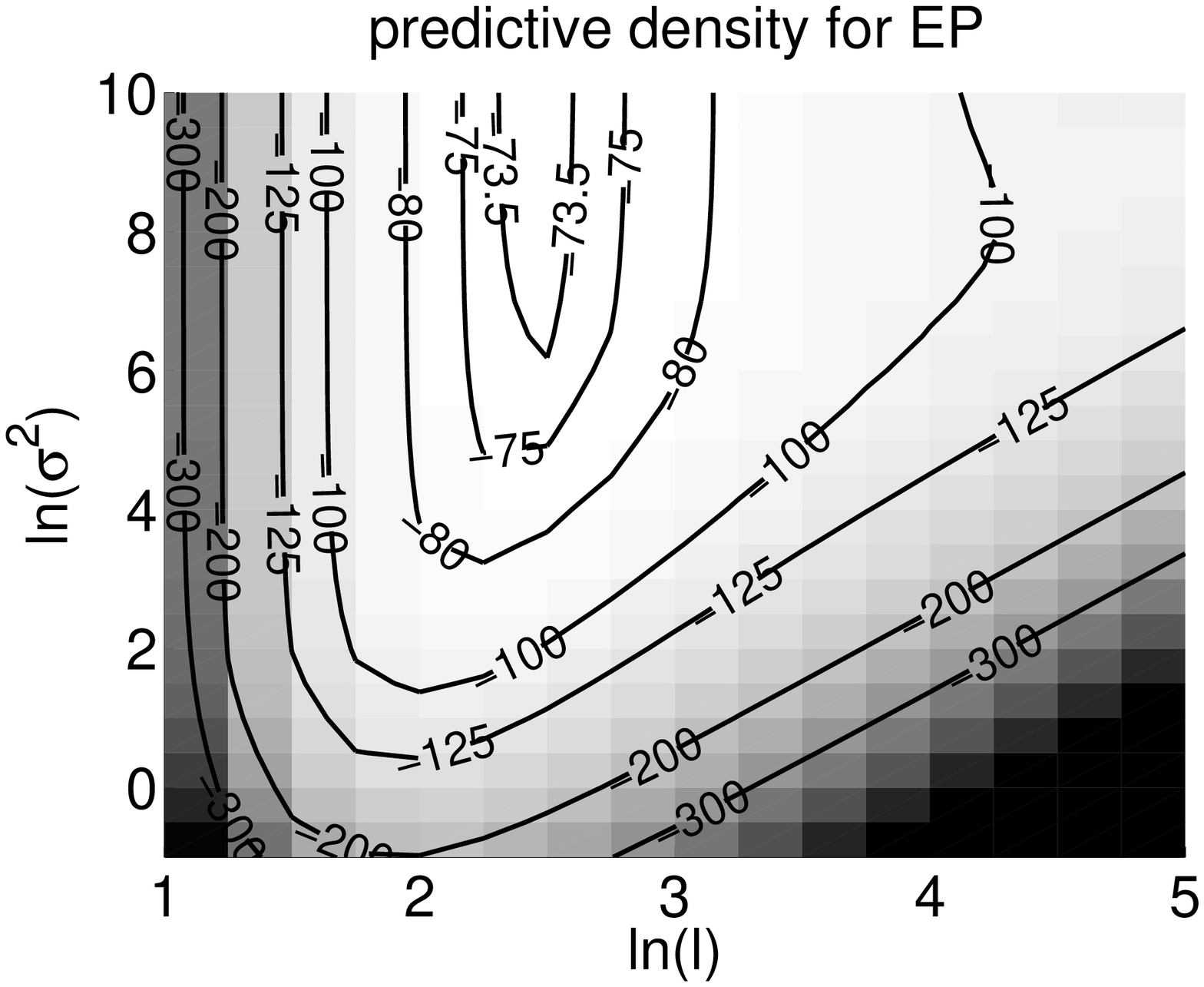}}
  \subfigure[]{\includegraphics[scale=0.19]{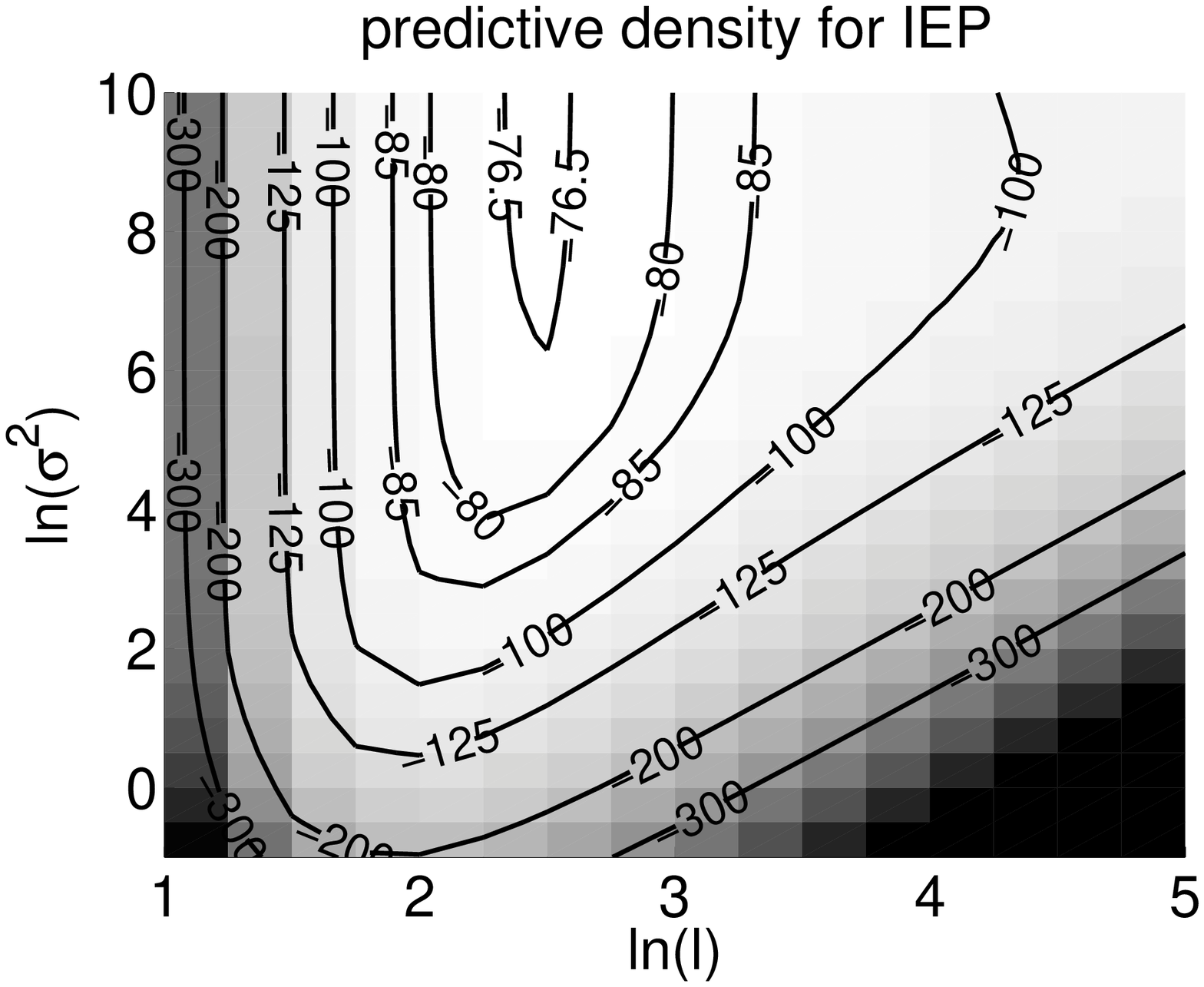}}
  \subfigure[]{\includegraphics[scale=0.19]{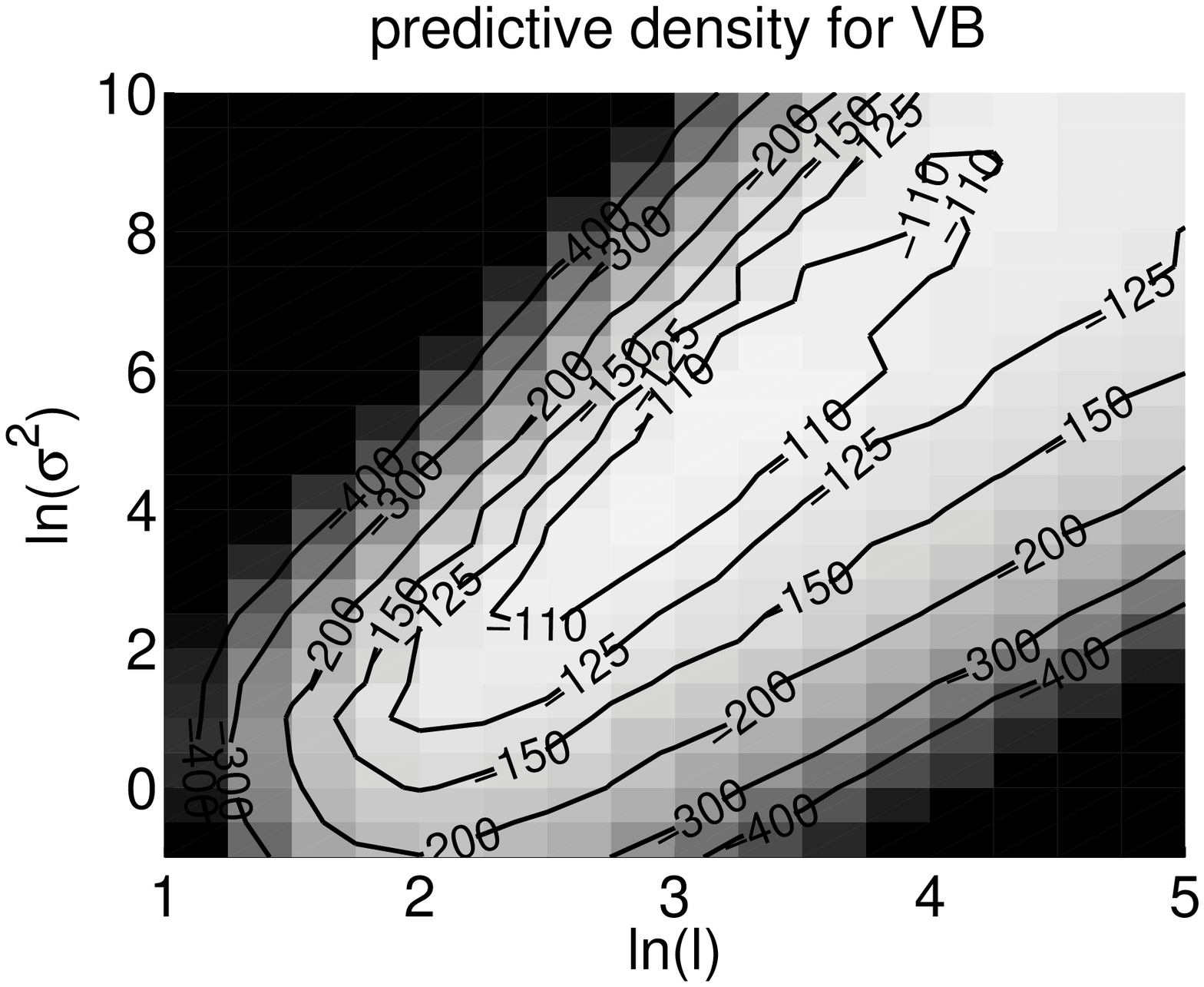}}
  \subfigure[]{\includegraphics[scale=0.19]{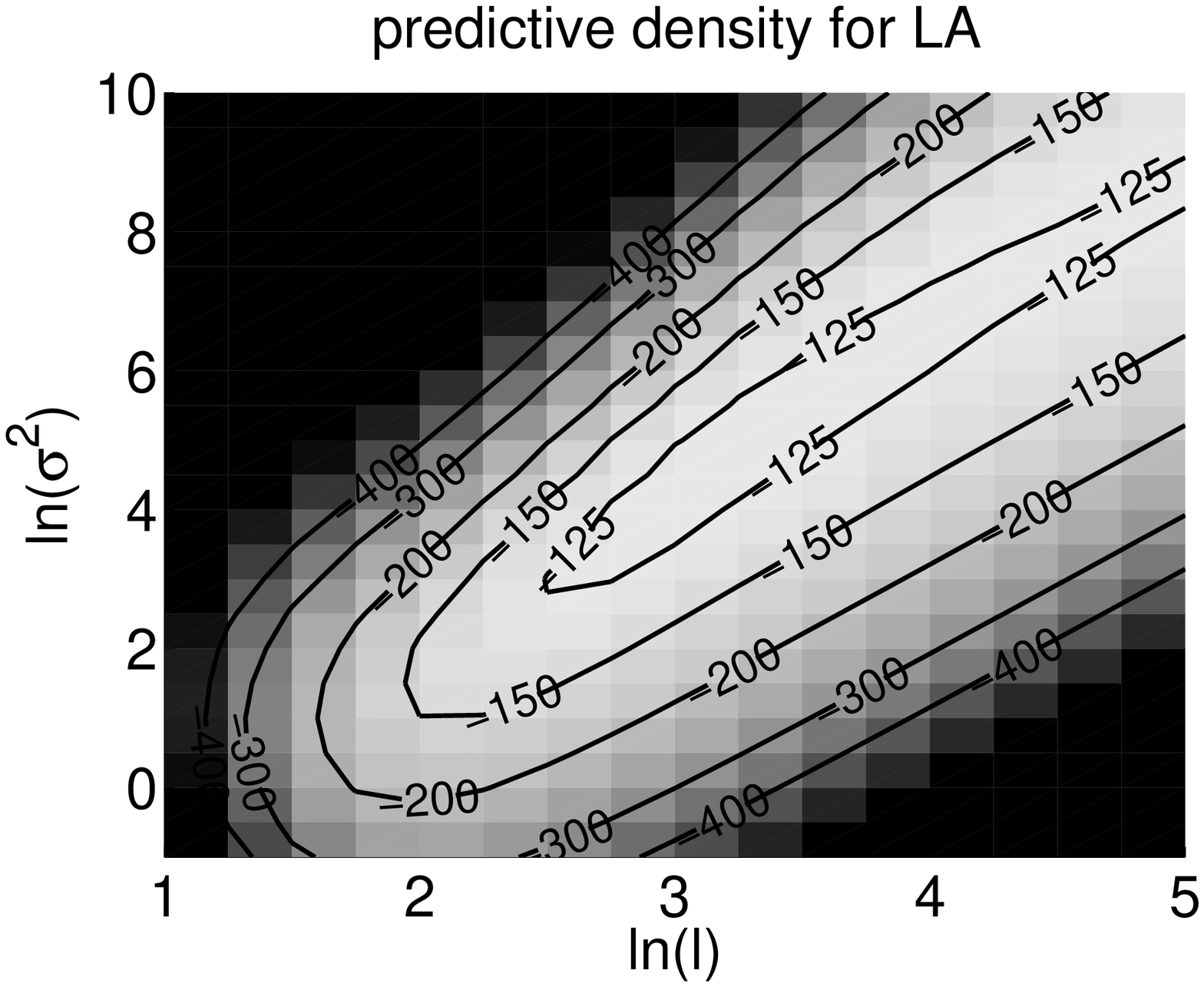}}
 \centering
  \subfigure[]{\includegraphics[scale=0.19]{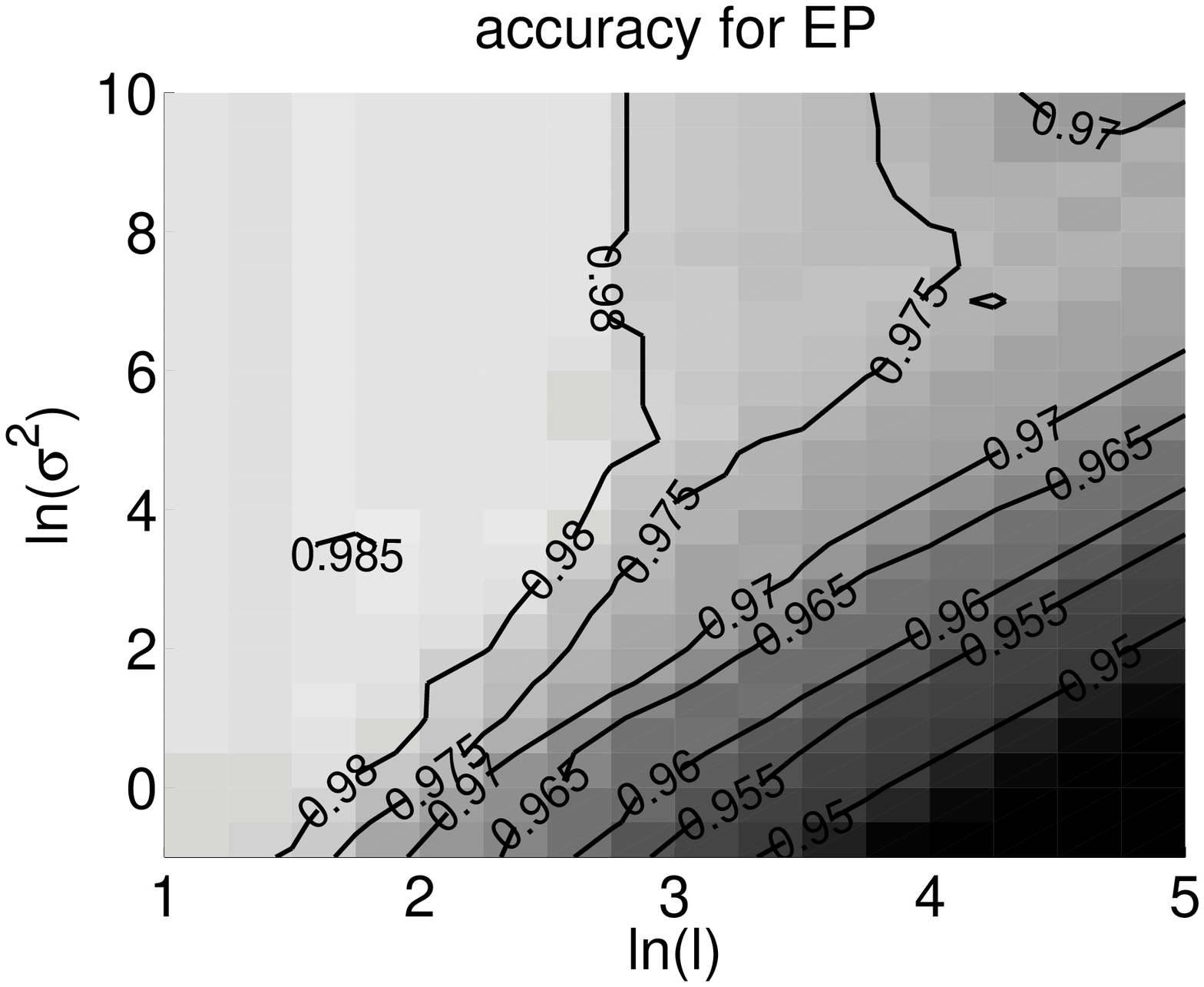}}
  \subfigure[]{\includegraphics[scale=0.19]{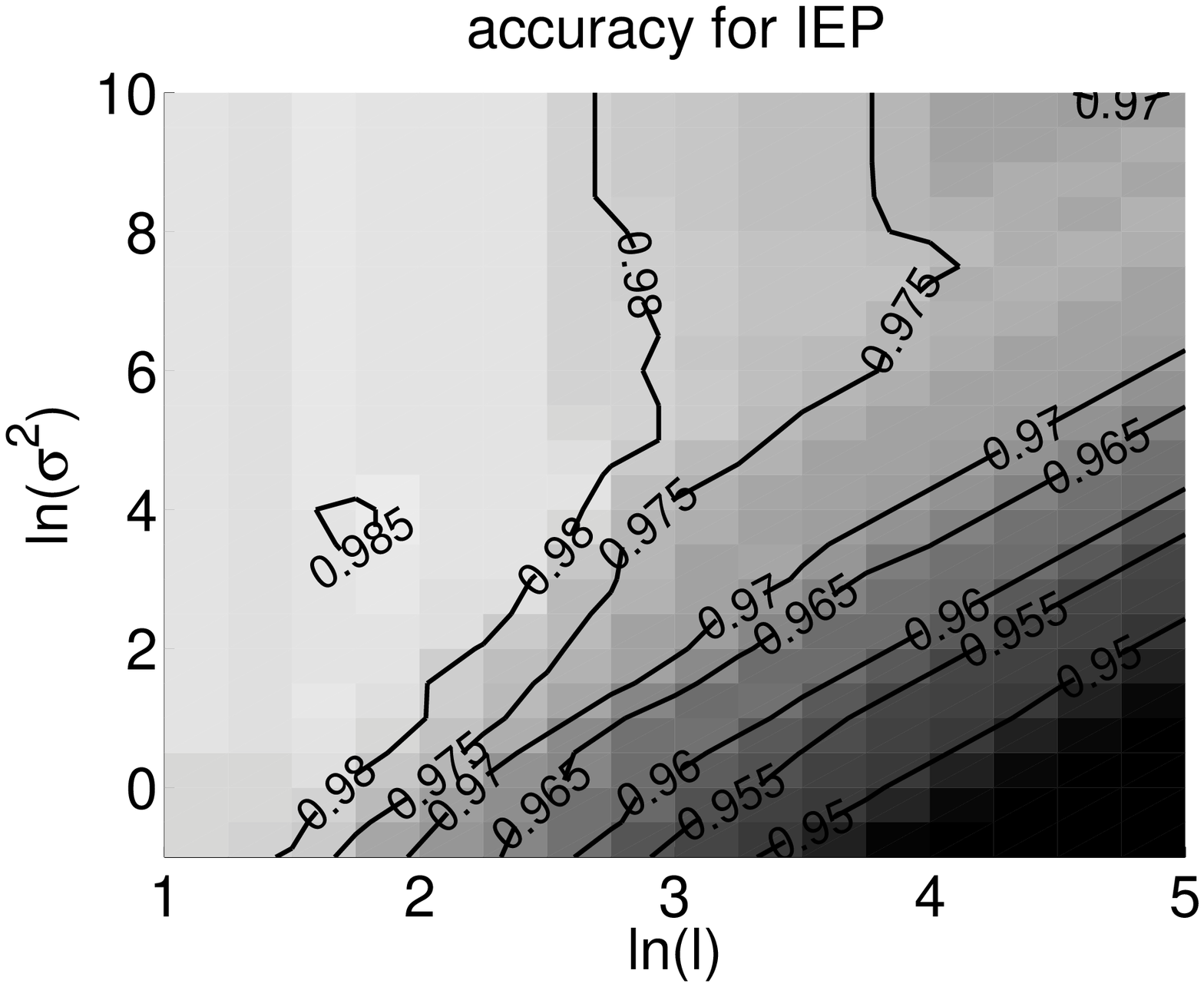}}
  \subfigure[]{\includegraphics[scale=0.19]{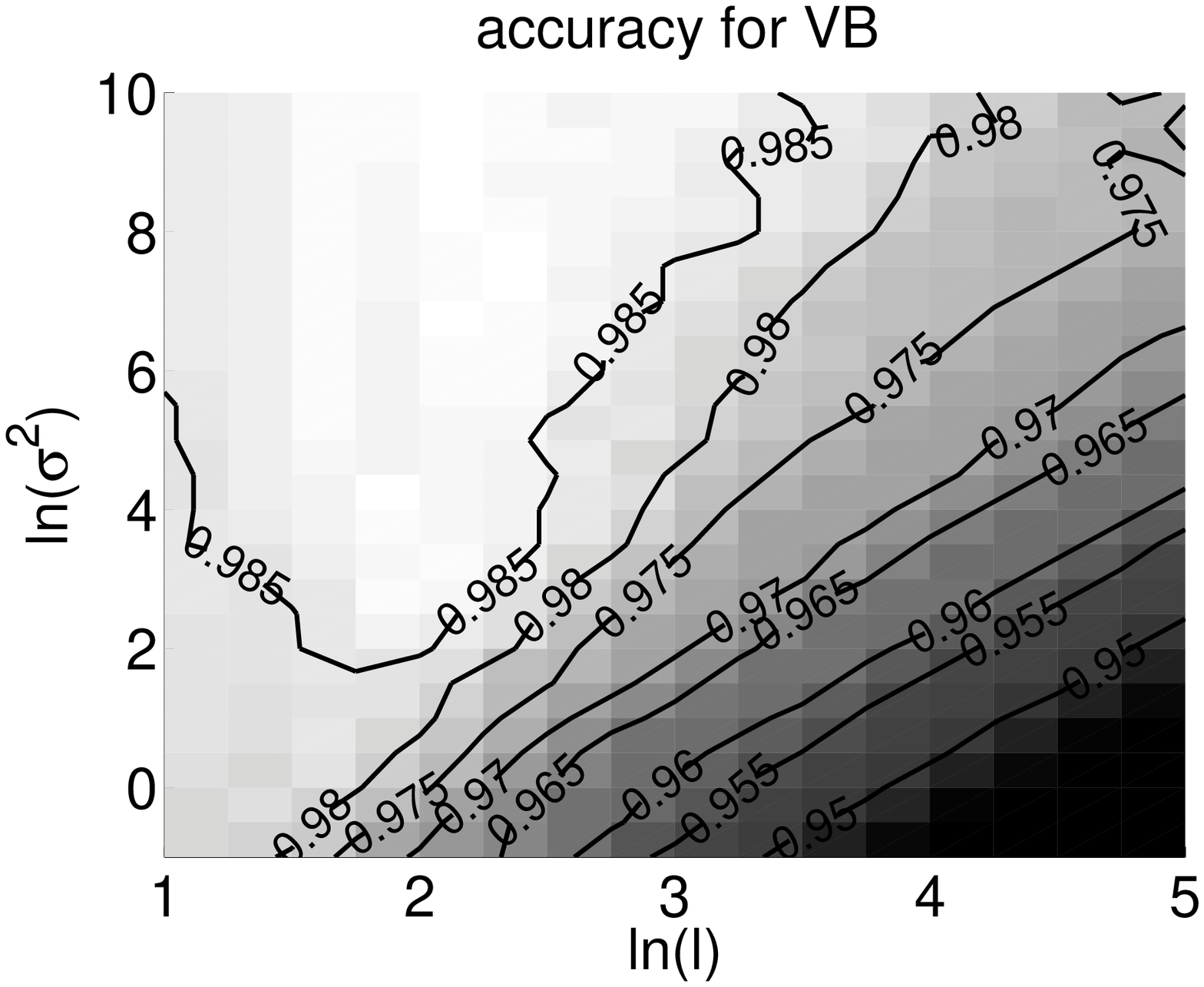}}
  \subfigure[]{\includegraphics[scale=0.19]{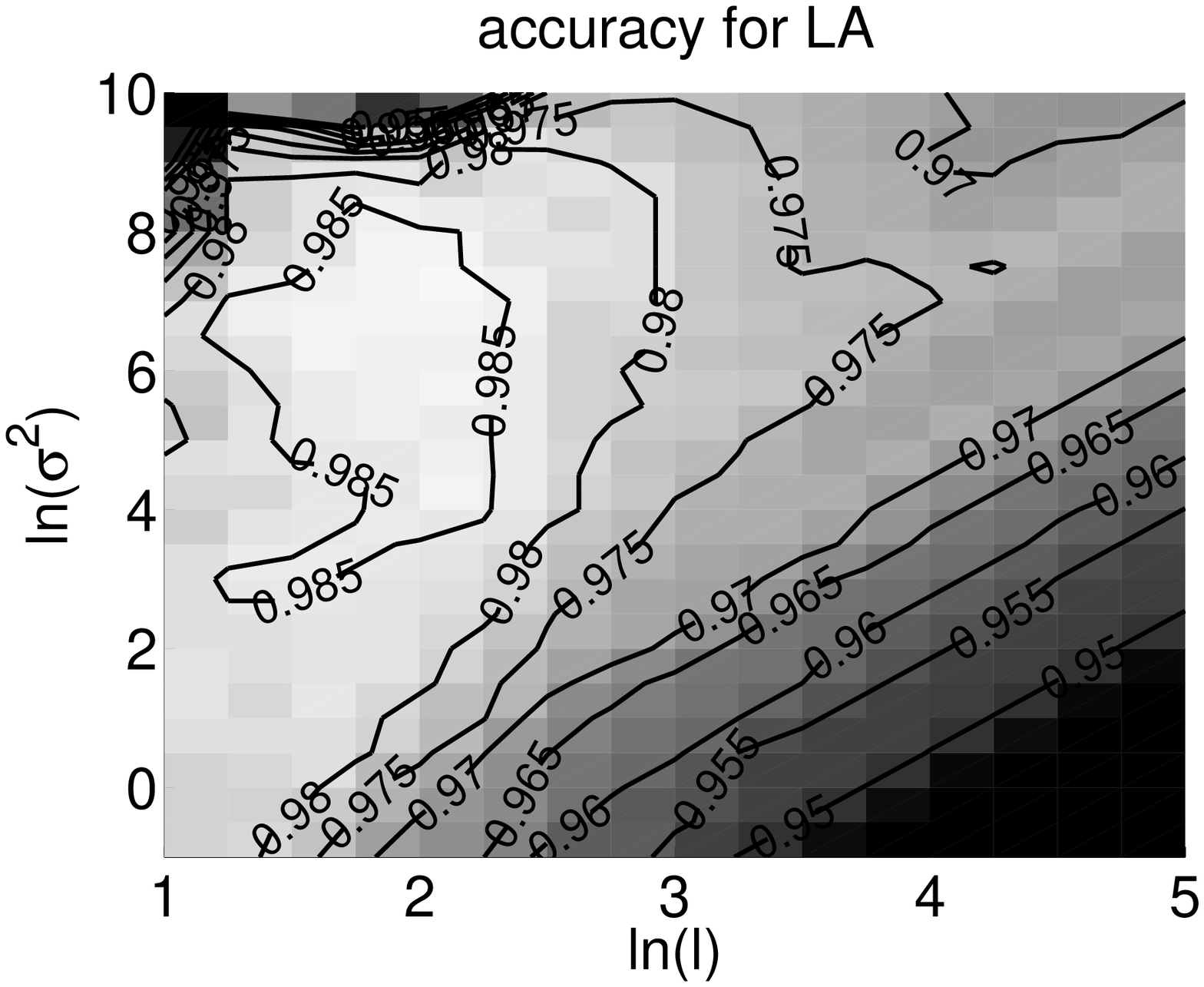}}
  \caption{Marginal likelihood approximations and predictive
    performances as a function of the log-lengthscale $\log(l)$ and
    log-magnitude $\log(\sigma^2)$ for EP, IEP, VB and LA on USPS 3
    vs. 5 vs. 7 data. The first row shows the log marginal likelihood
    approximations, the second row the log predictive densities in a
    test set, and the third row the classification accuracies in a
    test set.}
  \label{figure_surfaces}
\end{figure*}

\citet{kuss2005} and \citet{nickisch2008} discussed how a large value
of the magnitude hyperparameter $\sigma^2$ can lead to a skewed
posterior distribution in binary classification. In the multiclass
setting, similar behavior can be seen in the marginal distributions as
illustrated in Figures \ref{figure_toy_latents_xi} and
\ref{figure_nongaussian_latents}. A large $\sigma^2$ leads to a more
widely distributed prior which in turn is truncated more strongly by
the likelihood where it disagrees with the target class.
In the previous comparison, the hyperparameter values were chosen to
produce non-Gaussian marginal posterior distributions for
demonstration purposes.
However, usually the hyperparameters are estimated by maximizing the
marginal likelihood.
\citet{kuss2005} and \citet{nickisch2008} studied the suitability of
the marginal likelihood approximations for selecting hyperparameters
in binary classification.
They compared the calibration of predictive performance and the
marginal likelihood estimates on a grid of hyperparameter values.
In the following, we extend these comparisons to multiple classes
with the USPS dataset, for which similar considerations were done by
\citet{rasmussen2006} with the LA
method.

The upper row of Figure \ref{figure_surfaces} shows the log marginal
likelihood approximations for EP, IEP, and LA, and the lower bound on
evidence for VB as a function of the log-lengthscale $\log(l)$ and
log-magnitude $\log(\sigma^2)$ using the USPS 3 vs. 5 vs. 7 data.
The middle row shows the log predictive densities evaluated on the
test set, and the bottom row shows the corresponding classification
accuracies. The marginal likelihood approximations and predictive
densities for EP and IEP appear to be similar, but the maximum contour
of the log marginal likelihood for IEP (the contour labeled with -166
in plot (b)) does not coincide with the maximum contour of the
predictive density (the contour labeled with -76.5 in plot (f)), which
is why
a small bias can occur if the approximate marginal likelihood is used
for selecting hyperparameters. With EP there is a good agreement
between the maximum values in plots (a) and (e), and overall, the log
predictive densities are higher than with the other approximations.
The log predictive densities of VB and LA are small where
$\log(\sigma^2)$ is large (regions where $q(\f|\dataset,\theta)$ is likely to
be non-Gaussian), but also the marginal likelihood approximations
favor the areas of smaller $\log(\sigma^2)$ values.

There is a reasonable agreement with the marginal likelihood
approximations and classification accuracies with EP and IEP, although
the maximum accuracies are slightly lower than with VB and LA. The
maximum accuracies are very high with VB, but the region of the
highest accuracy does not agree with the region of the highest
estimate of the marginal likelihood.  With LA the marginal likelihood
estimate is calibrated better with the classification accuracy, but
the performance worsens when the posterior distribution is skewed with
large values of $\log(\sigma^2)$.

\subsection{Computational complexity and convergence}

In this section we consider the computational complexities of the approximate 
methods for one iteration with fixed hyperparameter values. Note that the 
following discussion is only approximate, and the practical efficiency of the 
algorithms depends much on implementations and the choices of convergence criteria.

Table \ref{table_complexities} summarizes the approximate scaling of
the number of computations as a function of $n$ and $c$. EP and IEP 
refer to the fully coupled and class-independent approximations, respectively, determined with 
the proposed nested EP algorithm. QIEP refers to the quadrature-based 
class-independent approximation proposed by \citet{seeger2006} and 
MCMC refers to Gibbs sampling with the multinomial probit model.
The first row (Posterior) describes the overall scaling of the mean
and covariance calculations related to the approximate conditional
posterior of $\f$.
The base computational cost resulting from the full GP prior scales as
$\mathcal{O}(n^3)$ due to the $n\times n$ matrix inversion (in
practice computed using Cholesky decomposition), which is required $c$
times for IEP, QIEP, and MCMC, and one additional time for EP and LA due to 
incorporation of the between-class correlations.
If the same prior covariance structure is used for all classes, VB has
the lowest cost, because only one matrix inversion is required per
iteration.

The second row (Likelihood) approximates the scaling of the number of
calculations that are required besides the posterior mean and
covariance evaluations (mainly likelihood related computations for one
iteration).
For both EP and IEP, this row describes the scaling of the computations 
needed for the tilted moment approximations done with inner EP algorithm. 
For QIEP the second row summarizes the number of one- and two-dimensional 
numerical quadratures (denoted by $n_{\textrm{q}}$ and $n_{\textrm{q}}^2$ respectively) 
required for the tilted moment evaluations under the 
independence assumption, and for LA the number of calculations required 
for evaluating the first and second order derivatives of the softmax likelihood. 
Each VB iteration
requires evaluating the expectations of the auxiliary variables either
by a quadrature or sampling, and the cost of one such operation is
denoted by $n_{\mathrm{q}}$ (for example, the number of quadrature design
points). Gibbs sampling with the multinomial probit likelihood
requires drawing from the conic truncation of a $c$-dimensional
normal distribution for each observation, and the cost of one draw is denoted 
by $n_{\mathrm{s}}$. The QIEP solution can be implemented efficiently because 
same function evaluations can be utilized in all of the $2c-1$ one-dimensional 
quadratures and the number of two-dimensional quadratures does not depend on $c$.
The cubic scaling in $c-1$ of the tilted moment evaluations in the nested EP and IEP
algorithms can be alleviated by reducing the number of inner-loop iterations 
$n_{\textrm{in}}$ as discussed in Section 3.2.3.

\begin{table}[!t]
  \caption{Approximate computational complexities of the
    various methods as a function of $n$ and $c$ for one iteration with fixed hyperparameters. The
    first row summarizes the scaling of the mean and covariance
    calculations related to the approximate conditional posterior of
    $\f$. The second row approximates the scaling of the number of
    calculations required for additional likelihood related
    computations. Parameter $n_{\mathrm{in}}$ refers to the number
    of inner EP iterations in the nested EP, $n_{\mathrm{q}}$ and $n_{\mathrm{q}}^2$ 
    to the cost of one- and two-dimensional numerical quadratures respectively,
    and $n_{\mathrm{s}}$ to the cost of sampling from a conic truncation of a
    $c$-variate Gaussian distribution.}
\label{table_complexities}
\begin{center}
  \begin{tabular}{ l | c | c | c | c | c | c }
    & EP & IEP & QIEP & VB & LA & MCMC \\
    \hline
    Posterior & $(c+1)n^3$ & $cn^3$ & $cn^3$ & $n^3$ & $(c+1)n^3$ & $cn^3$\\
    Likelihood & $nn_{\mathrm{in}} (c-1)^3$ & $nn_{\mathrm{in}} (c-1)^3$ &
    $n( (2c-1)n_{\mathrm{q}} $ & $n(c-1) 2n_{\mathrm{q}}$& $nc$ & $n cn_{\mathrm{s}}$ \\
    & & & $+2n_{\mathrm{q}}^2 )$ & & &
\end{tabular}
\end{center}
\end{table}

Using the USPS 3 vs. 5 vs. 7 dataset, we measured the CPU time
required for the posterior inference on $\f$ given nine different
preselected hyperparameter values from the grid of Figure
\ref{figure_surfaces}. With our implementations, LA was the fastest, and
EP and VB were about three times more expensive than LA.
Because of the efficient scaling (Table \ref{table_complexities}), VB
should be much faster, and probably closer to the running time of LA.
One reason for the slow performance may be our implementation based on
importance sampling steps, which may result in slower convergence due
to fluctuations.
The MCMC and LA-TKP approaches were overall very slow compared to LA.
%
One iteration of MCMC is relatively cheap, but in our experiments thousands
of posterior samples were required to obtain chains of sufficiently uncorrelated
samples which is why MCMC was over hundred times slower than LA.
LA-TKP requires roughly $c+1$ times the CPU time of LA for computing 
the predictions for each test input. Therefore, the computational cost 
of LA-TKP becomes quickly prohibitive as the number of test input increases.

\begin{figure*}[!t]
 \centering
  \subfigure[]{\includegraphics[scale=0.33]{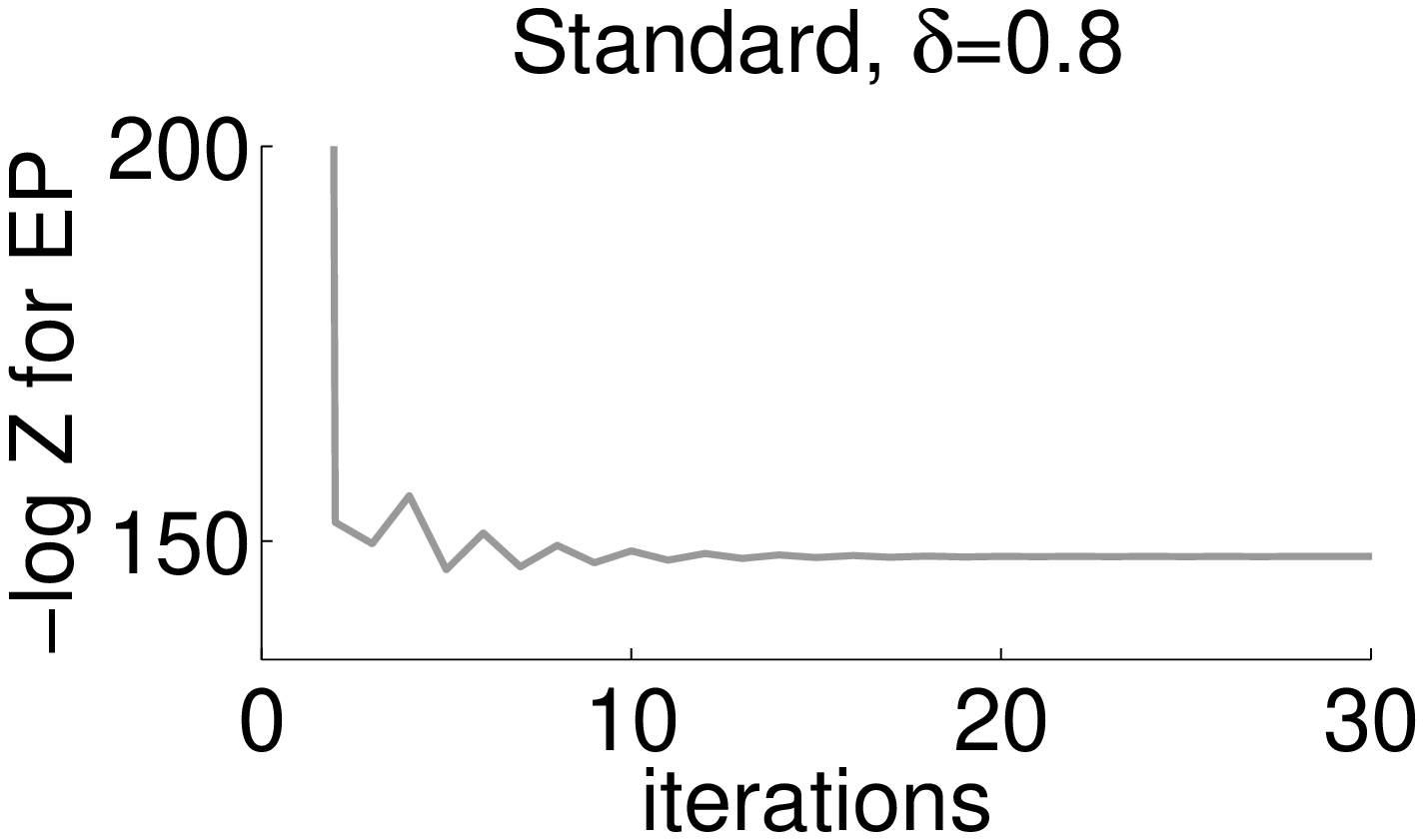}}
  \subfigure[]{\includegraphics[scale=0.33]{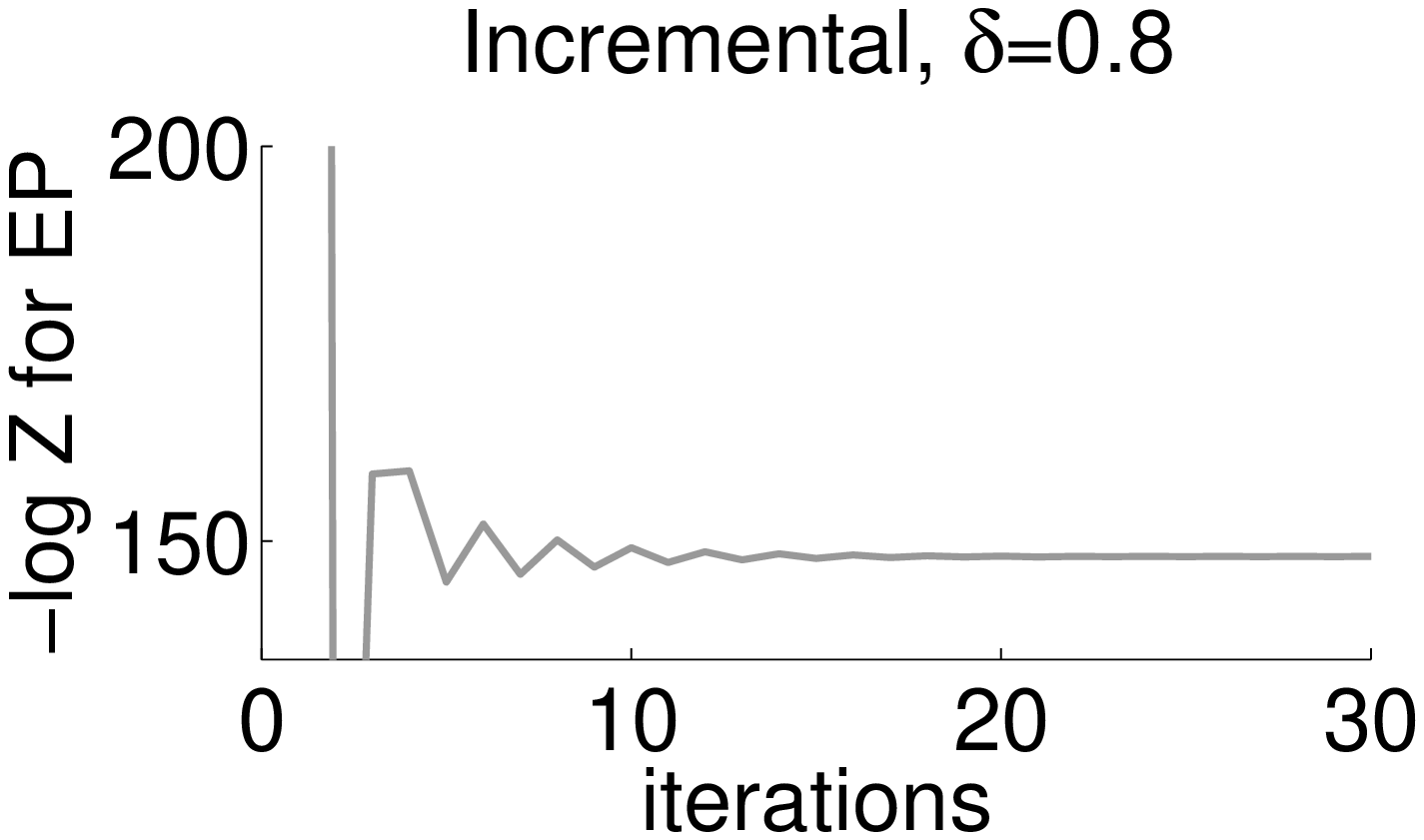}}
  \subfigure[]{\includegraphics[scale=0.33]{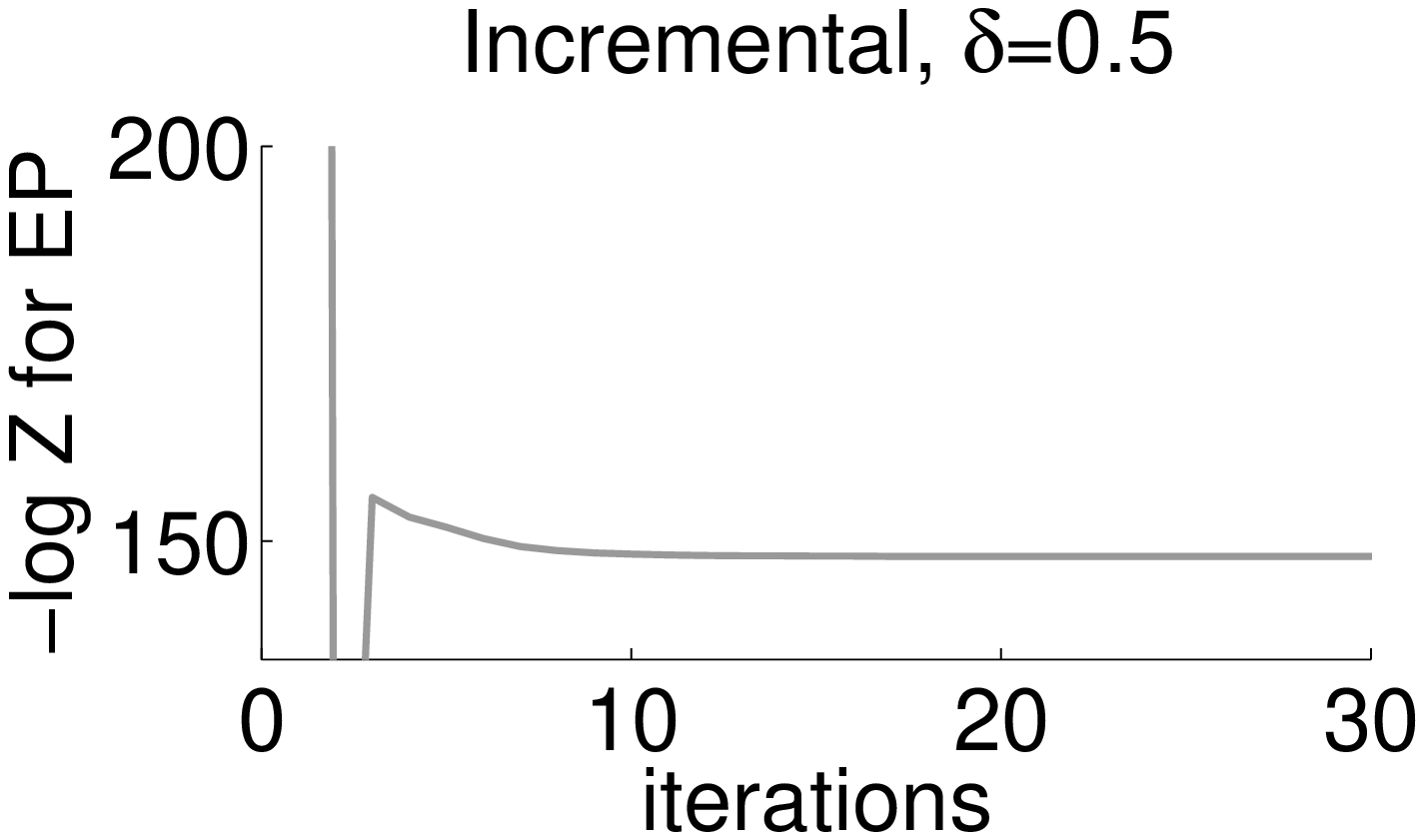}}
  \subfigure[]{\includegraphics[scale=0.33]{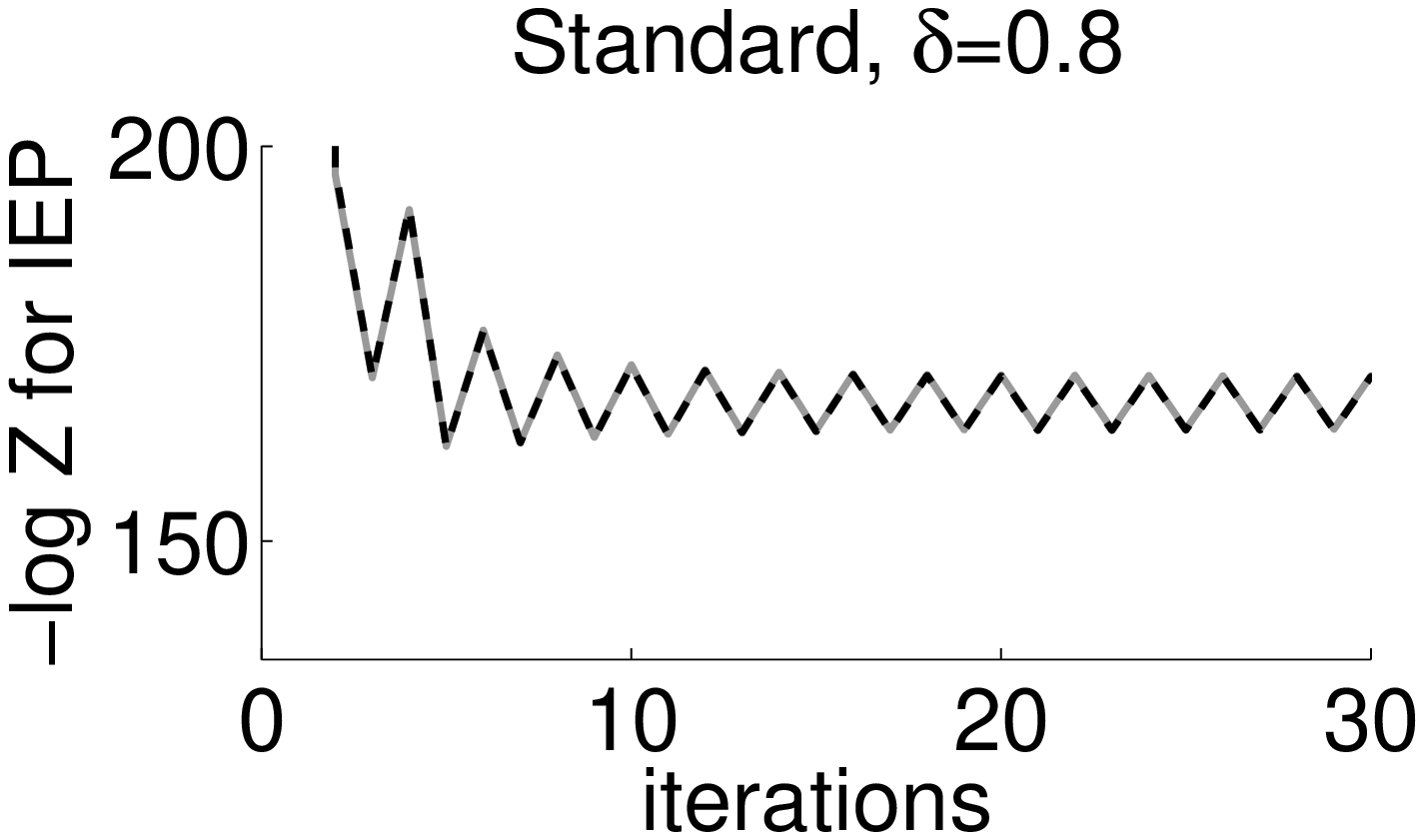}}
  \subfigure[]{\includegraphics[scale=0.33]{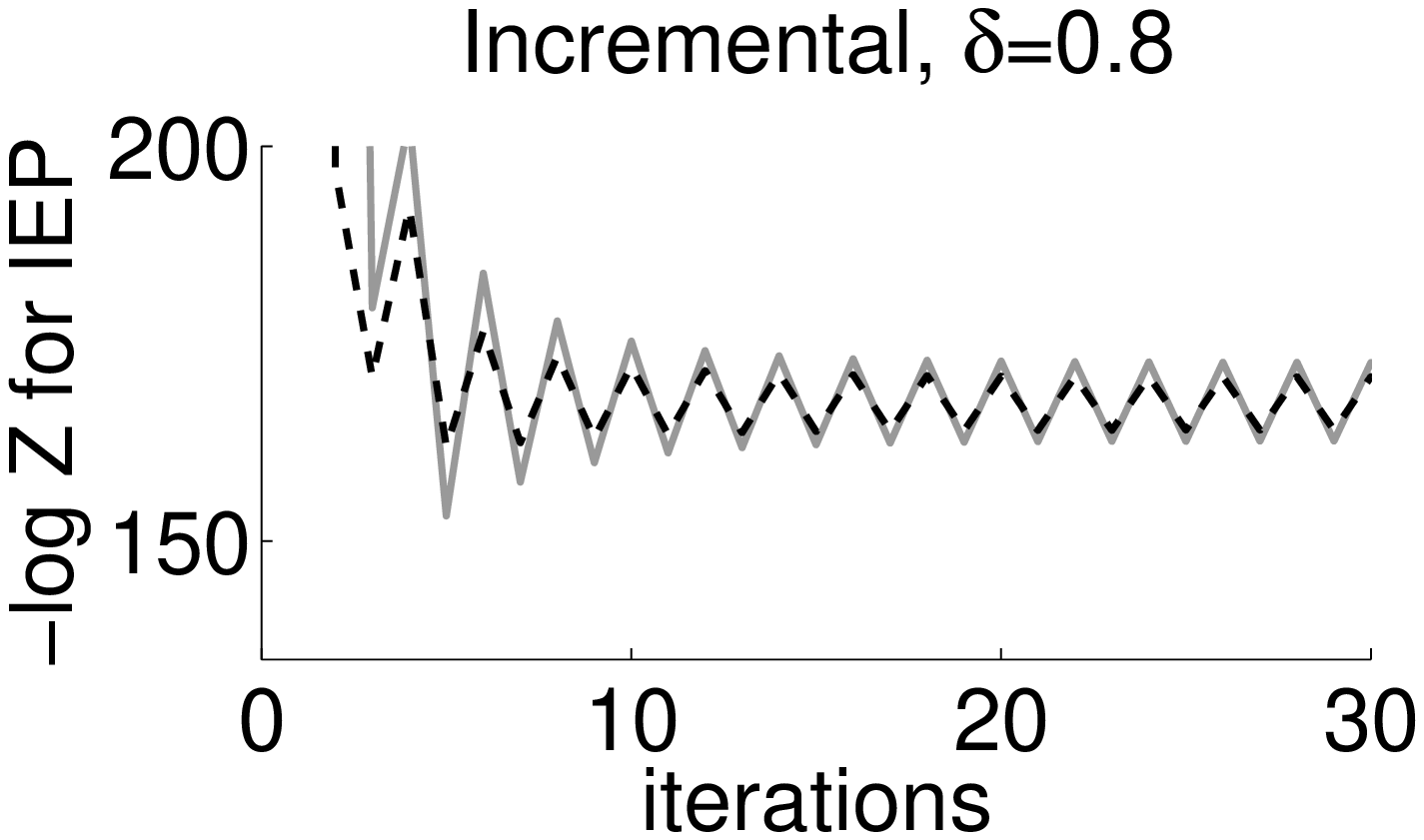}}
  \subfigure[]{\includegraphics[scale=0.33]{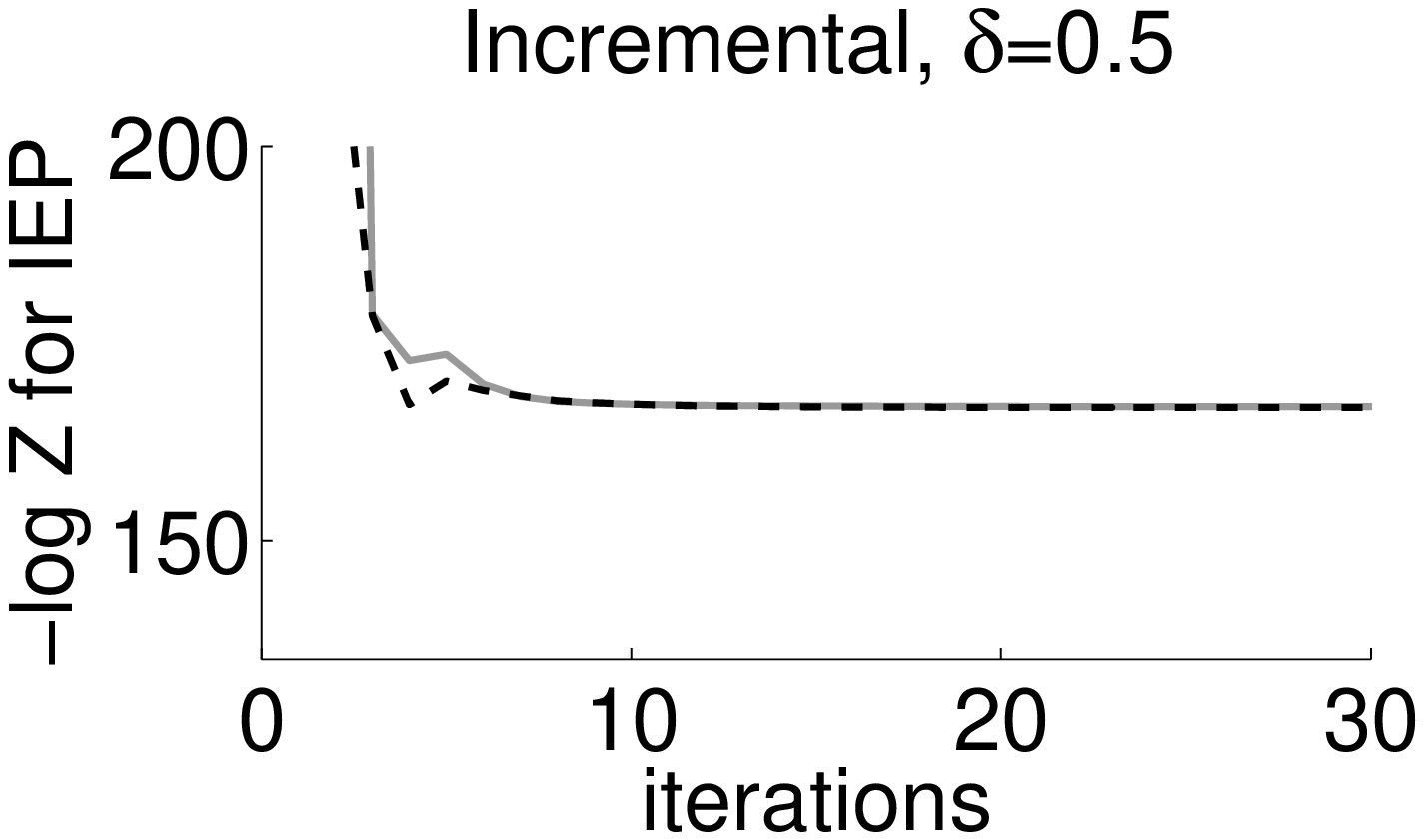}}
  \subfigure[]{\includegraphics[scale=0.33]{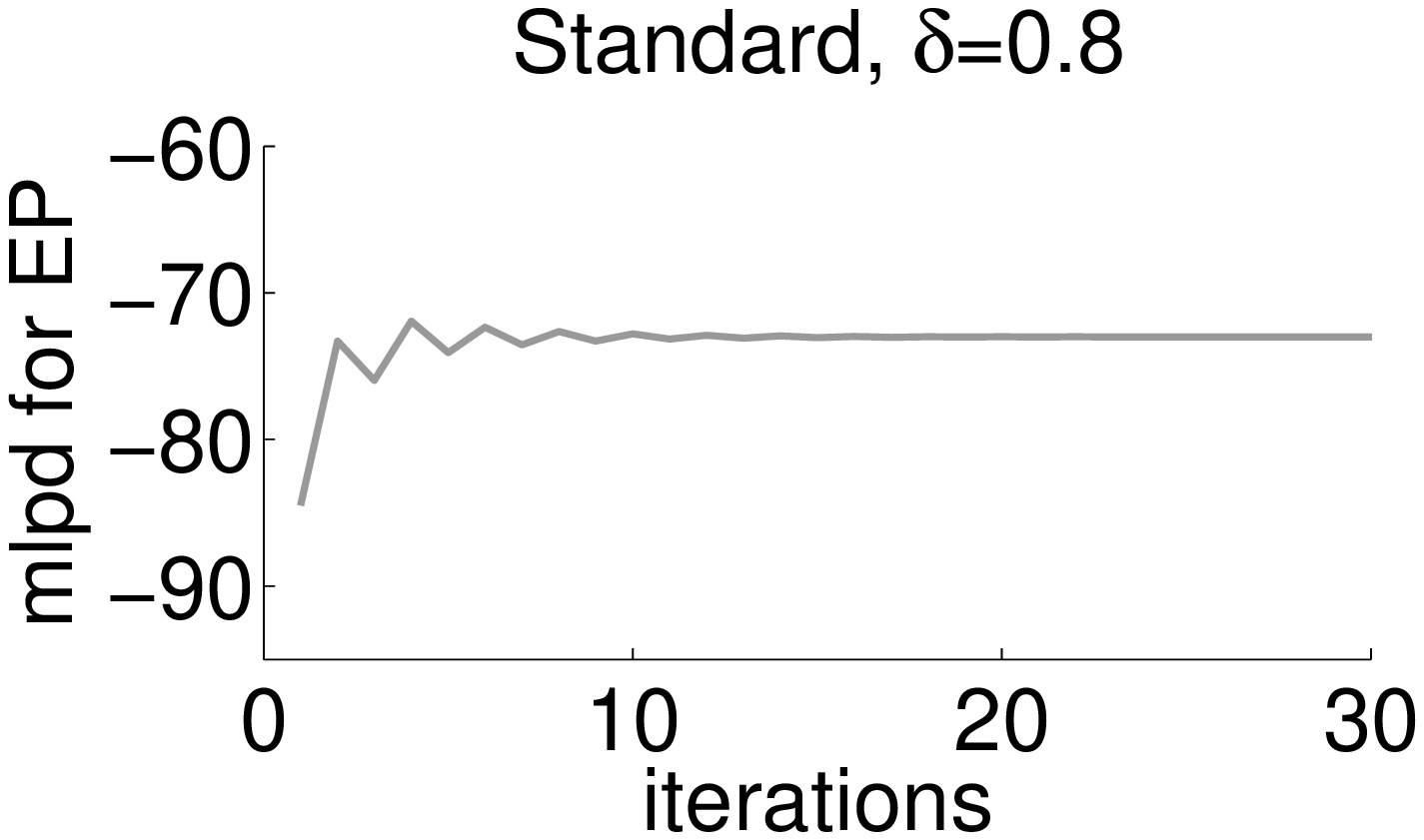}}
  \subfigure[]{\includegraphics[scale=0.33]{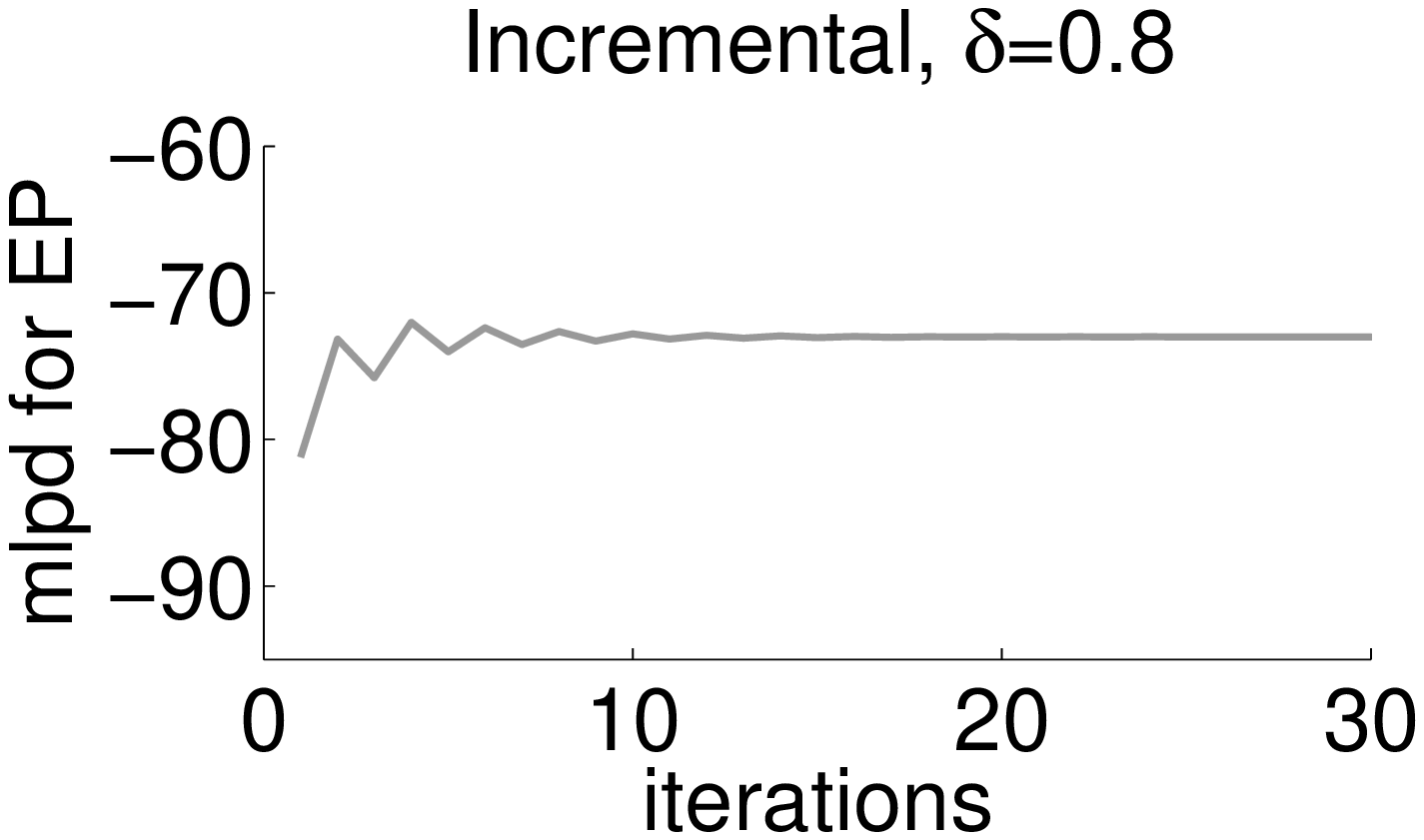}}
  \subfigure[]{\includegraphics[scale=0.33]{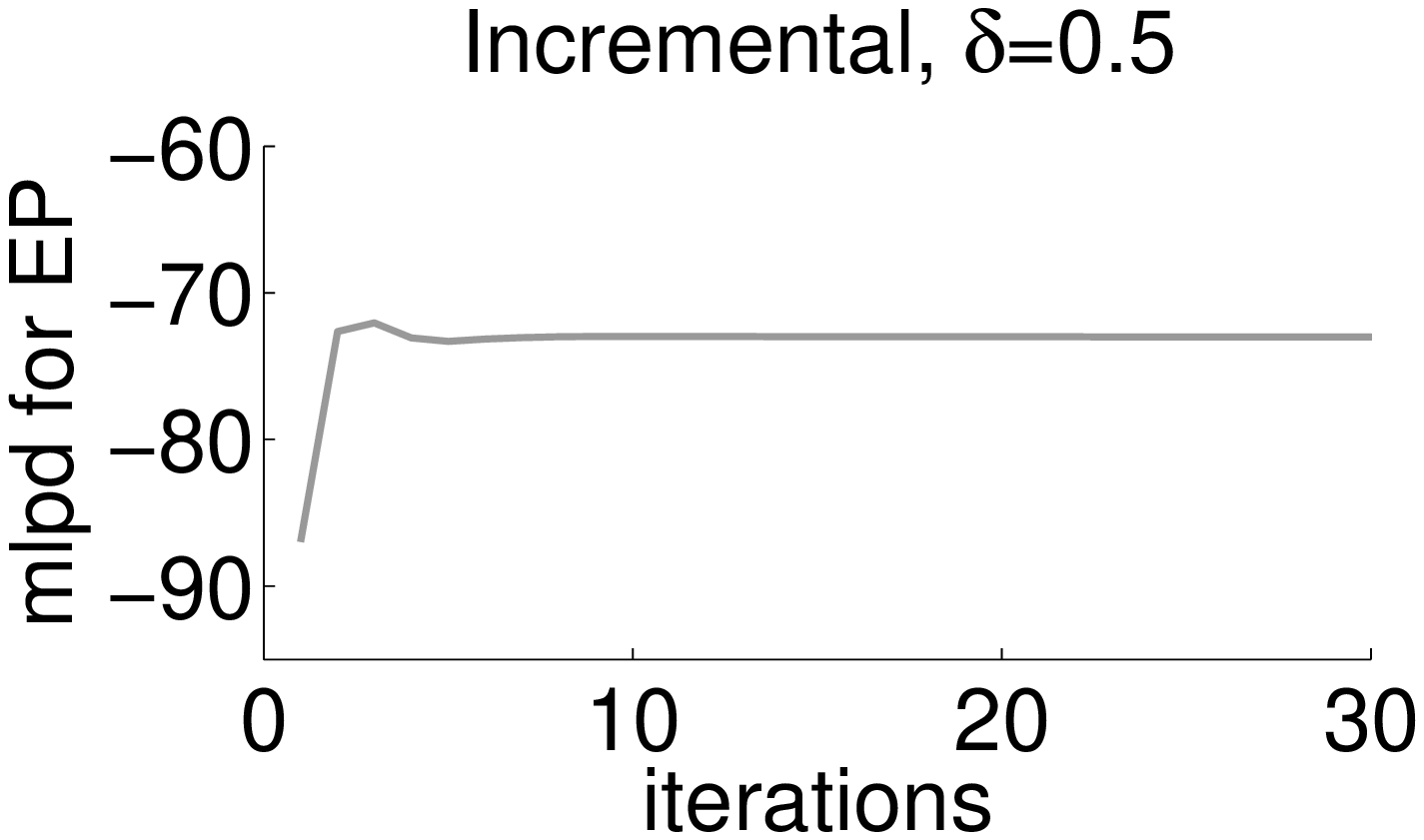}}
  \subfigure[]{\includegraphics[scale=0.33]{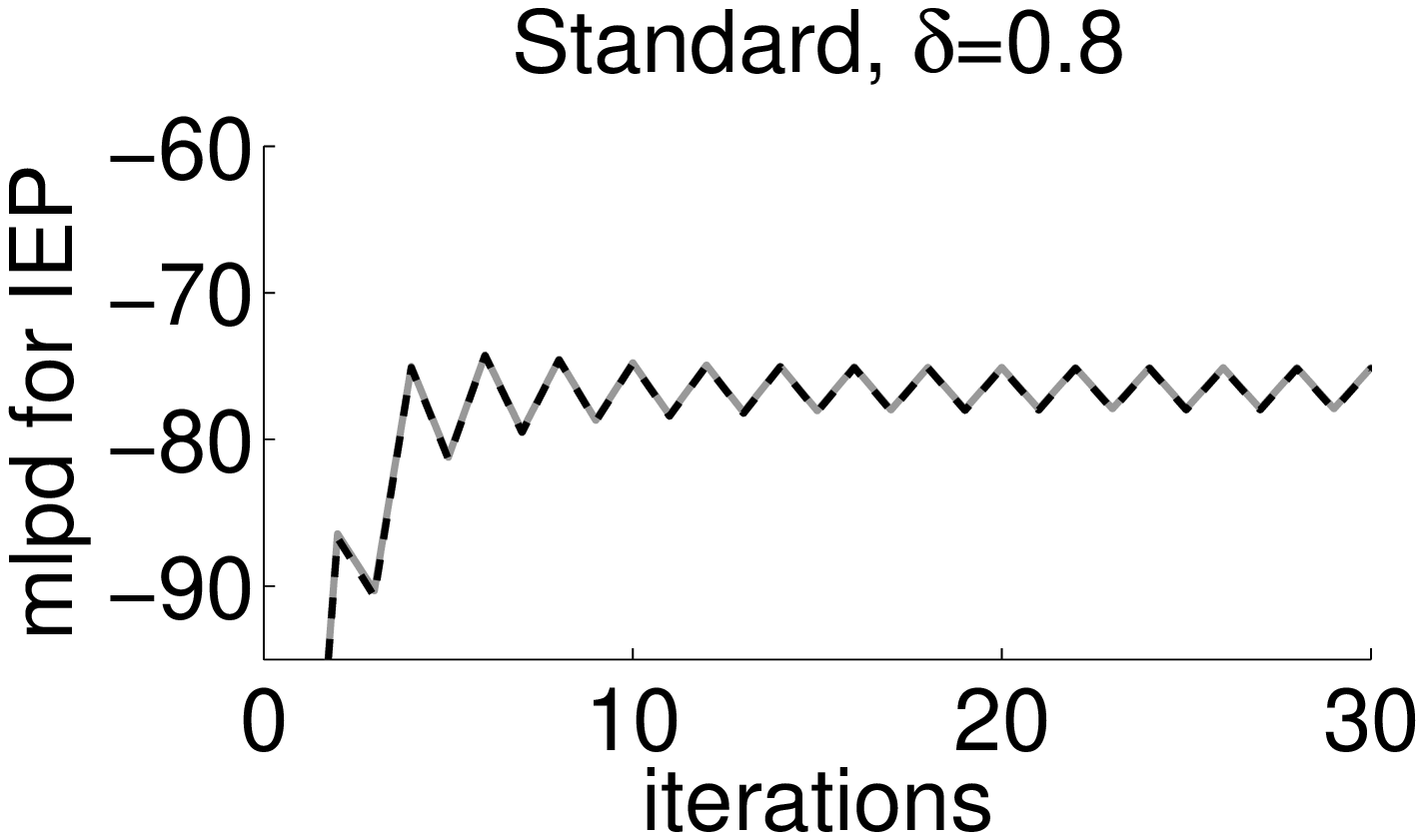}}
  \subfigure[]{\includegraphics[scale=0.33]{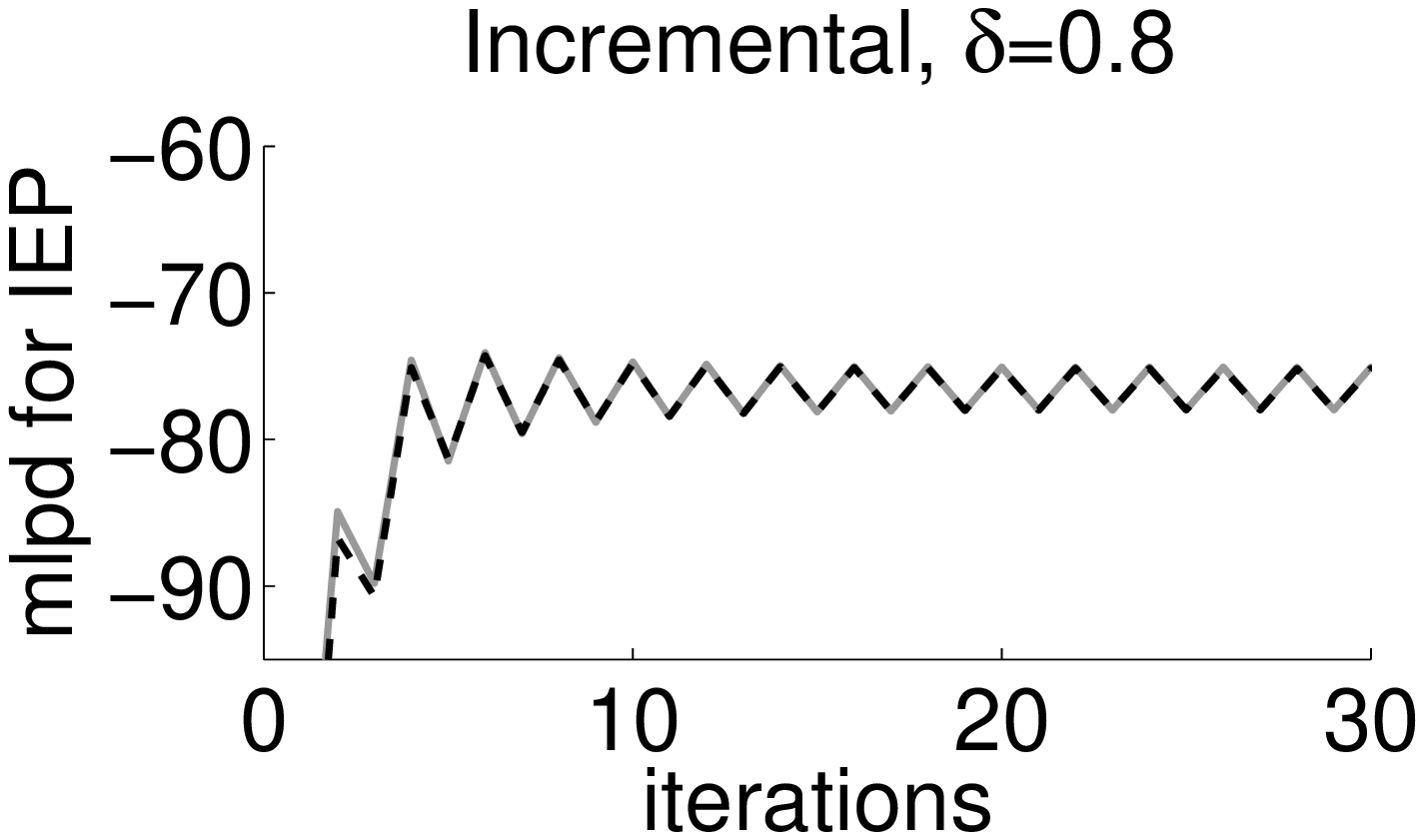}}
  \subfigure[]{\includegraphics[scale=0.33]{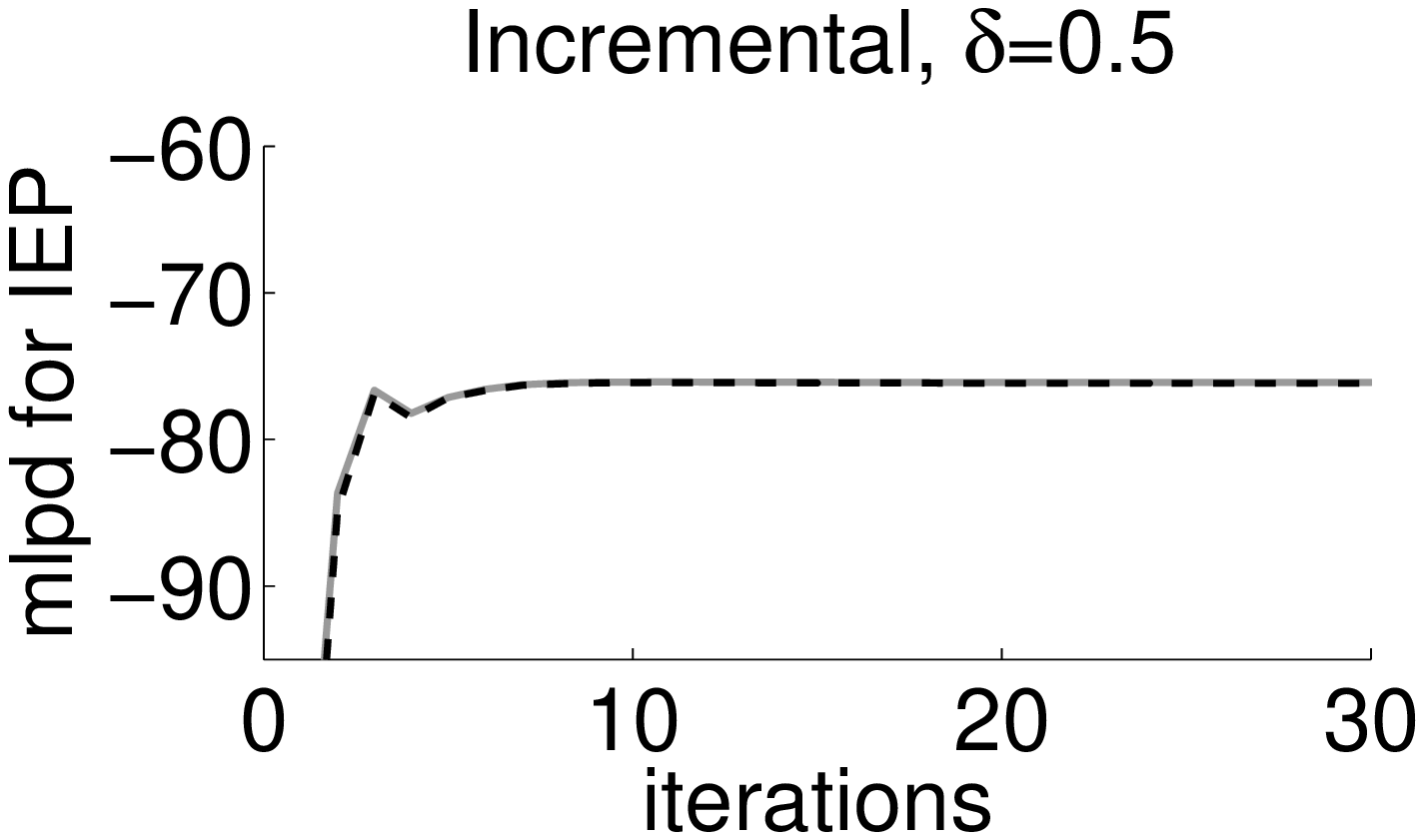}}
  \caption{A convergence comparison between EP and IEP using parallel
    updates in the outer EP loop with the USPS 3 vs. 5 vs. 7 data. 
    Plots (a)-(f) show the negative log marginal likelihood estimates $-\log Z_{\mathrm{EP}}$ 
    as a function of iterations for two different damping factors
    $\delta$, and plots (g)-(l) show the corresponding
    mean log predictive density evaluated using a separate test dataset.
    The hyperparameters of the squared exponential covariance function
    were fixed at $\log(\sigma^2)=8$ and $\log(l)=2.5$. 
    In the first column (Standard) the inner-loops of the nested EP (and IEP) algorithms 
    (the solid gray lines) are run until convergence at each outer-loop iteration, 
    whereas in the other columns (Incremental) only one inner-loop iteration 
    per site is done at each outer-loop iteration. 
    For comparison, the dashed black lines show also the IEP solution obtained with numerical
    quadratures.}
  \label{figure_convergence}
\end{figure*}

In the CPU time comparisons across the range of hyperparameter values 
producing variety of skewed and non-isotropic posterior distributions, fully coupled 
nested EP converged in fewer outer-loop iterations than nested IEP if the same 
convergence criteria were used.
Figure \ref{figure_convergence} illustrates the difference with the 
hyperparameters fixed at $\log(\sigma^2)=8$ and $\log(l)=2.5$ which results 
in good predictive performances on the independent test dataset with both 
methods (see Figure \ref{figure_surfaces}).
Figure \ref{figure_convergence} shows the log marginal likelihood
approximation $-\log Z_{\mathrm{EP}}$ and the mean log
predictive density (mlpd) in the test dataset after each iteration for both approximations. 
Note that the converged EP approximation satisfying the moment matching conditions
between $\hat{p}(\f_i)$ and $q(\f_i)$ corresponds to stationary points
of an objective function similar to $-\log Z_{\mathrm{EP}}$
\citep{minka2001b,opper2005}.
The solid gray lines correspond to EP and IEP approximations obtained with 
the proposed nested EP algorithm, and for comparison, the dashed black lines
show also the quadrature-based QIEP solution.

The convergence is illustrated both with a small amount of damping, 
damping factor $\delta=0.8$ (the first and second columns Figure 7), and 
with a larger amount of damping $\delta=0.5$ (the last column). 
Note that with the fully coupled nested EP algorithm the damping is applied on the inner-EP site
parameters $\Taut_i$ and $\Nut_i$, whereas with IEP and QIEP the damping is
applied on the natural exponential site parameters 
$\tilde{\bm{\nu}}$ and $\tilde{T}_i=\tilde{\Sigma}_i^{-1}$ ($\tilde{T}_i$ diagonal).
In the first column (Standard) the inner-loops of the nested EP (and IEP) 
algorithms are run until convergence at each outer-loop iteration, whereas 
in the remaining columns (Incremental) only one inner-loop iteration per 
site is done at each outer-loop iteration.
Based on Table \ref{table_complexities} the incremental updates ($n_{\mathrm{in}}=1$) 
reduce notably the computational burden of the inner-loop of the nested EP algorithm
which scales as $\mathcal{O}( n_{\mathrm{in}} (c-1)^3 )$.

Figure 7 shows that with standard updates the nested IEP gives almost identical 
approximation compared to the QIEP. However, when incremental updates are used, 
some differences can be seen in the early iterations but both methods converge into 
the same solution. 
The full nested EP algorithm oscillates less than IEP or QIEP with the same 
damping level but this may be partly caused by the different parameterization. 
However, with both damping levels full EP seems to converge in fewer iterations 
whereas there is a slow drift in $- \log Z_{\mathrm{EP}}$ and the mlpd score 
even after 20 iterations with nested IEP and QIEP. One explanation for this behavior
can be the large magnitude hyperparameter value which results into strongly non-Gaussian 
posterior distribution and the relatively large lengthscale which induces stronger 
between-class posterior dependencies through the likelihood terms.

\subsection{Predictive performance across datasets with hyperparameter
  estimation}


In this section, we compare the predictive performances of nested EP, nested IEP,
VB, LA, LA-TKP, and Gibbs sampling with the multinomial probit (MCMC) 
with estimation of hyperparameters on various benchmark datasets. 
All methods are compared using the USPS 3 vs. 5 vs. 7 data,
and the following five UCI Machine Learning Repository datasets:
New-thyroid, Teaching, Glass, Wine, and Image segmentation.  The
comparisons are also done using the USPS 10-class dataset, but only
for EP, IEP, VB, and LA due to the large $n$. The main characteristics
of the datasets are summarized in Table \ref{table_datasets}.

\begin{table}[!t]
  \caption{Datasets used in the experiments.}
\label{table_datasets}
\begin{center}
\begin{tabular}{ l | c | c | c | c | c}
  Dataset  & $n_{\mathrm{train}}$ & $n_{\mathrm{test}}$ & Classes ($c$)
  & Covariates ($d$) & ARD \\
  \hline
  New-thyroid & 215 & 215 (Ten-fold CV) & 3 & 5 & yes\\
  Teaching   & 151 & 151 (Ten-fold CV) & 3 & 5 & yes \\
  Glass   & 214 & 214 (Ten-fold CV) & 6 & 9 & yes\\
  Wine    & 178 & 178 (Ten-fold CV) & 3 & 13 & yes\\
  Image segmentation & 210 & 2100 & 7 & 18 & no\\
  USPS 3 vs. 5 vs. 7 & 1157& 1175 & 3 & 256 & no\\
  USPS 10-class & 4649 & 4649 & 10 & 256 & no
\end{tabular}
\end{center}
\end{table}

For all the datasets, we standardize the covariates to zero mean and
unit variance, and use the squared exponential covariance function
with the same hyperparameters for all classes.
Only one lengthscale parameter is assumed for the Image segmentation
and USPS datasets, but otherwise individual lengthscale parameter
(Automatic Relevance Determination, ARD) is set for each input
dimension. We place a weakly informative prior on the lengthscale and
magnitude parameters by choosing a half Student-$t$ distribution with
four degrees of freedom and a variance equal to one hundred. With MCMC
we sample the hyperparameters, and with the other methods, we use
gradient-based type-II MAP estimation to select the hyperparameter
values.
%
The predictive performance is measured using a ten-fold
cross-validation (CV) with four of the datasets, and predetermined
training and test sets for three of the datasets (See Table
\ref{table_datasets}).

\begin{figure*}[!t]
  \centering
  \subfigure[New-thyroid]{\includegraphics[scale=0.22]{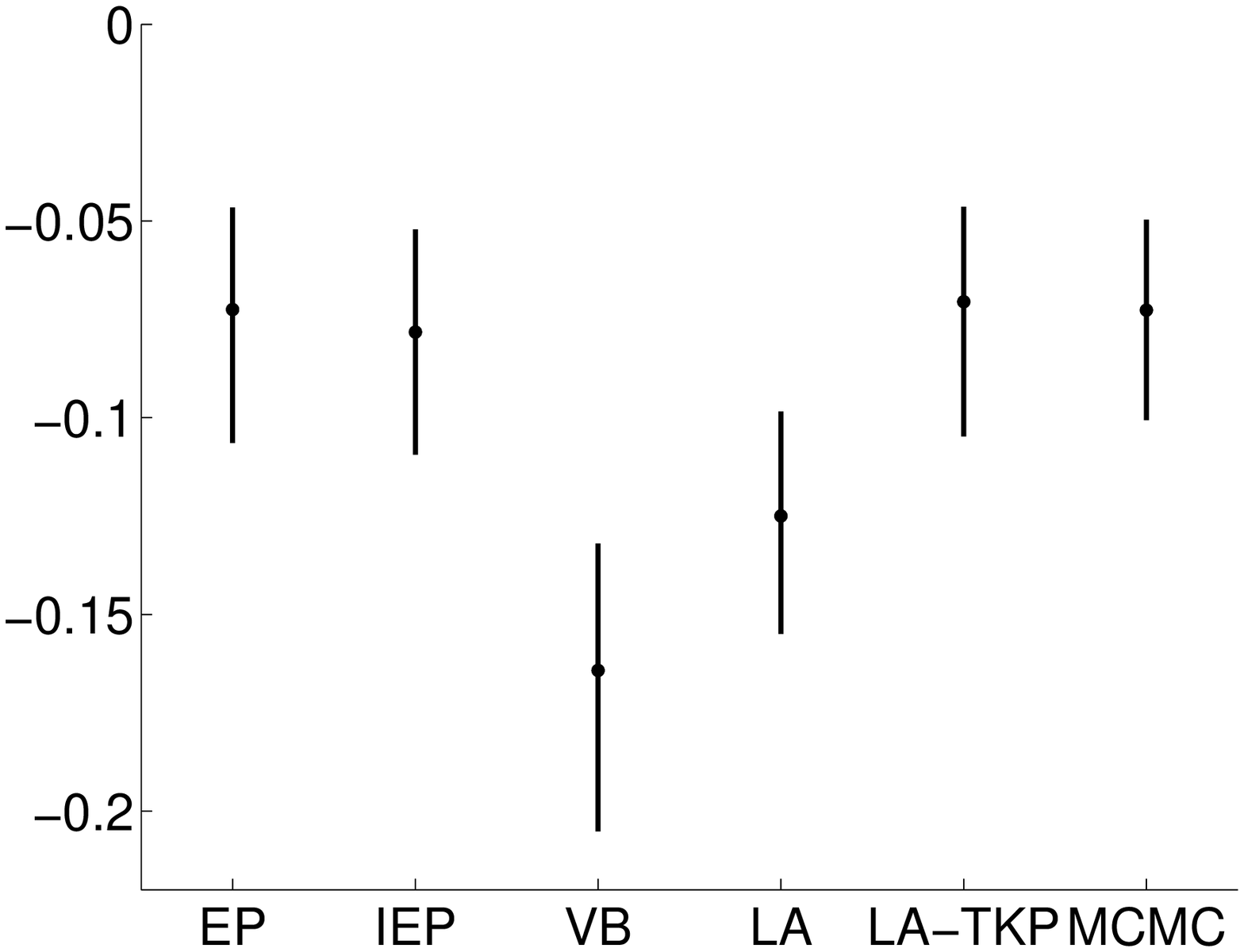}}
  \subfigure[Teaching]{\includegraphics[scale=0.22]{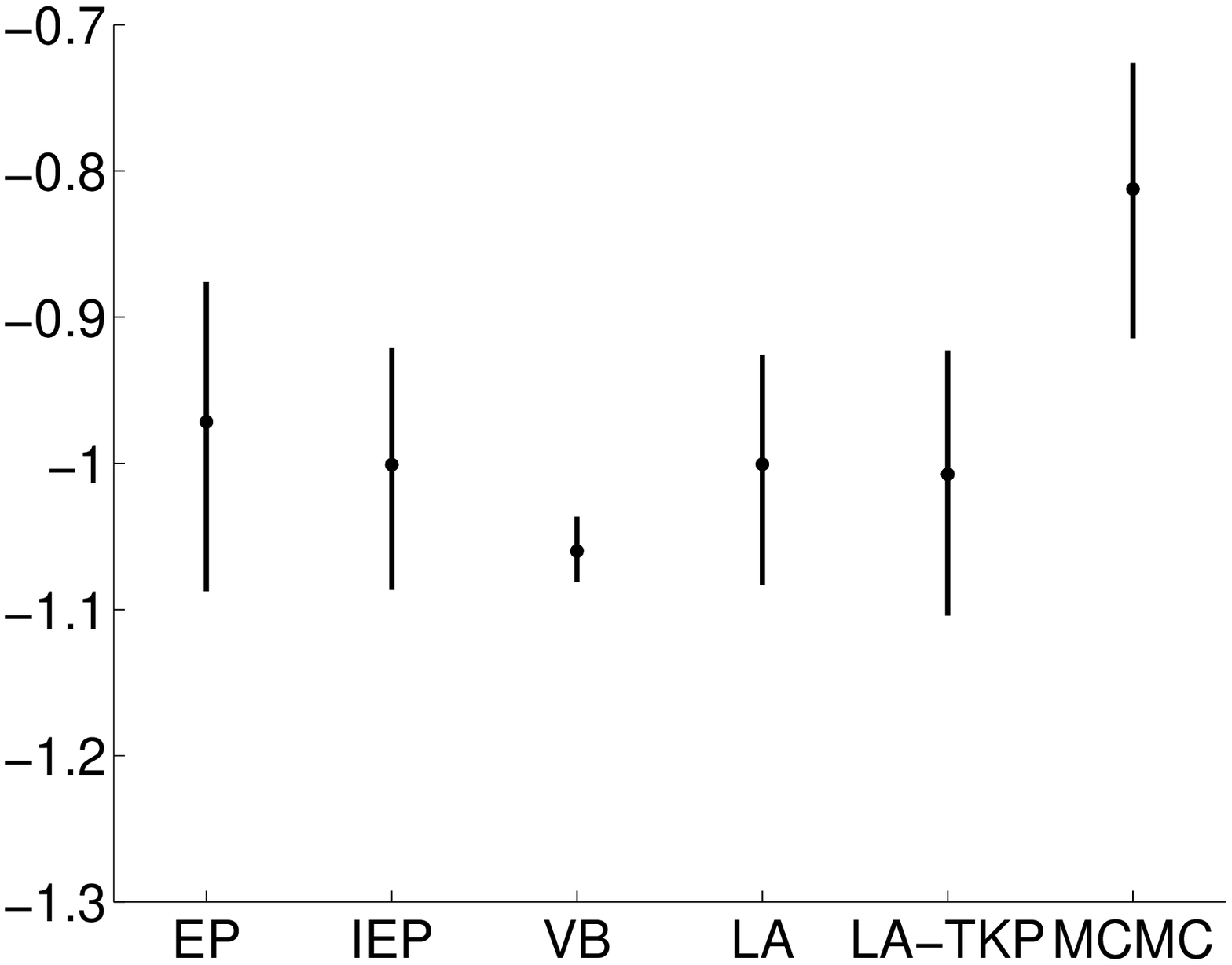}}
  \subfigure[Glass]{\includegraphics[scale=0.22]{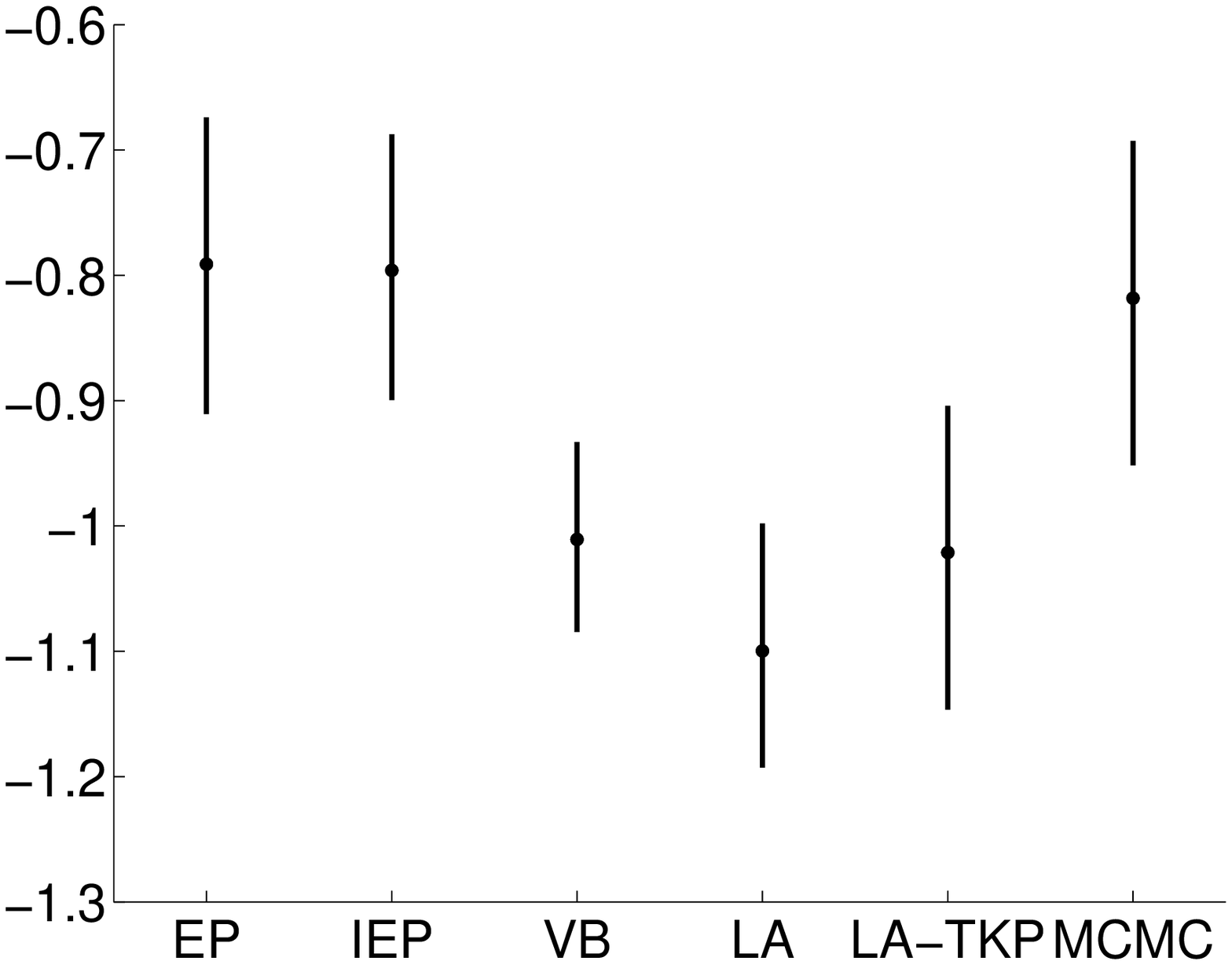}}
  \centering
  \subfigure[New-thyroid - pairwise]{\includegraphics[scale=0.22]{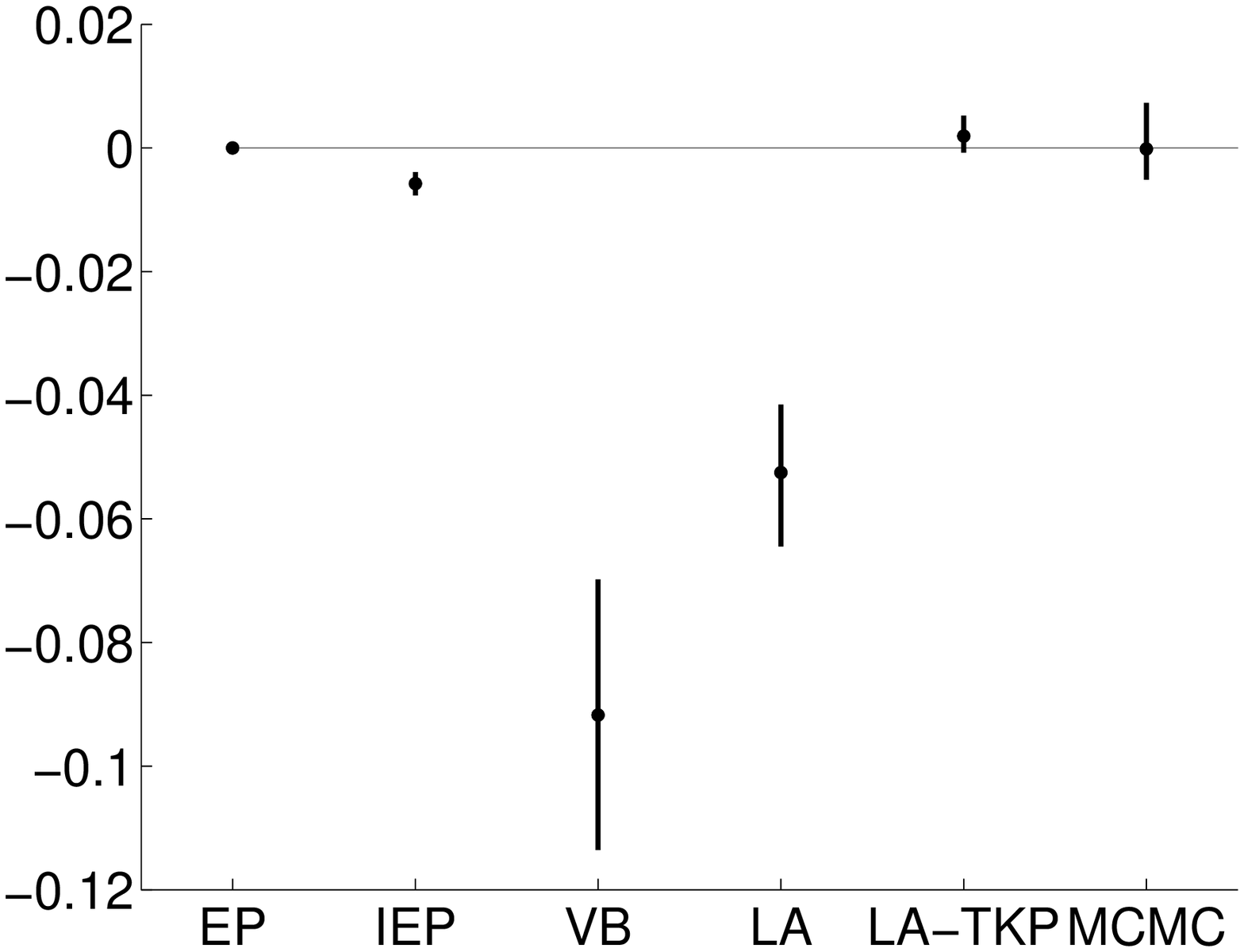}}
  \subfigure[Teaching - pairwise]{\includegraphics[scale=0.22]{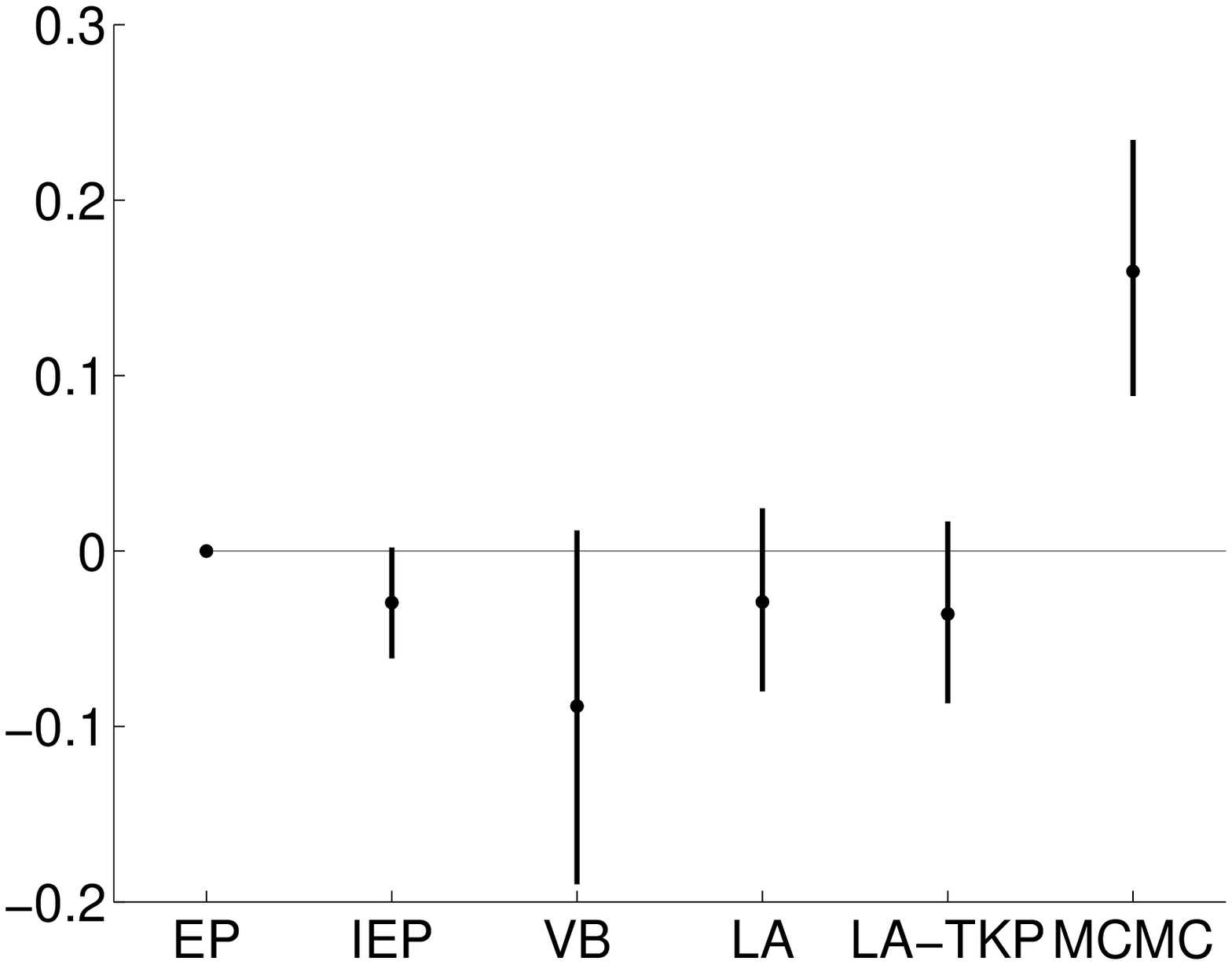}}
  \subfigure[Glass - pairwise]{\includegraphics[scale=0.22]{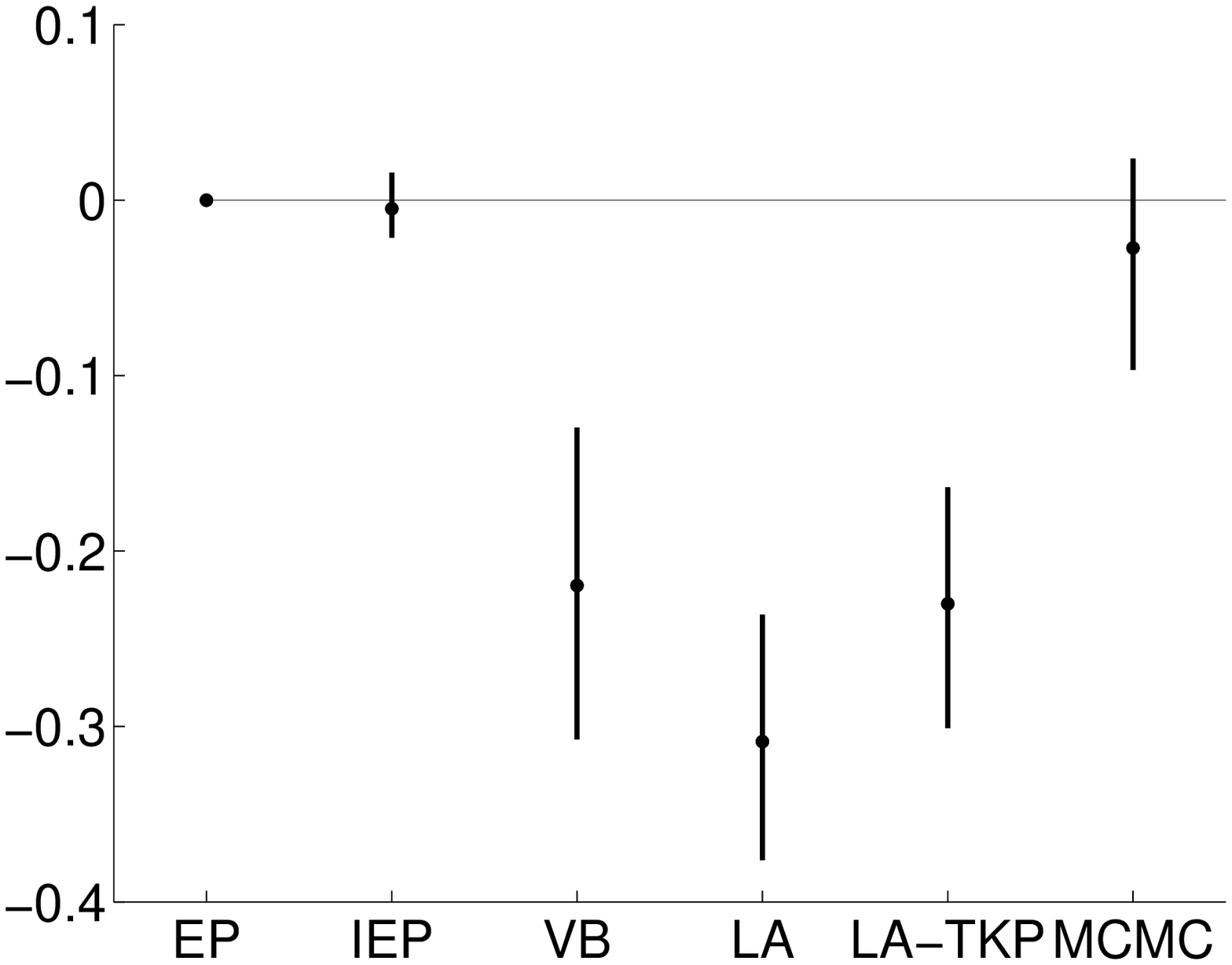}}
  \centering
  \subfigure[Wine]{\includegraphics[scale=0.22]{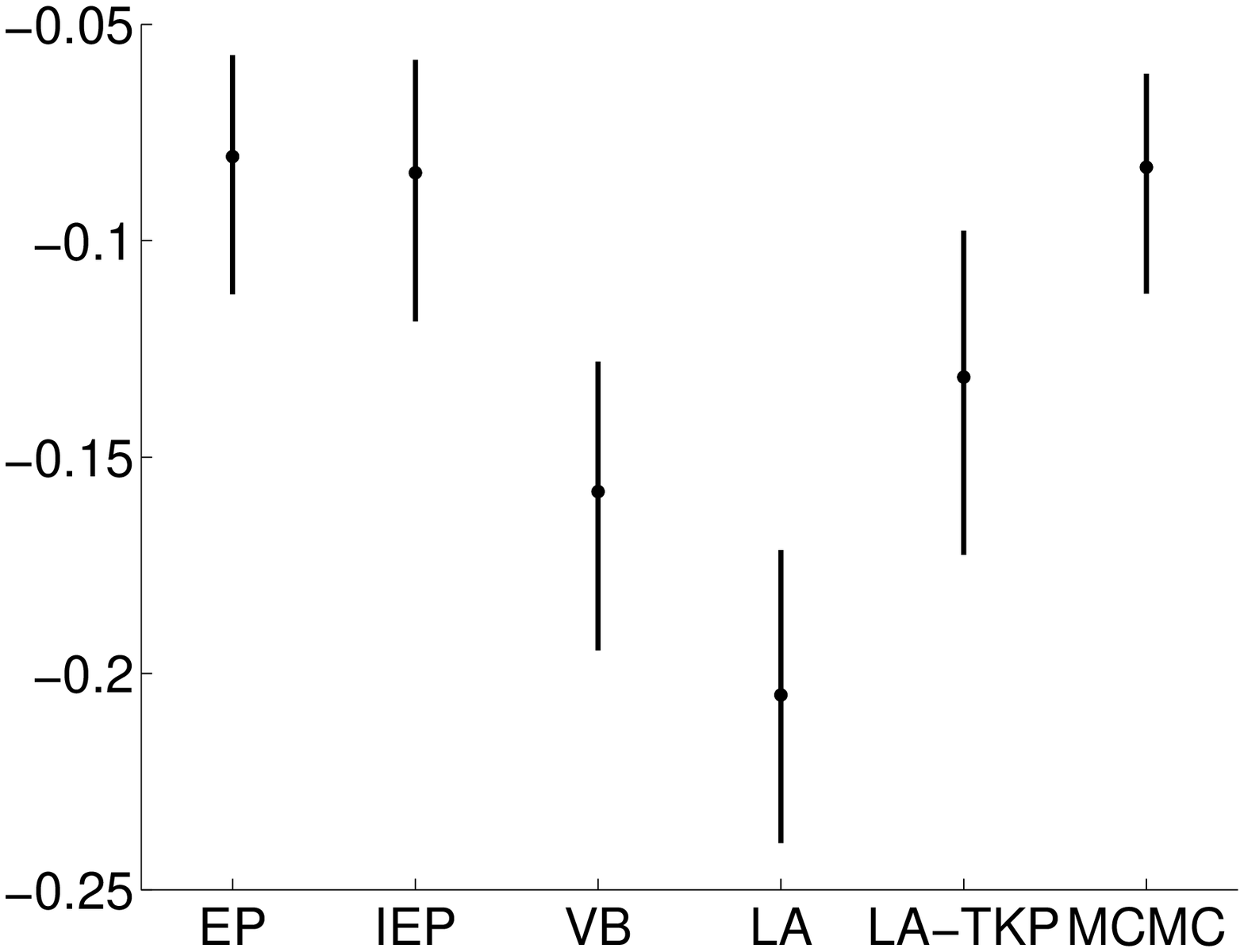}}
  \subfigure[Segmentation]{\includegraphics[scale=0.22]{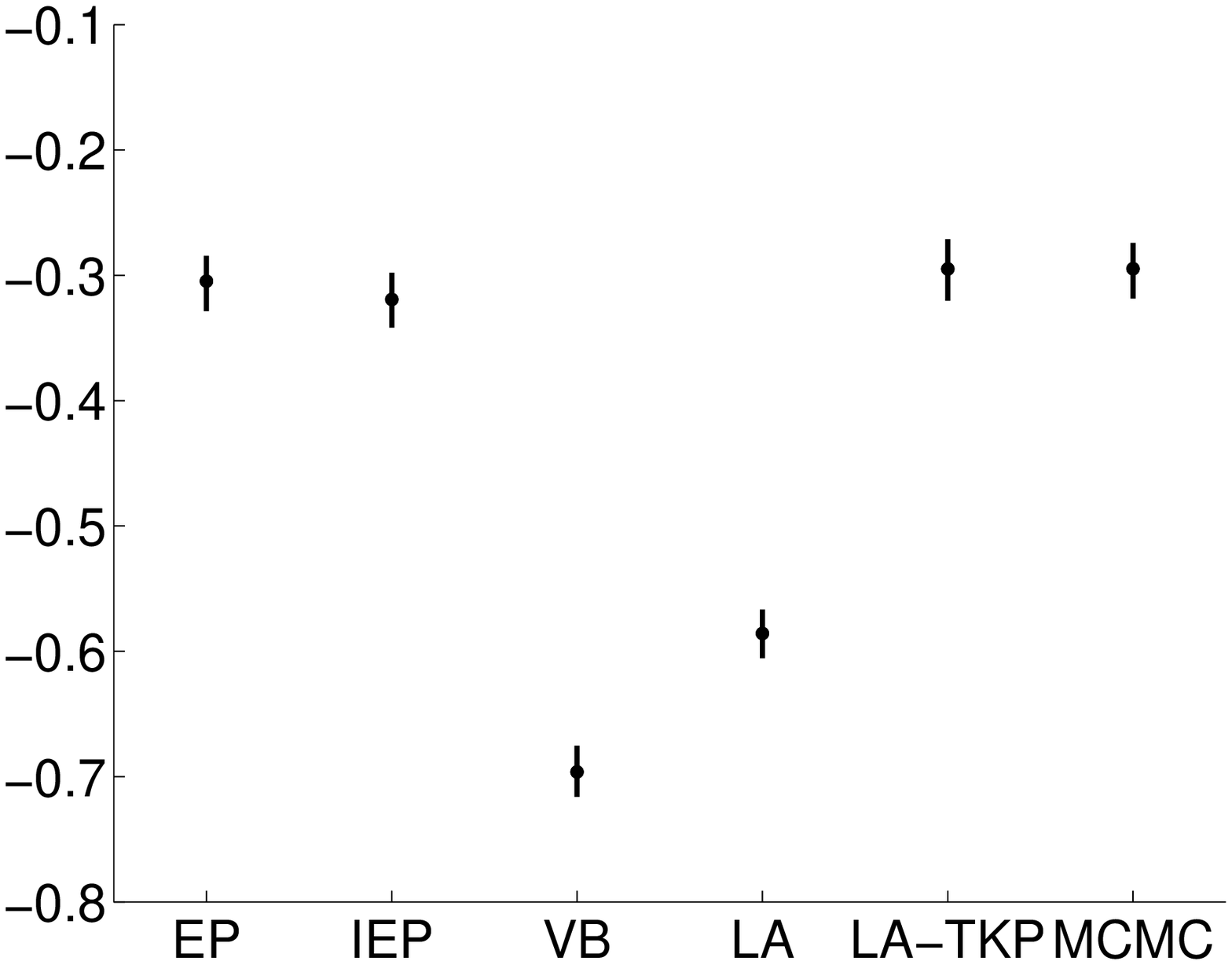}}
  \subfigure[USPS 3-5-7]{\includegraphics[scale=0.22]{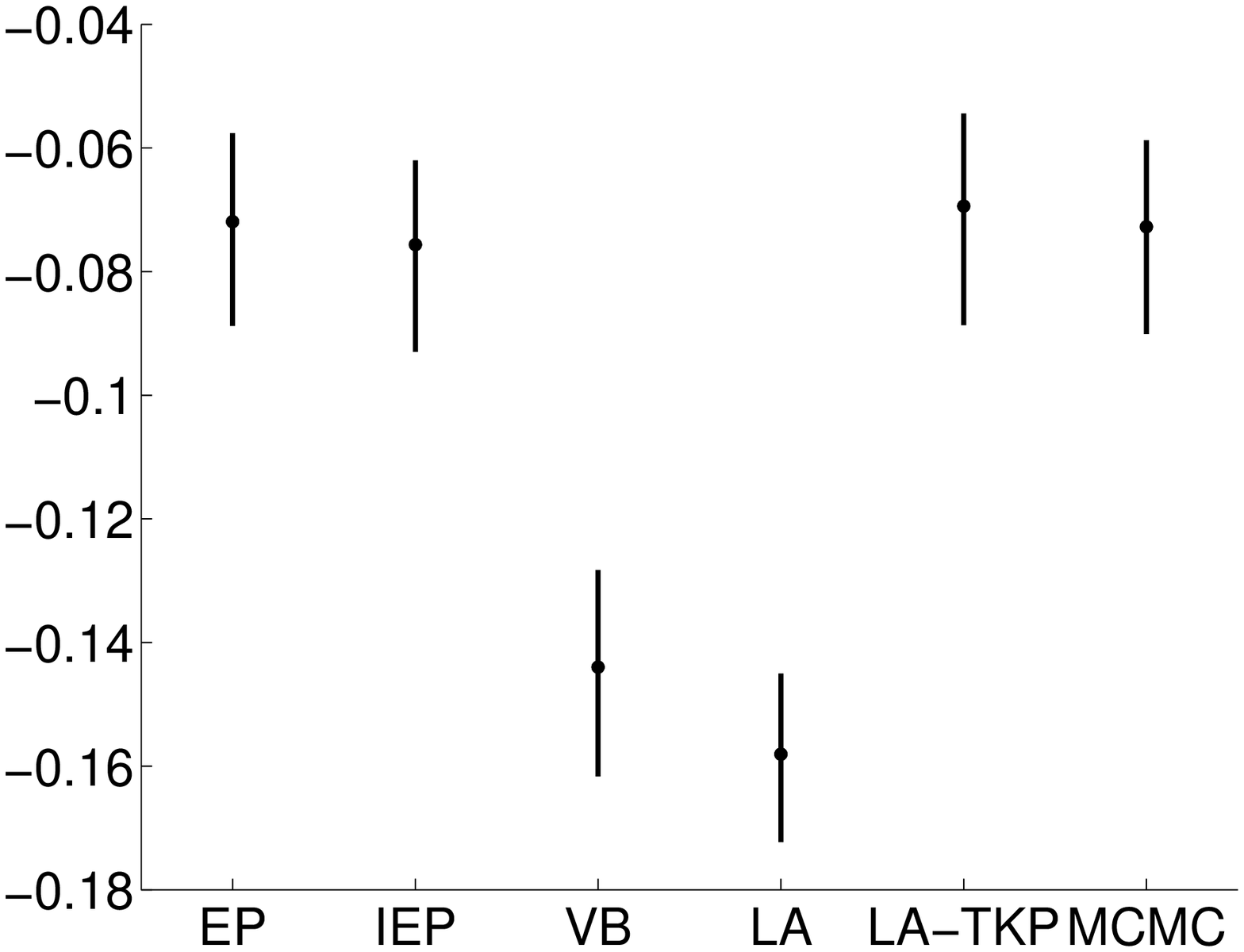}}
  \centering
  \subfigure[Wine - pairwise]{\includegraphics[scale=0.22]{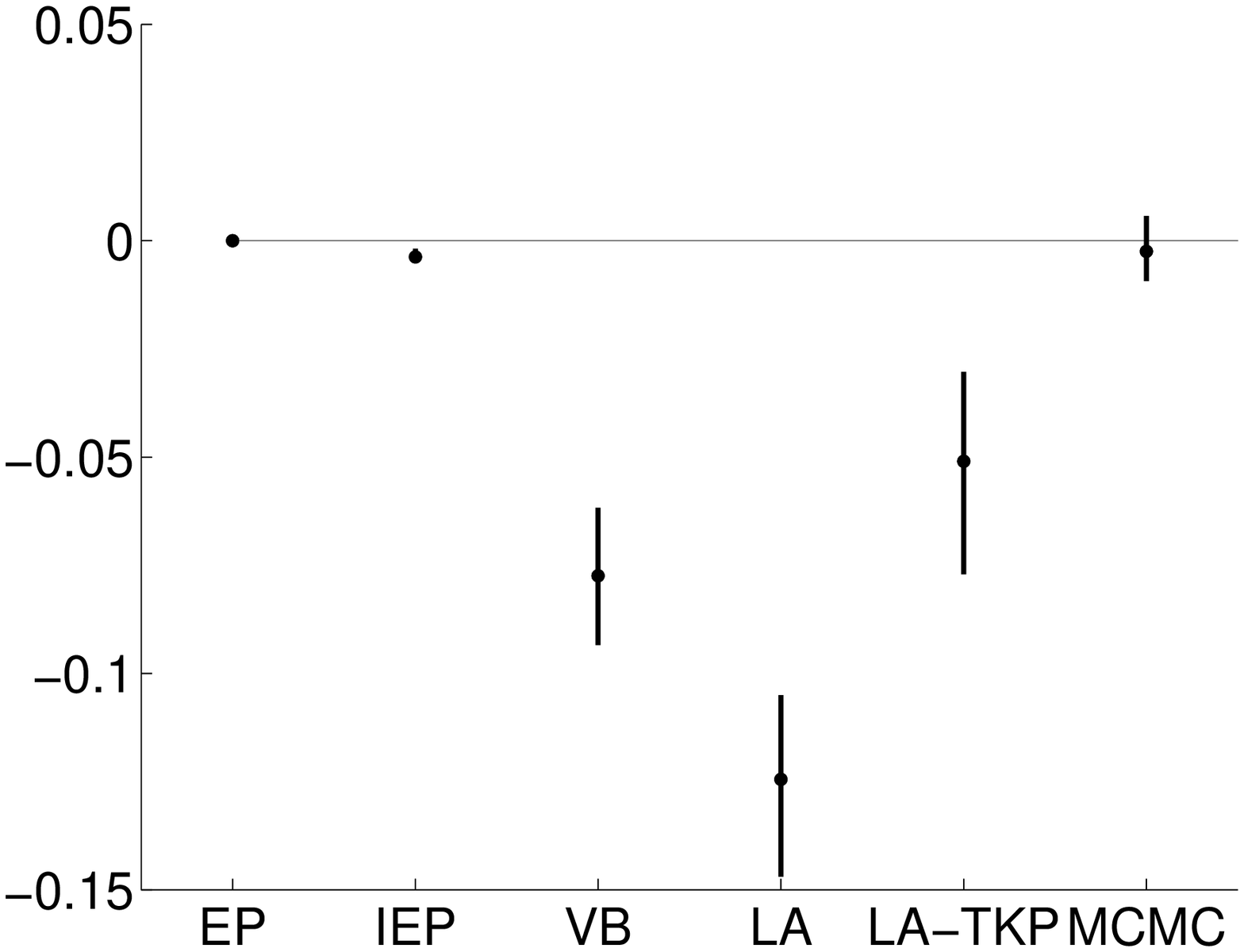}}
  \subfigure[Segmentation - pairwise]{\includegraphics[scale=0.22]{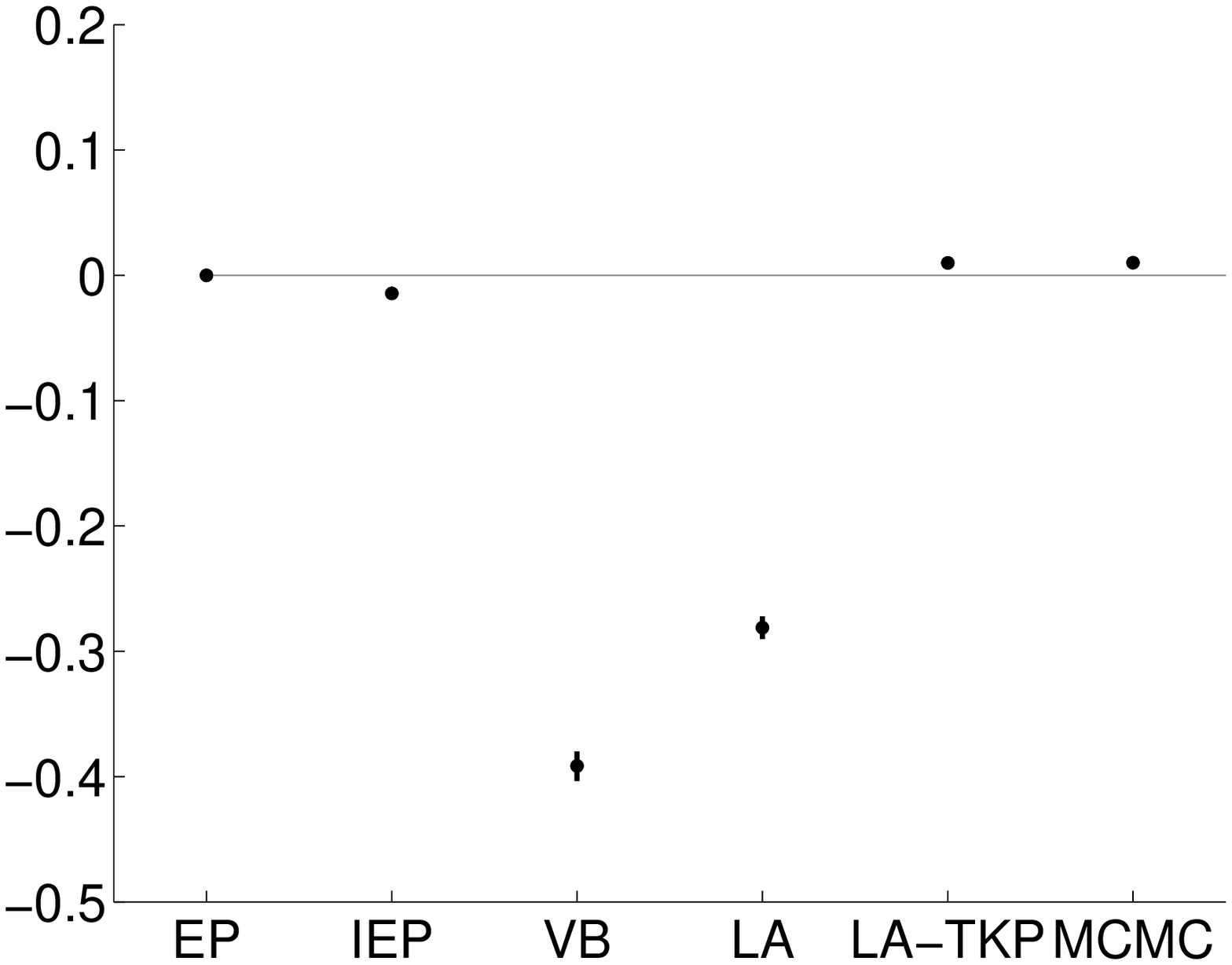}}
  \subfigure[USPS 3-5-7 - pairwise]{\includegraphics[scale=0.22]{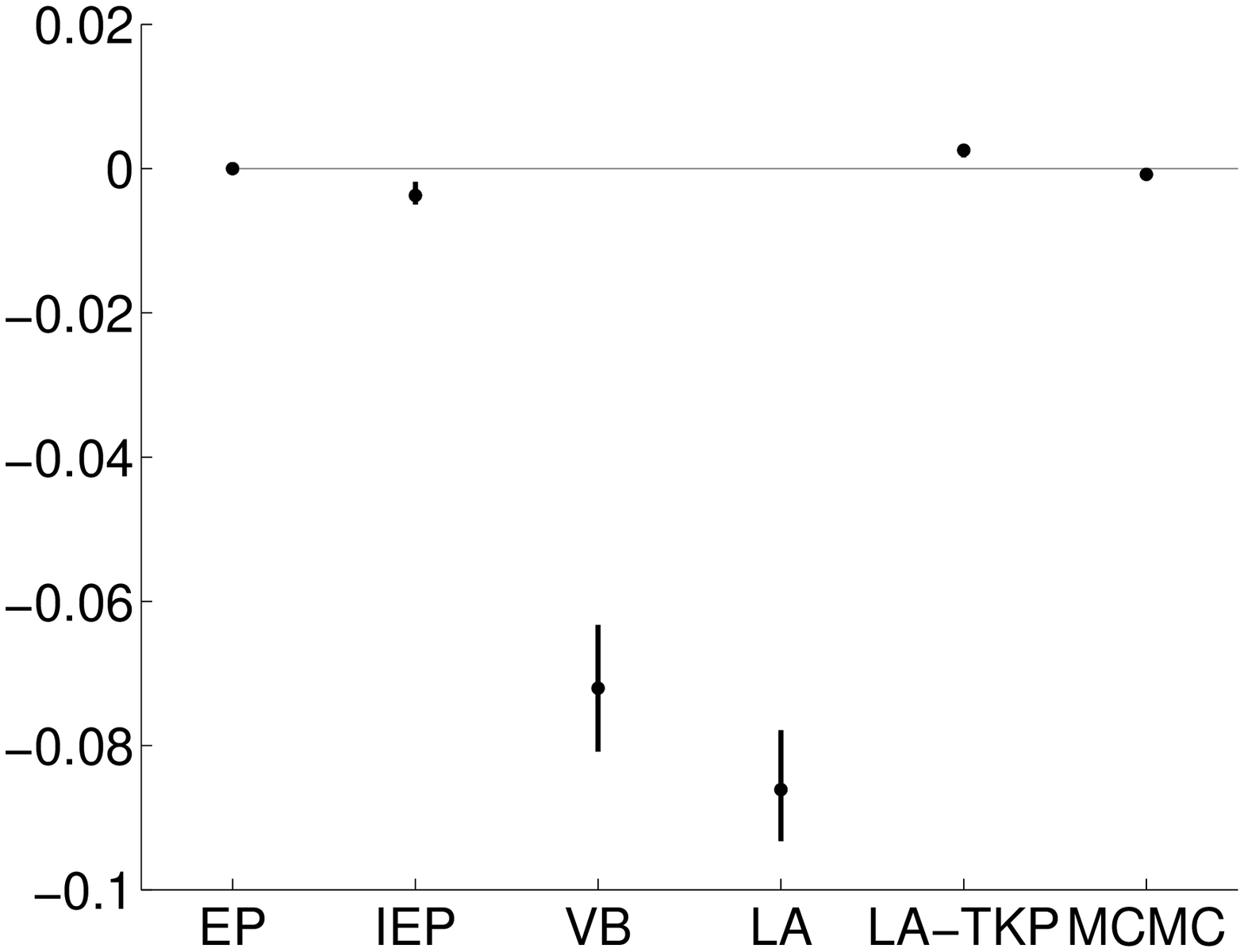}}
  \caption{The first and third rows: The mean log predictive densities
    and their 95\% credible intervals for six datasets (See Table
    \ref{table_datasets}) using EP, IEP, VB, LA, LA-TKP, and MCMC with
    Gibbs sampling. The second and fourth rows: Pairwise differences
    of the log predictive densities with respect to EP (mean + 95\%
    credible intervals). Values above zero indicate that a method is
    performing better than EP. }
  \label{figure_results_lpd}
\end{figure*}

The first and third rows of Figure \ref{figure_results_lpd} visualize
the mean log predictive predictive densities (MLPD) and their 95\%
credible intervals for six datasets estimated using the Bayesian
bootstrap method as described by \citet{vehtari2002}.
To highlight the differences between the methods more clearly, we
compute pairwise differences of the log posterior predictive densities
with respect to EP. The second and fourth rows of Figure
\ref{figure_results_lpd} show the mean values and the 95\% credible
intervals of the pairwise differences. 
The comparisons reveal that EP performs well when compared to MCMC;
only in the Teaching and Image segmentation datasets MCMC is
significantly better. IEP performs worse than EP in all datasets
except Teaching and Glass. The predictive densities of VB and LA are
overall worse than EP, IEP or MCMC. LA-TKP improves the performance of
LA with all datasets except Teaching.

\begin{figure*}[!t]
  \centering
  \subfigure[New-thyroid]{\includegraphics[scale=0.22]{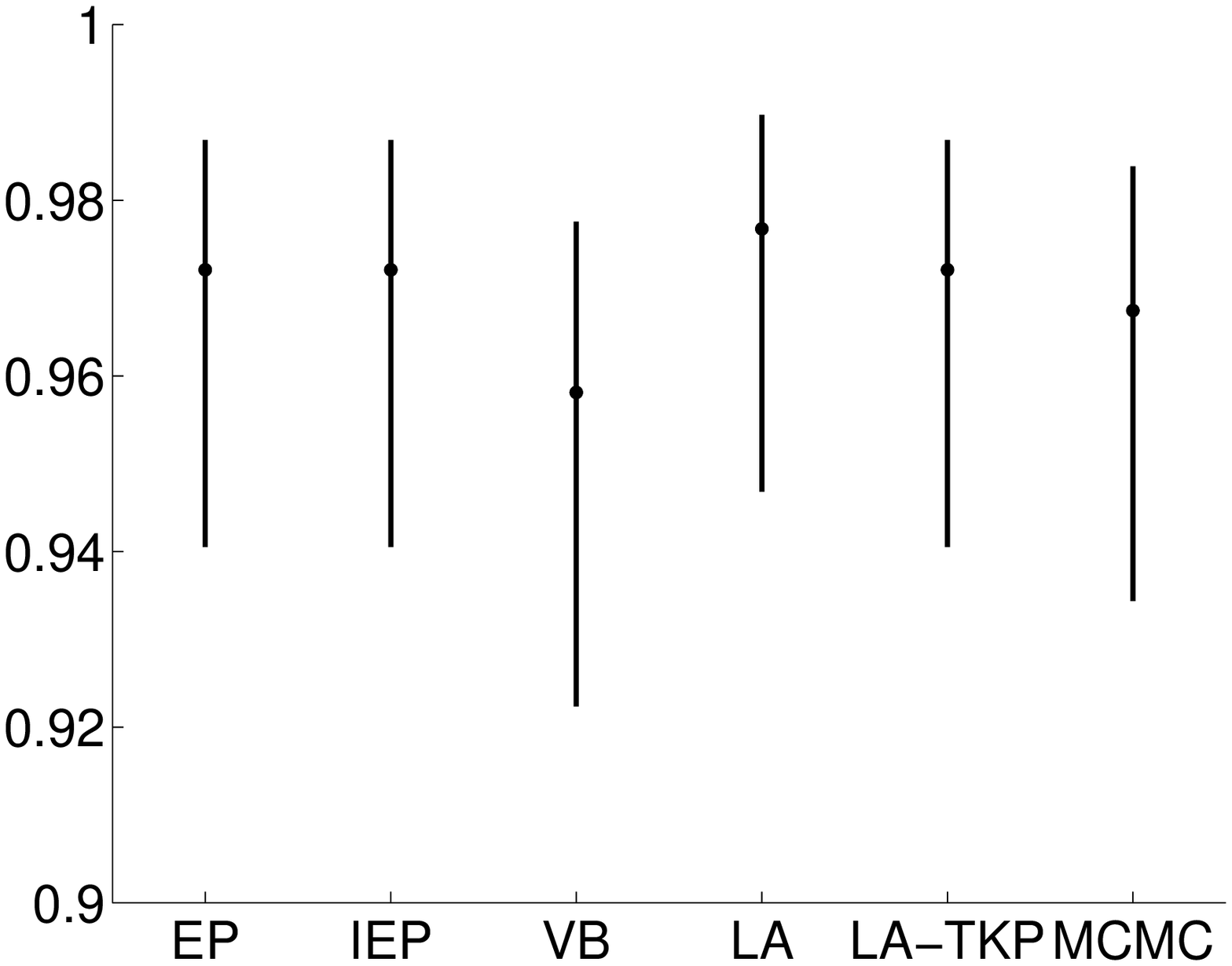}}
  \subfigure[Teaching]{\includegraphics[scale=0.22]{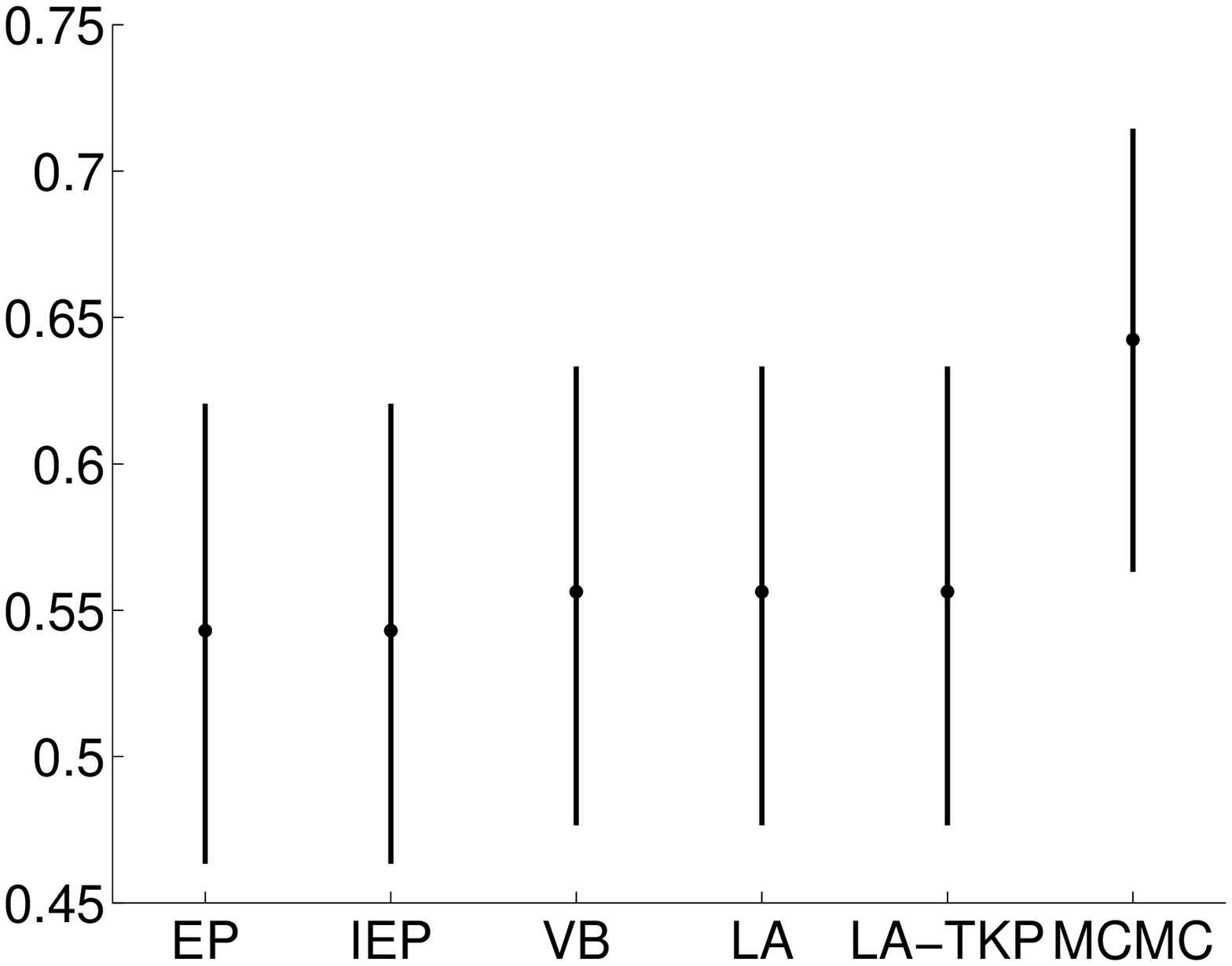}}
  \subfigure[Glass]{\includegraphics[scale=0.22]{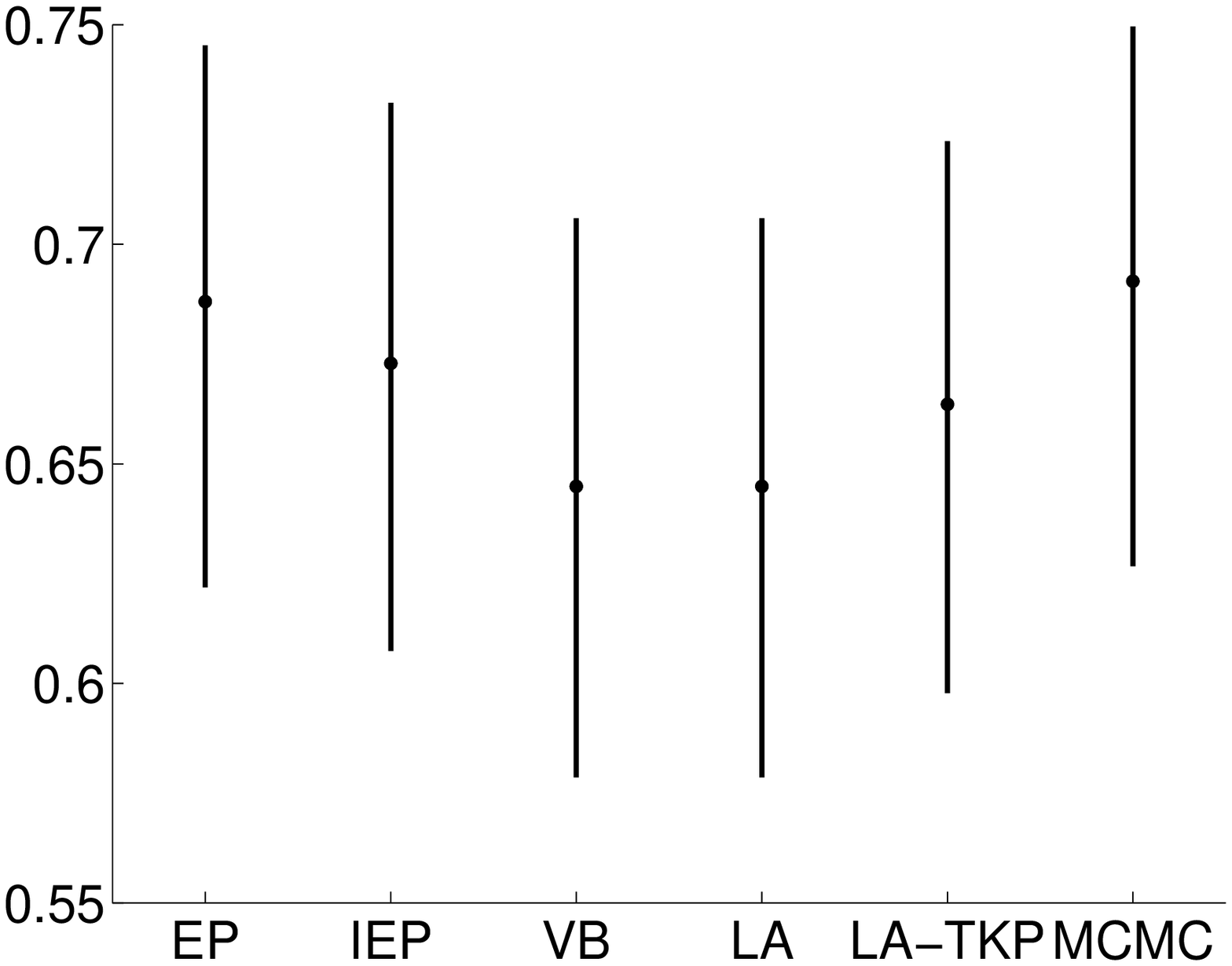}}
  \centering
  \subfigure[New-thyroid - pairwise]{\includegraphics[scale=0.22]{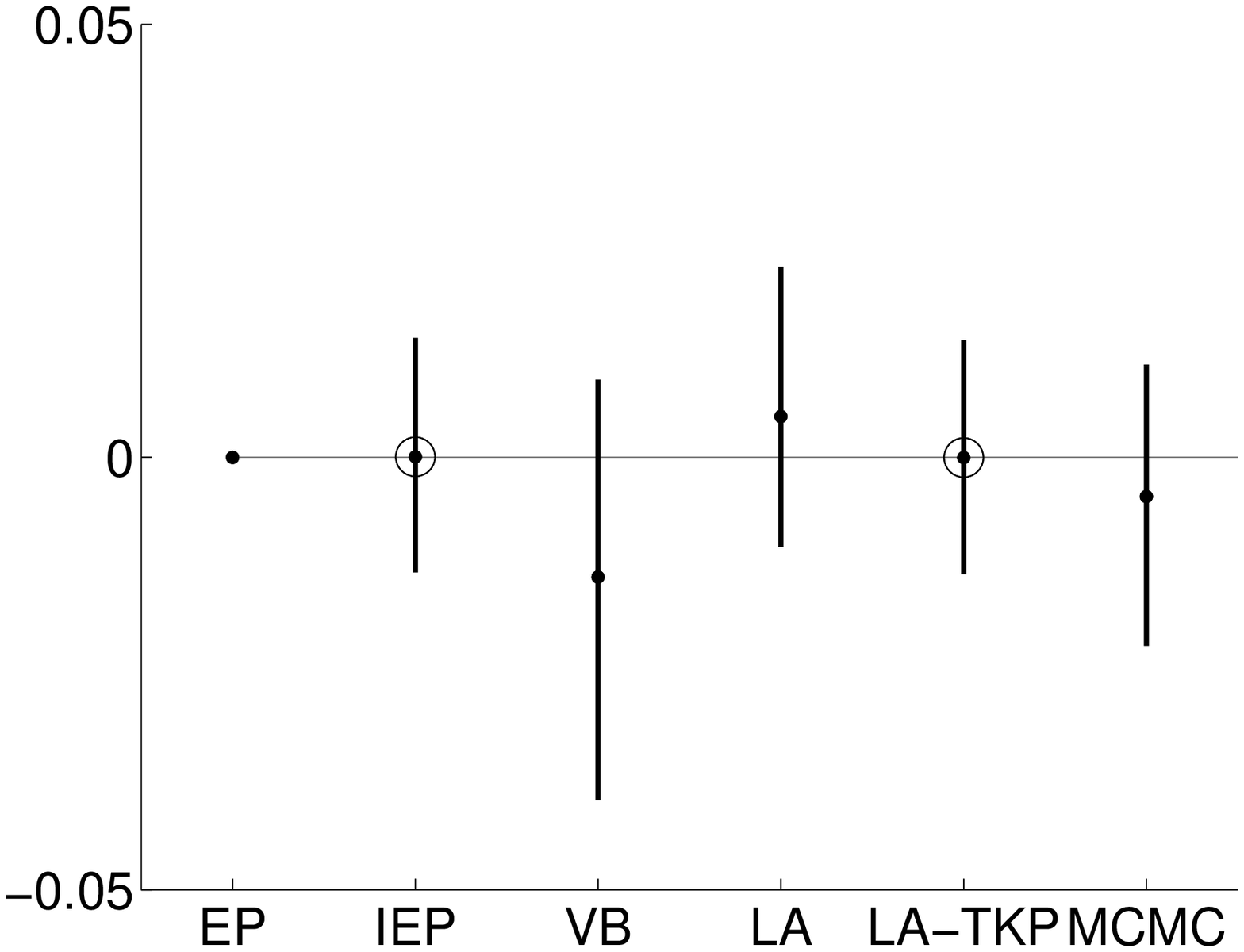}}
  \subfigure[Teaching - pairwise]{\includegraphics[scale=0.22]{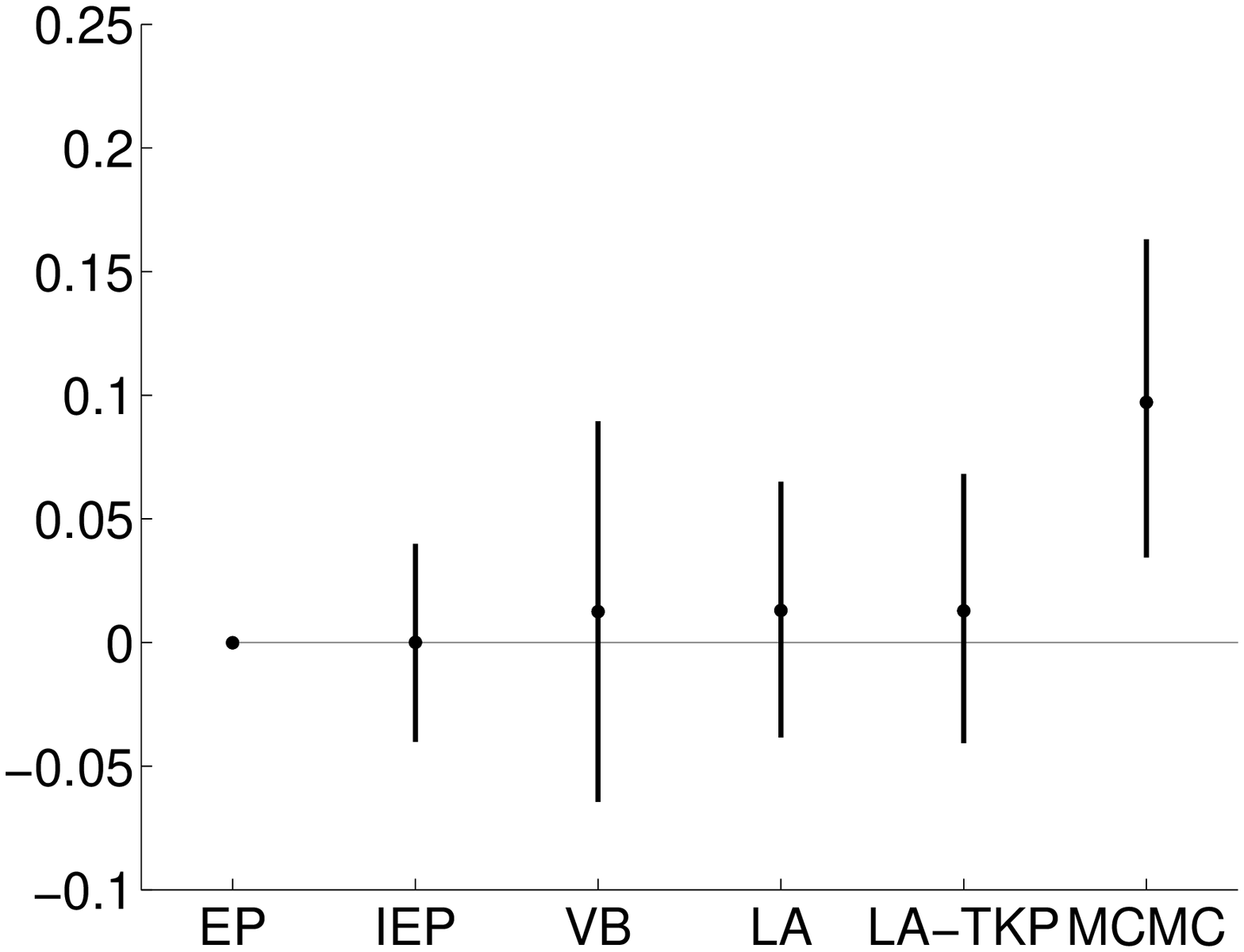}}
  \subfigure[Glass - pairwise]{\includegraphics[scale=0.22]{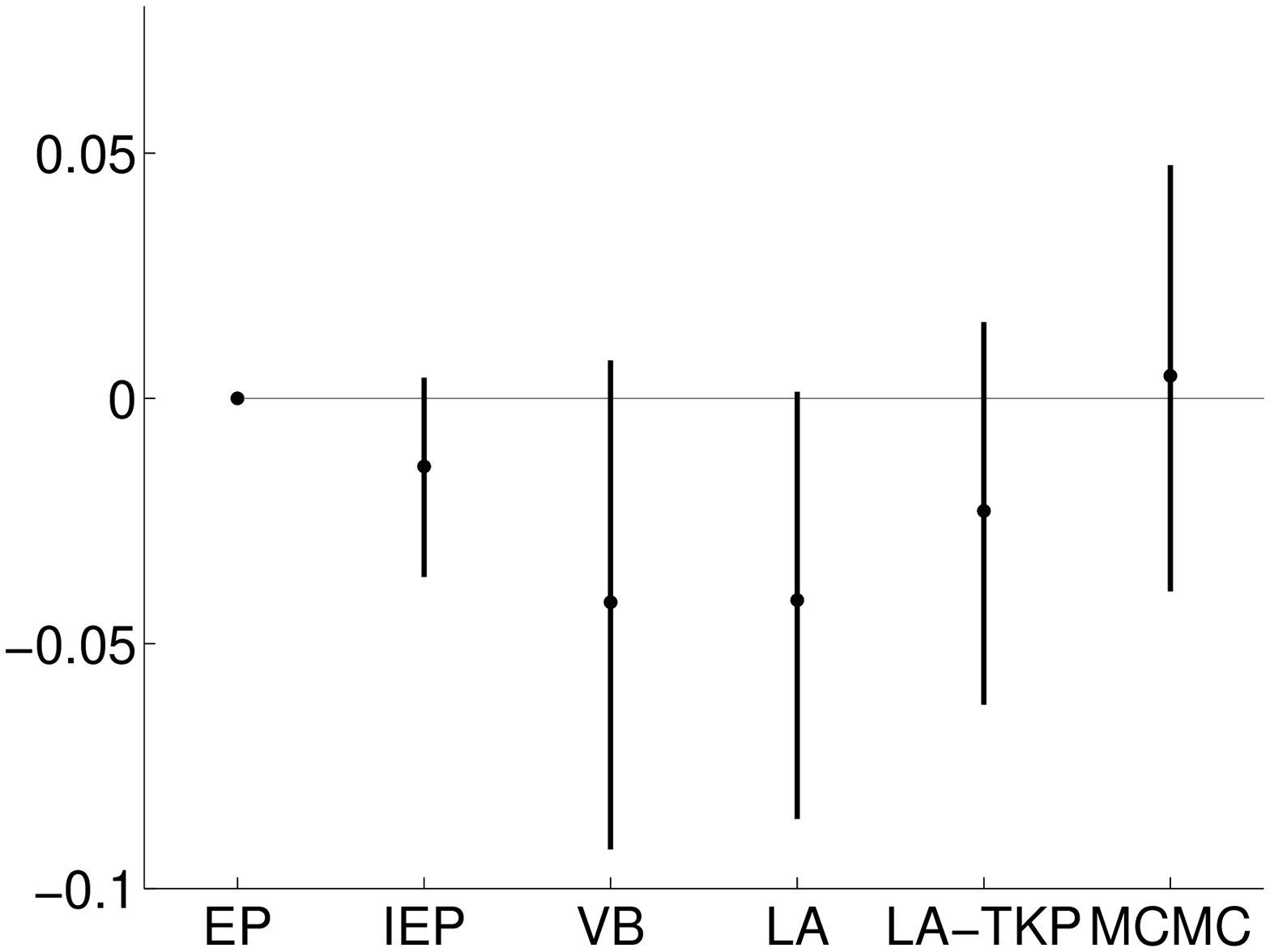}}
  \centering
  \subfigure[Wine]{\includegraphics[scale=0.22]{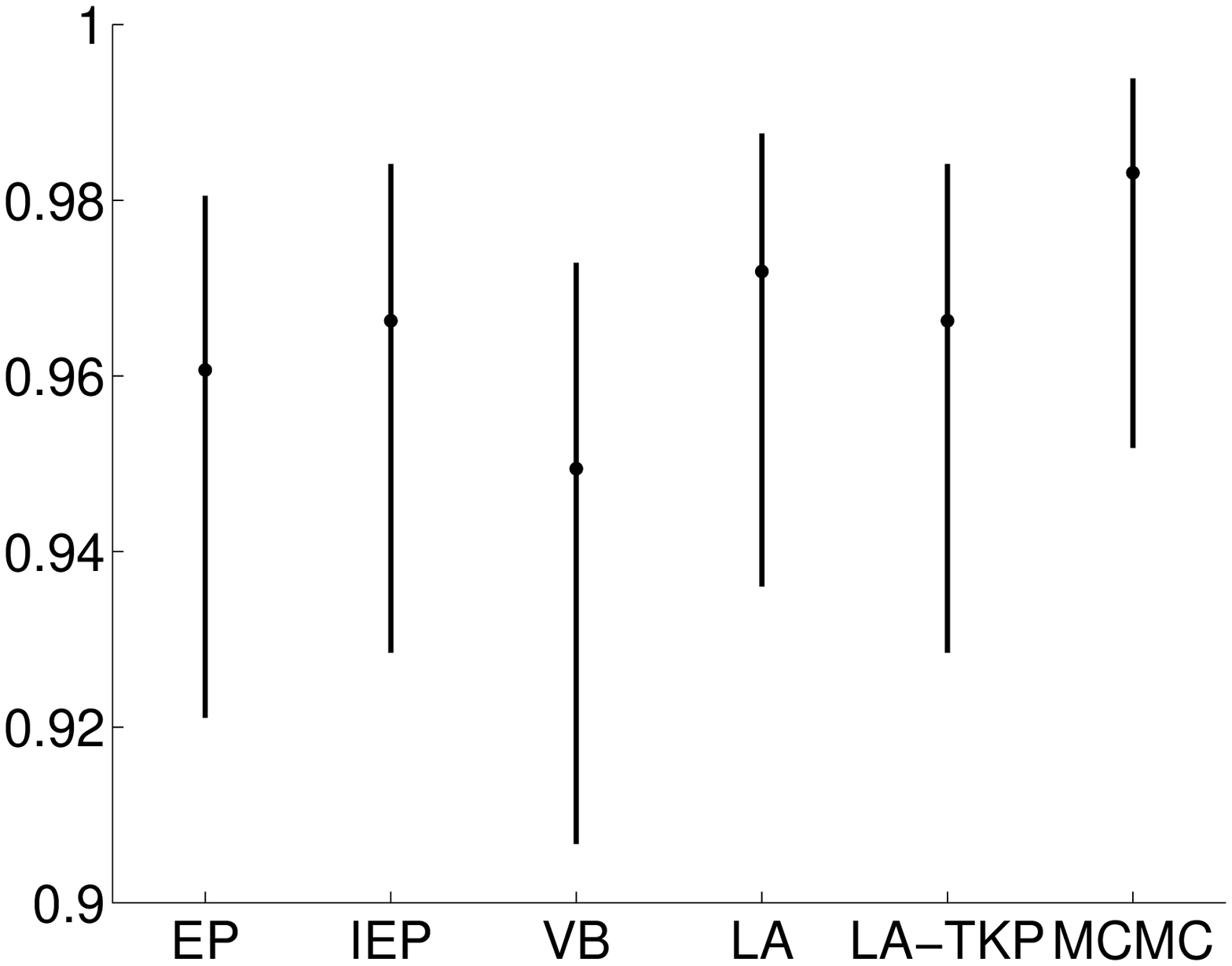}}
  \subfigure[Segmentation]{\includegraphics[scale=0.22]{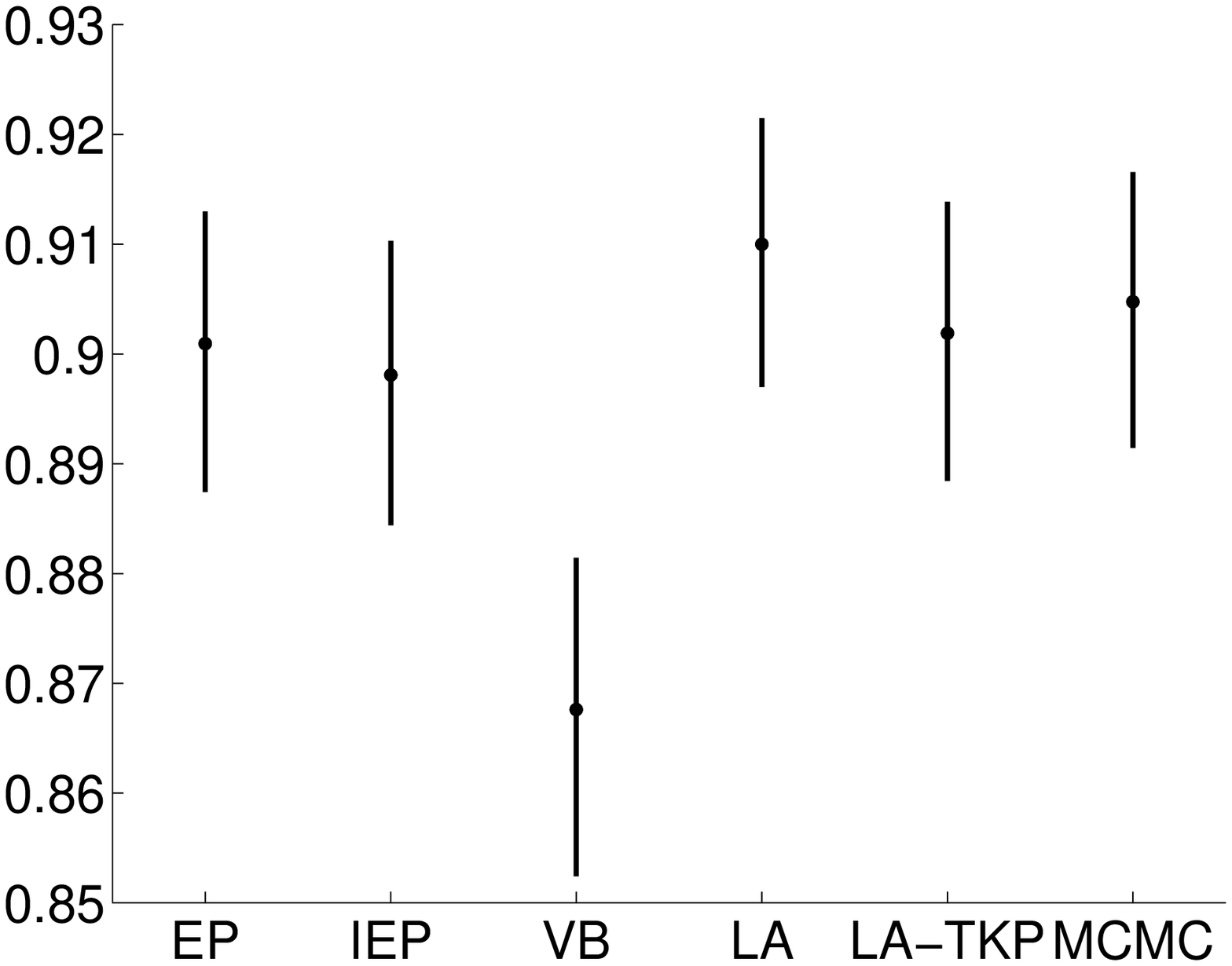}}
  \subfigure[USPS 3-5-7]{\includegraphics[scale=0.22]{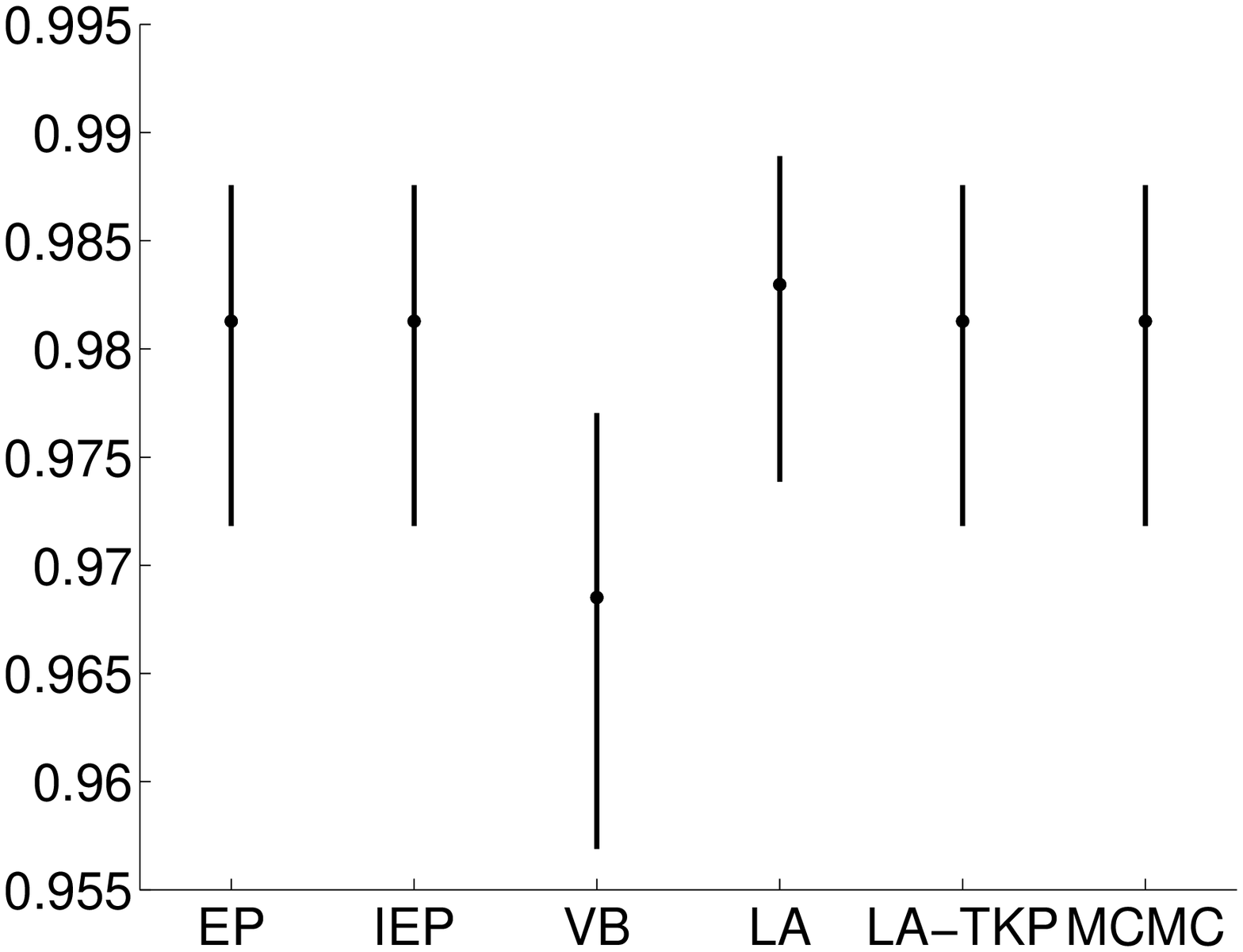}}
  \centering
  \subfigure[Wine - pairwise]{\includegraphics[scale=0.22]{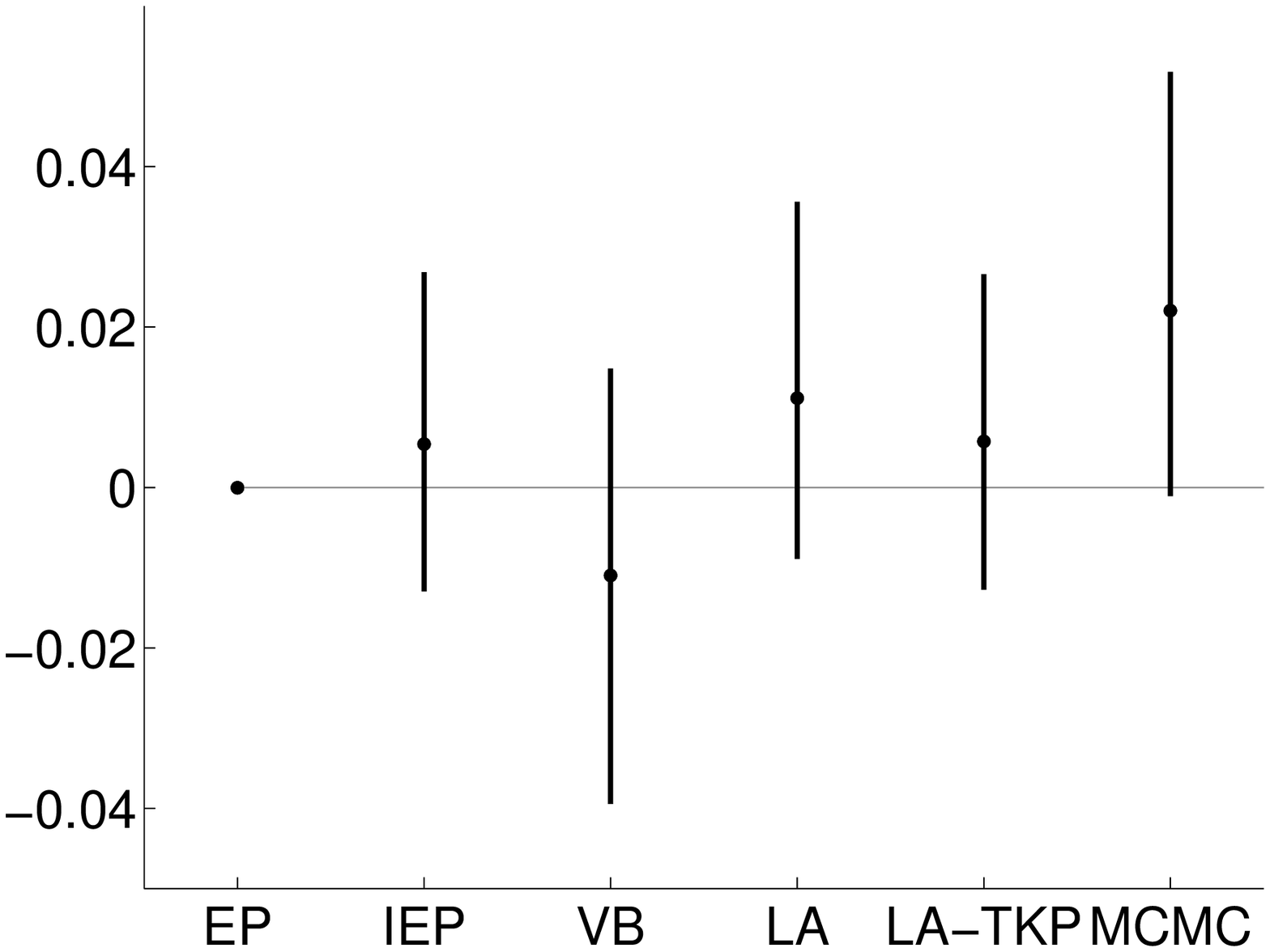}}
  \subfigure[Segmentation - pairwise]{\includegraphics[scale=0.22]{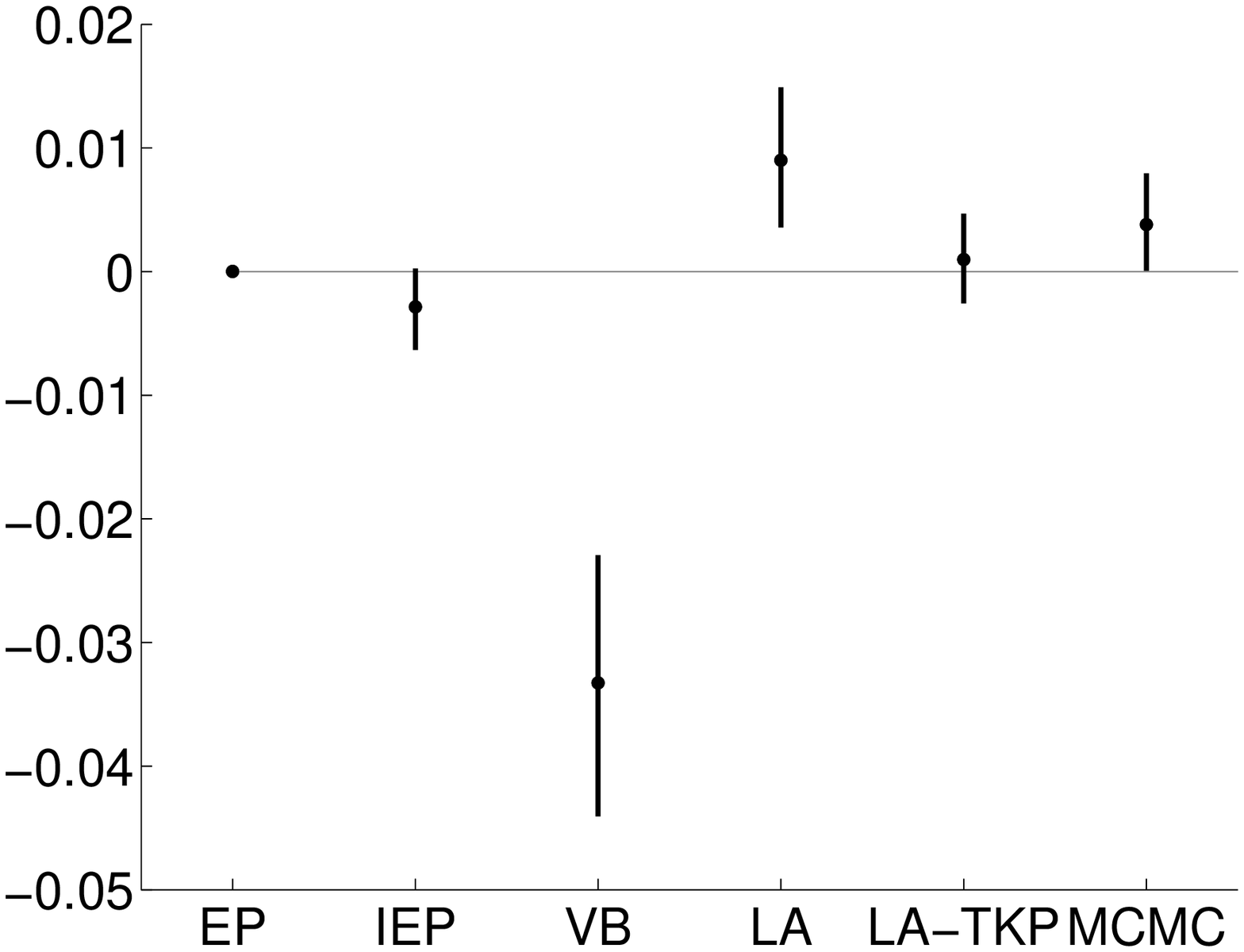}}
  \subfigure[USPS 3-5-7 - pairwise]{\includegraphics[scale=0.22]{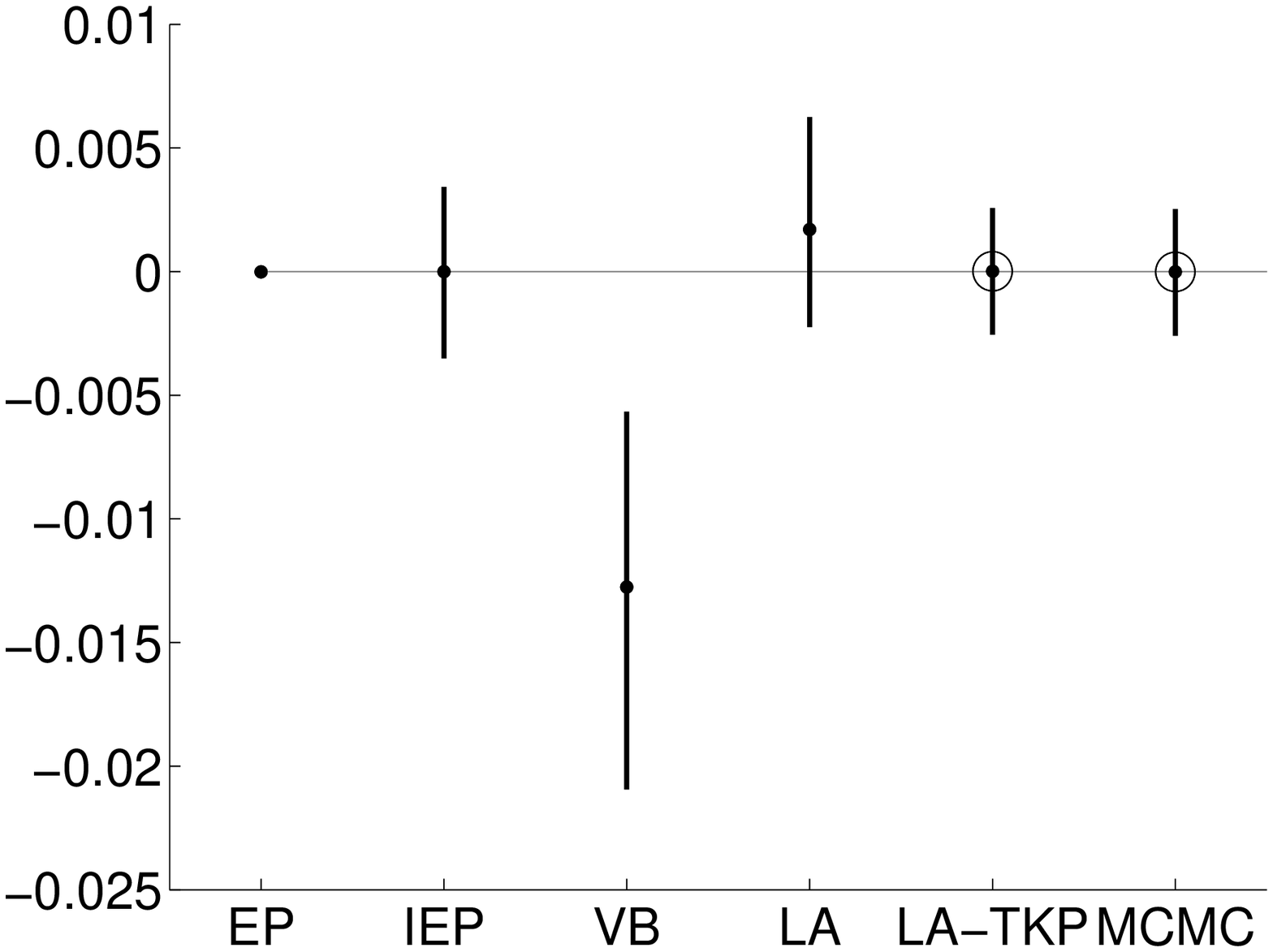}}
  \caption{The first and third rows: The classification accuracies and
    their 95 \% credible intervals for six datasets (See Table
    \ref{table_datasets}) using EP, IEP, VB, LA, LA-TKP, and MCMC with
    Gibbs sampling. The second and fourth rows: Pairwise differences
    of the classification accuracies with respect to EP (mean + 95\%
    credible intervals). Values above zero indicate that a method is
    performing better than EP. A small circle at the mean value is
    plotted if the predictions are exactly the same as with EP. }
  \label{figure_results_acc}
\end{figure*}

The first and third rows of Figure \ref{figure_results_acc} shows the
mean classification accuracies and their 95\% credible intervals,
and the second and fourth rows show the pairwise mean differences of
the classification accuracies together with the 95\% credible
intervals with respect to EP.
Because the difference of the classification outcomes for each
observation is a discrete variable with three possible values (worse,
same, or better than EP), we assume a multinomial model with a
non-informative Dirichlet prior distribution for the comparison test.
In a case where the method has exactly the same predictions as EP, a
small circle is plotted around the mean value. 
The differences between all methods are small. In the Teaching
dataset, where the overall accuracy is the lowest, the MCMC estimate
is significantly better than any other method. There is no
statistically significant difference between EP and IEP; IEP performs
slightly better in the Wine dataset, but EP has better accuracy in the
Glass and Image segmentation datasets, which both have more than three
target classes, and in which the overall classification accuracies are
among the lowest.
LA has good classification accuracy, and performs better than EP
in Image segmentation.
A possible explanation for this is the different shape of the softmax
likelihood function used by LA.
If classification accuracy is the only criterion, the LA-TKP
correction seems unnecessary. VB has the lowest classification
performance and is significantly worse than the other methods in the
Image segmentation and USPS 3 vs. 5 vs. 7 datasets, which is probably
caused by a worse estimate of the hyperparameter values.

\begin{figure*}[!t]
 \centering
  \subfigure[USPS]{\includegraphics[scale=0.2]{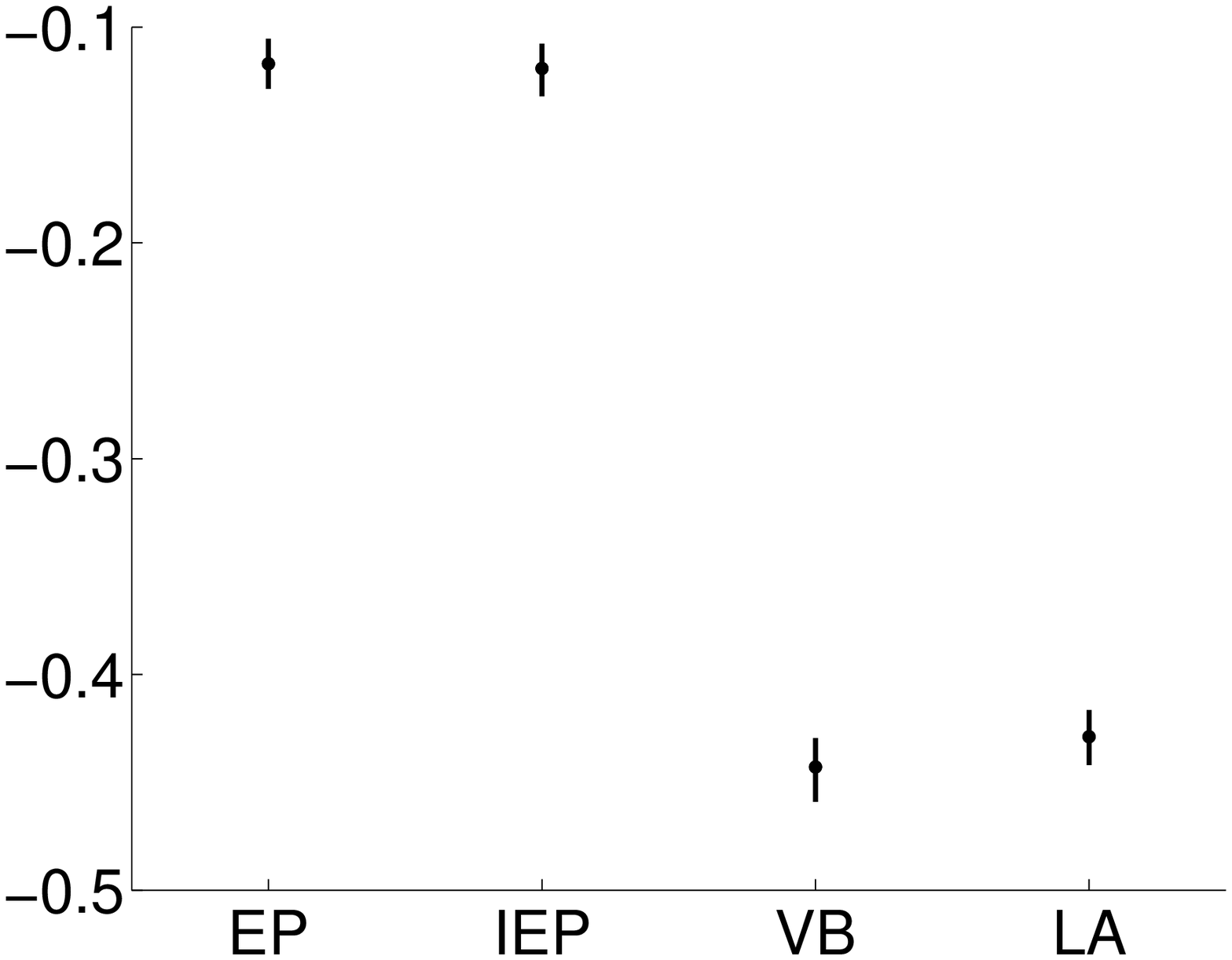}}
  \subfigure[USPS - pairwise]{\includegraphics[scale=0.2]{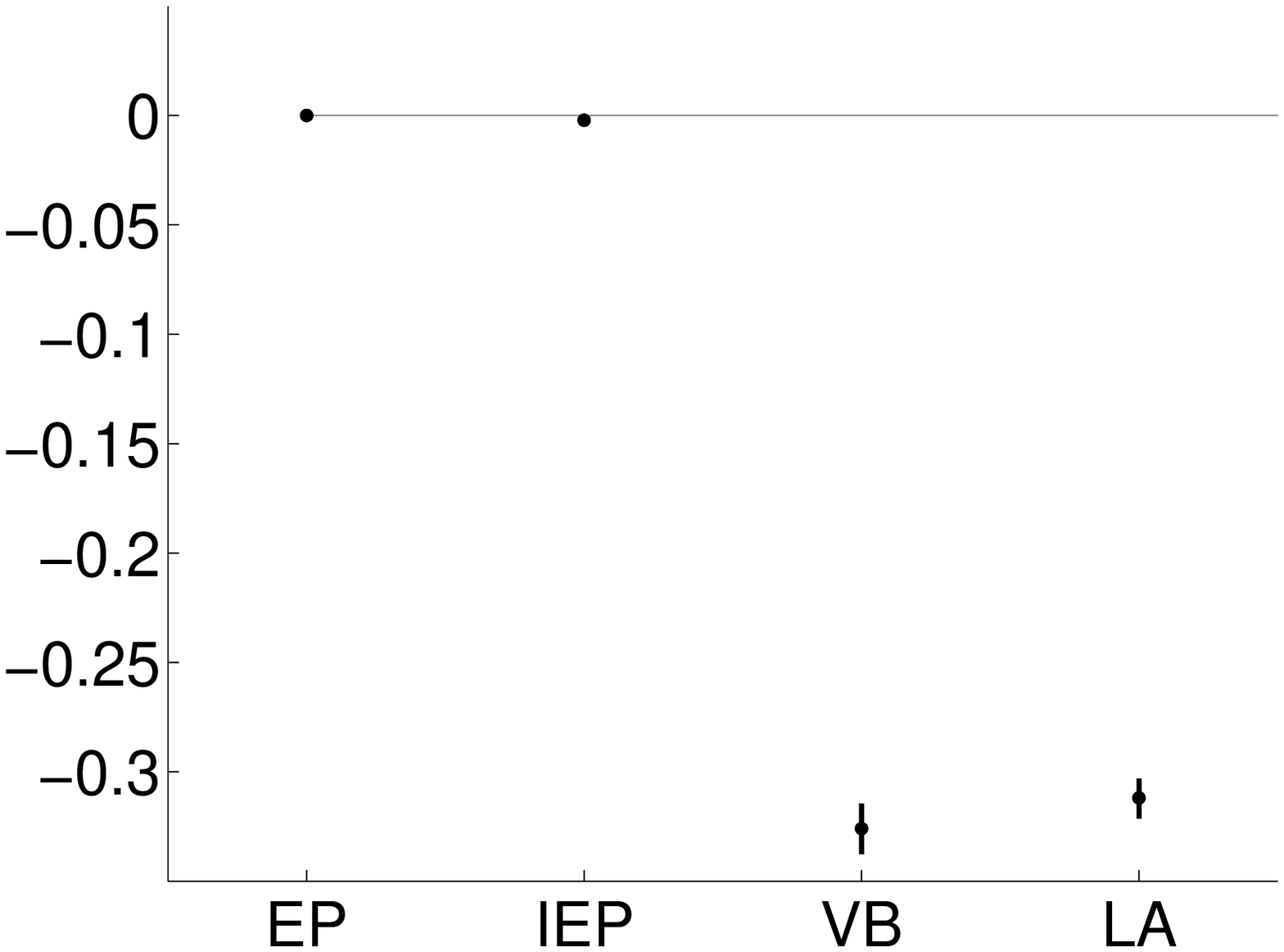}}
  \subfigure[USPS]{\includegraphics[scale=0.2]{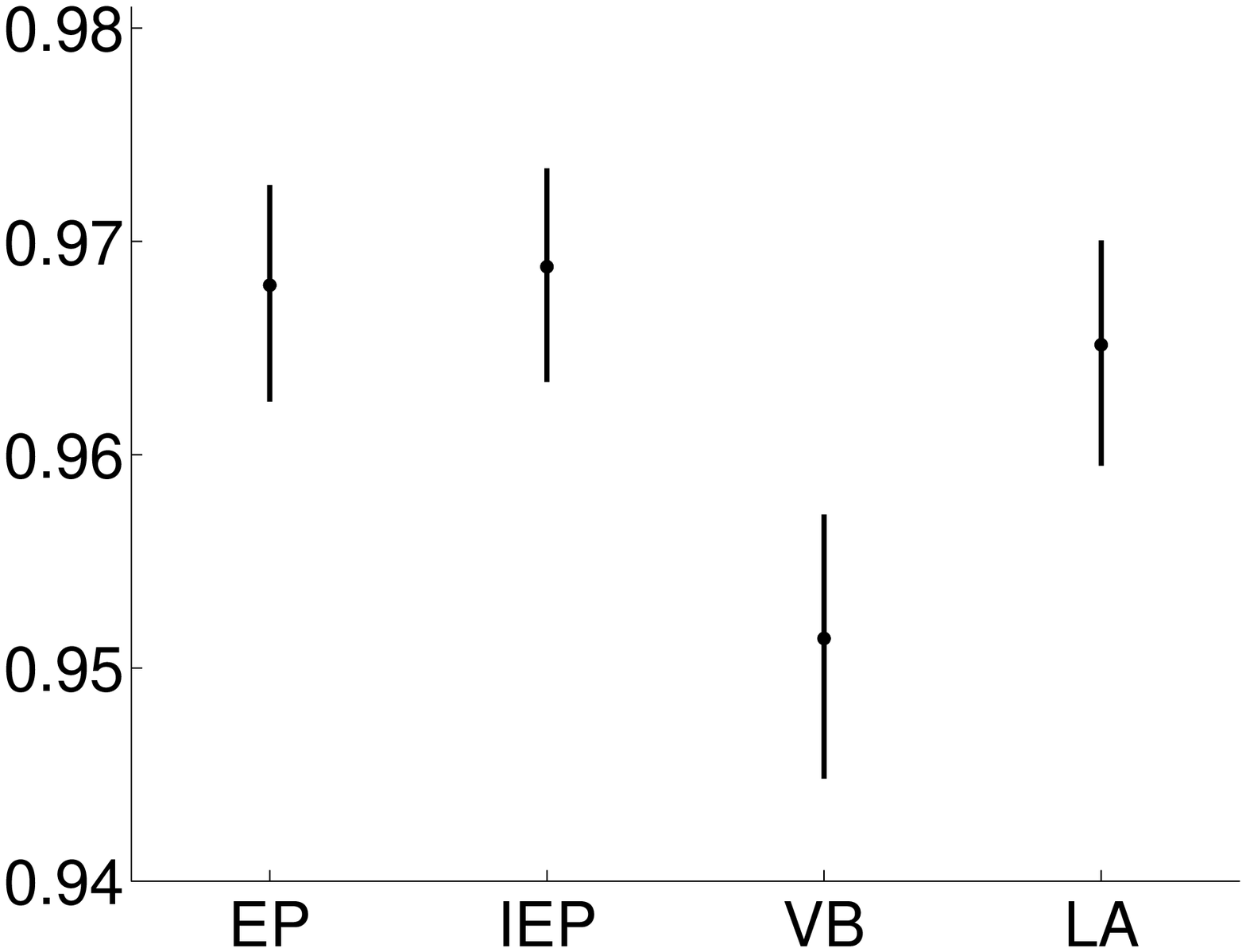}}
  \subfigure[USPS - pairwise]{\includegraphics[scale=0.2]{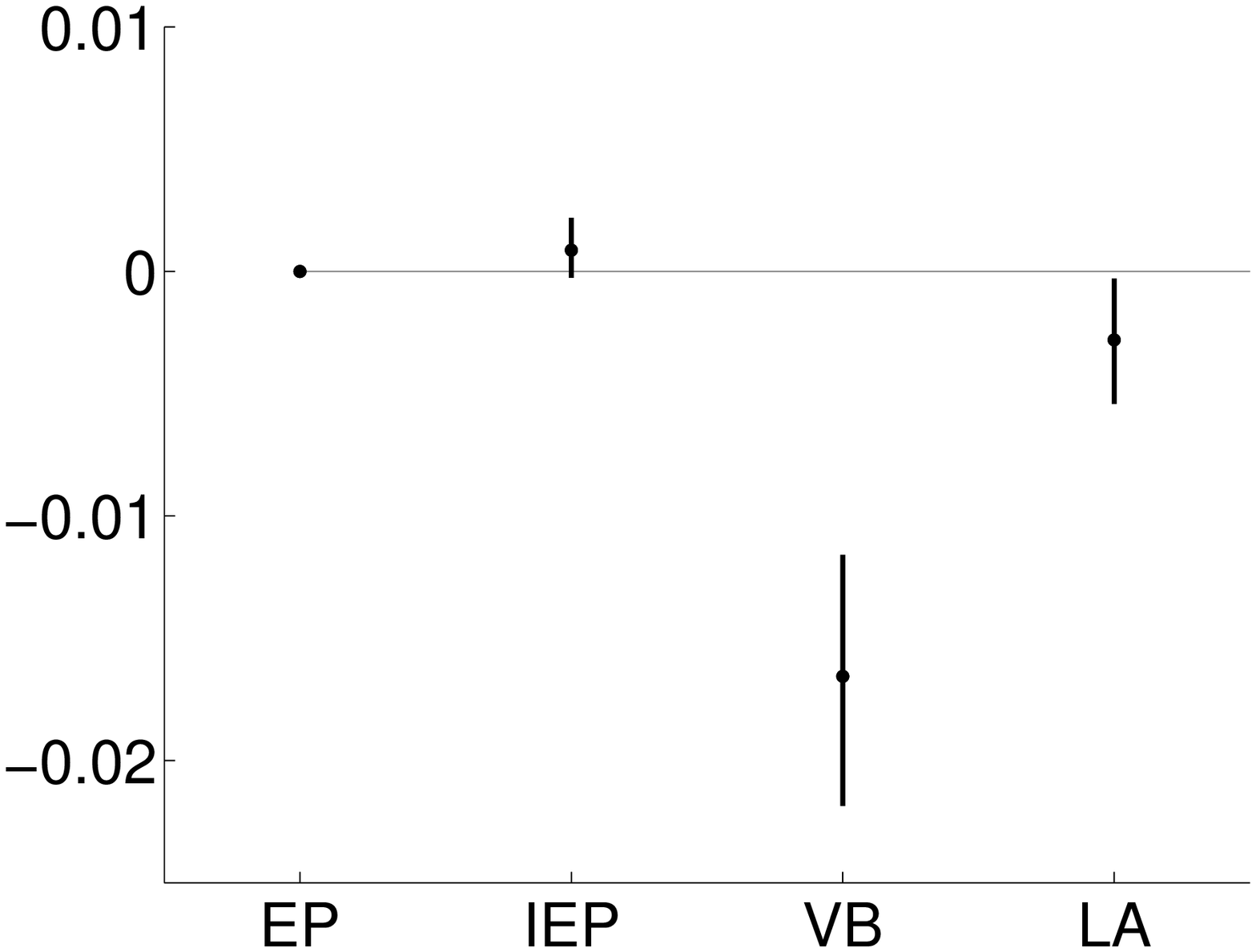}}
  \caption{The mean log predictive densities (a) and classification
    accuracies (c) for the USPS 10-class dataset (See Table
    \ref{table_datasets}) using EP, IEP, VB, and LA. The pairwise
    differences of the log predictive densities and the classification
    accuracies with respect to EP are shown in (b) and (d),
    respectively. In panels (b) and (d) values above zero indicate
    that a method is performing better than EP. In all panels mean and
    95\% credible intervals are shown. }
  \label{figure_results_uspsall}
\end{figure*}

Finally, we summarize the MLPD scores and classification accuracies of
EP, IEP, VB, and LA with the USPS 10-class dataset in Figure
\ref{figure_results_uspsall}. Both EP approaches are significantly
better than VB or LA with both measures. Considering the EP
approaches, EP achieves slightly better MLPD score, whereas IEP is
slightly better in terms of classification accuracy, but the
differences are not statistically significant.

\section{Conclusions and further research}

EP approaches for GP classification with the multinomial probit model
have already been proposed by \citet{seeger2006} and \citet{girolami2007}.
In this paper, we have complemented their work with a novel
quadrature-free nested EP algorithm that maintains all between-class
posterior dependencies but still scales linearly in the number of
classes. Our comparisons show that when the hyperparameters are
determined by optimizing the marginal likelihood, nested EP is
a consistent approximate method compared to full MCMC. In terms of
predictive density, nested EP is close to MCMC, and more accurate
compared to VB and LA, but if only the classification accuracy is
concerned, all the approximations perform similarly. LA-TKP improves
the predictive density estimates of LA but the computational cost
becomes increasingly demanding if larger number of predictions are
needed.

In our comparisons the predictive accuracies of the full EP and IEP 
solutions obtained using the nested EP algorithm are
similar for practical purposes.
However, our visualizations show that the approximate marginal
posterior distributions of the latent values provided by full EP are
clearly more accurate, although the full nested EP solution can be calculated
with similar computational burden than nested IEP. 
Because there is no convergence guarantee for the standard EP
algorithm, it is worth to notice the differences in the convergence
properties of full EP and IEP observed in our experiments.
With the same hyperparameter values, nested IEP converged more slowly and
required more damping than full nested EP.
This can be due to slower propagation of information caused by the
independence assumptions, and this behavior can get worse as the
between-class posterior couplings get stronger with certain
hyperparameter values.
Given all these observations, we prefer full EP over IEP.

Models in which each likelihood term related to a certain observation
depends on multiple latent values, such as the multinomial probit, are
challenging for EP because a straightforward quadrature-based
implementation may become computationally infeasible unless independence 
assumptions between the latent values or some other simplifications are made.
In the presented nested EP approach, we have applied inner EP approximations
for each likelihood term within an outer EP framework in a
computationally efficient manner.
This approach could be applicable also for other similar multi-latent
models which admit integral representations consisting of simple
factorized functions each depending on a linear transformation of the
latent variables.

A drawback with GP classifiers is the fundamental computational
scaling $\mathcal{O}(n^3)$ resulting from the prior structure.
To speed up the inference in multiclass GP classification, sparse
approximations such as the informative vector machine (IVM) have been
proposed \citep{seeger2004,girolami2006,seeger2006}.
IVM uses the information provided by all observations to form an
active subset which is then used to form the posterior mean and
covariance approximations. The presented EP approach could be extended
to IVM in a similar fashion as described by \citet{seeger2004}. The
accurate marginal approximations of full EP could be useful in
determining the relative entropy measures used as a scoring criterion
to select the active set. To speed up the computations, the inner EP
site parameters could be updated iteratively even for the observations
not in the active set in a similar fashion as described in Section
\ref{sec_efficient}.
Recently, a similar approach to IVM called predictive active set
selection (PASS-GP) has been proposed by \citet{henao2010} to lower
the computational complexity in binary GP classification.
PASS-GP uses the approximate cavity and cavity predictive distributions of EP to
determine a representative active set.
The proposed EP approach could prove useful when extending PASS-GP to
multiple classes, because it provides accurate marginal predictive
density estimates.

\newpage

\appendix
\section{Approximating tilted moments using EP}
\label{app_moment}

For convenience, we summarize the inner EP algorithm for approximating
the tilted moments resulting from a multinomial probit likelihood. Essentially
the same algorithm was presented by \citet{minka2001a} for classification with
the Bayes point machine and later by \citet{qi2004} for the binary probit
classifier. To facilitate a computationally efficient implementation, the
following algorithm description is written with an emphasis to reduce the
number of vector and matrix operations in a similar fashion as in the general EP
formulation presented by \citet[][appendix C]{cseke2011}.

We want to approximate the normalization, mean and covariance of the
tilted distribution
\begin{align}
\hat{p}(\w_i)
=\hat{Z}_i^{-1} \mathcal{N}(\w_i|\mw,\Sw) \prod_{j=1,j\neq \yi}^c \Phi(\w_i^T\z_{i,j}).
\end{align}
This is done with the EP algorithm which results in the Gaussian
approximation
\begin{align}
  \hat{q}(\w_i)=Z_{\hat{q}_i}^{-1} \mathcal{N}(\w_i|\mw,\Sw)
  \prod_{j=1,j\neq \yi}^c
  \tilde{Z}_j\mathcal{N}(\w_i^T \z_{i,j}|\nut_{i,j} \taut_{i,j}^{-1},\taut^{-1}_{i,j}),
\end{align}
where we have used the natural parameters $\taut_{i,j}$
(precision) 
and $\nut_{i,j}$ (location) 
for the site approximations. The index $i$ denotes the $i$'th
observation, and to clarify the notation below, we leave out this
index from the inner EP terms.
In the first outer-loop, the site parameters $\taut$ and $\nut$ are
initialized to zero, $\bm{\mu}_{\hat{q}_i}$ to $\mw$, and
$\Sigma_{\hat{q}_i}$ to $\Sw$. After the first outer-loop, these
parameters are initialized to their last values from the previous
iteration for speed-up.
The following steps are repeated until convergence.
\begin{enumerate}

\item Cavity evaluations:
\begin{eqnarray}
v_{-j}&=&(v_j^{-1}-\taut_j)^{-1}\\
m_{-j}&=&v_{-j}(v_j^{-1}m_j-\nut_j),
\end{eqnarray}
where scalars $v_j=\z_{i,j}^T\Sigma_{\hat{q}_i} \z_{i,j}$ and
$m_j=\z_{i,j}^T\bm{\mu}_{\hat{q}_i}$ corresponds to marginal
distribution of latent $\w_i^T\z_{i,j}$.

\item Tilted moments:
\begin{eqnarray}
\hat{Z}_j&=&\Phi(z_j)\\
\hat{m}_j&=&\rho_j\nu_{-j}+m_{-j}\\
\hat{v}_j&=&v_{-j}-v_{-j}^2\gamma_j,
\end{eqnarray}
where $z_j=m_{-j}(1+v_{-j})^{-1/2}$,
$\rho_j=\frac{\mathcal{N}(z_j)}{\Phi(z_j)}(1+v_{-j})^{-1/2}$ and
$\gamma_j=\rho_j^2+z_j\rho_j(1+v_{-j})^{-1/2}$.

\item Site updates with damping:
\begin{eqnarray}
  \Delta\taut_j&=&\delta(\hat{v}_j^{-1}-v_j^{-1})\\
  \Delta\nut_j&=&\delta(\hat{v}_j^{-1}\hat{m}_j-v_j^{-1}m_j),
\end{eqnarray}
where $\delta\in (0,1]$ is the damping factor.

\item Rank-1 covariance update:
\begin{eqnarray}
\Sigma_{\hat{q}_i}^{\mathrm{new}}
&=&\Sigma_{\hat{q}_i}-\bm{\vartheta}_j(1+\Delta\taut_j v_j)^{-1}\Delta\taut_j\bm{\vartheta}_j^T\\
\bm{\mu}_{\hat{q}_i}^{\mathrm{new}}&=&\bm{\mu}_{\hat{q}_i}+\bm{\vartheta}_j(1+\Delta\taut_j v_j)^{-1}(\Delta\nut_j-\Delta\taut_j
m_j),
\end{eqnarray}
where $\bm{\vartheta}_i=\Sigma_{\hat{q}_i} \z_{i,j}$.
\end{enumerate}
Alternatively, the rank-1 updates of step 4 could be replaced by only
one parallel covariance update after each sweep over the sites indexed
by $j$.

\section{Details of posterior computations}

The site covariance is $\tilde{T}=D-DR(R^TDR)^{-1}R^TD$, where the
matrix $R$ is a $cn\times n$ matrix of $c$ times stacked identity
matrices $I_n$, and
$D=\mathrm{diag}\left[\pi_1^1,\ldots,\pi_n^1,\pi_1^2,\ldots,\pi_n^2,\ldots,\pi_1^c,\ldots,\pi_n^c\right]^T$.
To compute predictions at a test point $\x_*$, we need to first
evaluate the mean and covariance of
$\f_*=\left[f_*^1,f_*^2,\ldots,f_*^c\right]^T$ as
\begin{align}
  \mathrm{E}[\f_*]&=K_*^T(I - M K)\tilde{\bm{\nu}}\label{eq_pred_mean}\\
  \mathrm{Cov}[\f_*]&=K_{*,*}-K_*^TMK_*,\label{eq_pred_cov}
\end{align}
where $M=\tilde{T}(I+K\tilde{T})^{-1}$, and $K_*$ is the $c\times cn$
covariance matrix between the test point and the training points, and
$K_{*,*}$ is the $c\times c$ covariance matrix for the test point.
The matrix $M$ in the equations \eqref{eq_pred_mean} and
\eqref{eq_pred_cov} can be evaluated using
\begin{align}
M=B-BRP^{-1}R^TB,
\end{align}
where $B=D^{1/2} A^{-1} D^{1/2}$, $P=R^TD^{1/2} A^{-1} D^{1/2}R$, and
$A=I+D^{1/2} K D^{1/2}$.
%
%
To evaluate $M$, we compute the Cholesky decompositions of $P$ and the
$c$ diagonal blocks of $A$, which results in the scaling
$\mathcal{O}((c+1)n^3)$.
The predictive mean and covariance can be computed by using the block
diagonal structure of $B$ and the sparse structure $K_*$.
Given $\mathrm{E}[\f_*]$ and $\mathrm{Cov}[\f_*]$, the integration
over the posterior uncertainty of $\f_*$ required to compute the
predictive class probabilities, is equivalent to the tilted moment
evaluation, and can be approximated as described in appendix A.

The marginal likelihood approximation of EP can be computed as
\begin{eqnarray}
  \log Z_{\mathrm{EP}}&=& \frac{1}{2}\tilde{\bm{\nu}}^T\bm{\mu}
  -\frac{1}{2}\log |I+K\tilde{T}|
  +\sum_i \log Z_{\hat{q}_i}
  +\frac{1}{2} \sum_i \left(\bm{\mu}_{-i}^T\Sigma_{-i}^{-1}\bm{\mu}_{-i} + \log|\Sigma_{-i}|\right)\nonumber\\
  &&-\frac{1}{2}\sum_i \left(\bm{\mu}_{i}^T\Sigma_{i}^{-1}\bm{\mu}_{i} + \log|\Sigma_{i}|\right)\label{eq_evidence},
\end{eqnarray}
where the vector $\bm{\mu}_{i}$ of length $c$ and the $c\times c$
matrix $\Sigma_{i}$ are the $i$'th marginal mean and covariance.
Similarly, $\bm{\mu}_{-i}$ and $\Sigma_{-i}$ are the cavity mean and
covariance.
The moments $\hat{Z}_i$ in \eqref{eq_evidence} are the normalization
terms from the algorithm described in appendix A.
Finally, the determinant term in \eqref{eq_evidence} can be evaluated
as
\begin{eqnarray}
|I+K\tilde{T}|=|A||R^TDR|^{-1}|P|.
\end{eqnarray}

The gradients of the log marginal likelihood with respect to $\theta$
can be obtained by calculating only the explicit derivatives of the
first two terms of \eqref{eq_evidence}. The implicit derivatives with
respect to the site parameters and cavity parameters (in their natural
exponential forms) of the outer EP cancel each other out in the
convergence \citep{opper2005,seeger2005}. Since the likelihood does
not depend on any hyperparameters, the explicit derivatives of $\log
Z_{\hat{q}_i}$ are zero. Also the implicit derivatives of $\log
Z_{\hat{q}_i}$ with respect to the inner EP parameters cancel out
because these terms are formed as marginal likelihood approximations
with the inner EP, which has the same previously mentioned property of
the EP algorithm.

\vskip 0.2in
\bibliography{multiep}

\end{document}